# A General Statistic Framework for Genome-based Disease Risk Prediction

Long Ma[1], Nan Lin[1] and Momiao Xiong[1,*]

[1] Division of Biostatistics, The University of Texas School of Public Health, Houston, TX 77030, USA

Running Title: biomarker identification for risk prediction

*Address for correspondence and reprints: Dr. Momiao Xiong, Human Genetics Center, The University of Texas Health Science Center at Houston, P.O. Box 20186, Houston, Texas 77225, (Phone): 713-500-9894, (Fax): 713-500-0900, E-mail: Momiao.Xiong@uth.tmc.edu.

## Abstract

Fast and more economical next generation sequencing (NGS) technologies will generate unprecedentedly massive and highly-dimensional genomic and epigenomic variation data. In the near future, a routine part of medical records will include the sequenced genomes. How to efficiently extract biomarkers for risk prediction and treatment selection from millions or dozens of millions of genomic variants raises a great challenge. Traditional paradigms for identifying variants of clinical validity are to test association of the variants. However, significantly associated genetic variants may or may not be usefulness for diagnosis and prognosis of diseases. Alternative to association studies for finding genetic variants of predictive utility is to systematically search variants that contain sufficient information for phenotype prediction. To achieve this, we introduce concepts of sufficient dimension reduction (SDR) and coordinate hypothesis which project the original high dimensional data to very low dimensional space while preserving all information on response phenotypes. We then formulate a clinically significant genetic variant discovery problem into the sparse SDR problem and develop algorithms that can select significant genetic variants from up to or even ten millions of predictors with the aid of a split-and-conquer approach.  The sparse SDR is in turn formulated as a sparse optimal scoring problem, but with penalty which can remove row vectors from the basis matrix. To speed up computation, we apply the alternating direction method of multipliers to solving the sparse optimal scoring problem which can easily be implemented in parallel. To illustrate its application, the proposed method is applied to a coronary artery disease (CAD) dataset from the Wellcome Trust Case Control Consortium (WTCCC) study, Rheumatoid Arthritis (RA) dataset from the GWAS of North American Rheumatoid Arthritis Consortium (NARAC) and  the early-onset myocardial infarction (EOMI) exome sequence datasets which have European origin from the NHLBI's Exome Sequencing Project.

## Introduction

Although there is heated debate about whether DNA variation has value to predict diseases[1-4], genetic variations that are being generated by next generation sequencing (NGS) technologies will provide invaluable information for disease prediction and prevention[5]. However, how to efficiently extract genetic variants for risk prediction and treatment selection from millions or dozens of millions of genomic variants is among the biggest challenges in genomic research and medicine[6]. Innovative approaches should be developed to address these challenges.

Traditional paradigms for identifying variants of clinical validity test association of the variants[7]. The variants are ranked by P-values or odds ratios. A few "top-SNPs" with the smallest P-values are then used to predict disease risk of individuals[8,9]. However, most variants have small effect sizes and the often low to moderate heritability of common diseases. Many associated genetic variants, on their own, have little or no predictive utility. The studies based on a few number of the loci demonstrate limited predictive value of disease[10-14]. The best predictors may not always be at the top of the list and variable selection by P-values will miss many important predictors[8].

To overcome this limitation, rather than using a few highly significant SNPs from genome-wide association studies (GWAS), some researchers use penalized techniques such as the penalized pseudo-likelihood approaches and the bridge regression[15], the LASSO[16], the SCAD and other folded-concave penalties[17]. The Dantzig selector[18] and their related methods[19] have recently been developed to directly search a set of SNPs for disease risk prediction from a large number of genetic variants[20] with mild success.

The widely used methods for disease risk prediction have several serious limitations. Dimension reduction in which we attempt to identify linear or nonlinear combinations of the original set of variables or chose a subset of variables from the original set[21] is an essential issue for genome-based prediction of common diseases. The first limitation of the current methods for disease risk prediction is to use unsupervised dimension reduction in which we fail to make use of the phenotype information.  The second limitation of these methods is their lack of algorithms to systematically search millions of genetic variants for disease risk prediction. We observe that the maximum number of features which the LASSO type methods are predictive is less than 50,000. Simple application of the penalized pseudo-likelihood approaches to millions or dozen of millions of features is infeasible. How to select an informative subset of features from millions of variables is still an open question. The genome-based prediction of common diseases often demands extremely intensive computations. A practical solution is to employ the power of parallel computing. The third limitation is that it is difficult to adapt the current methods for risk prediction to parallel computing.

The purpose of this paper is to address the limitations of these traditional methods for disease risk prediction.  Dimension reduction and variable selection are a dominant theme in the genome-based disease risk prediction. To overcome the most important limitation of the currently used unsupervised dimension reduction in disease risk prediction, which will lose individual disease status information, we propose a supervised dimension reduction method in which both genetic variant information and disease information will be used.   An essential concept in supervised dimension reduction is sufficient dimension reduction (SDR) and testing of the coordinate relation hypothesis[22,23].  Since the dimension of genomic variation data is extremely high and the most genetic variants across the genome are irrelevant with disease, we replace the original predictor vector (genetic variants) with its prediction onto a low dimensional space of the predictor space –which results in a minimal set of linear combinations of the original predictors, without loss of information on disease or response[24,25]. Such low dimensional space is called dimension reduction subspace[22]. Minimum dimension reduction subspace is referred to as central subspace (CS). All information of the predictors about disease or response in the genetic dataset is contained in the central subspace. The central subspace provides sufficient information to predict response or risk. We only need to study relationships between responses (disease risk) and variables in the central space regardless of the original predictors.

The number of methods including sliced inverse regression (SIR)[26], sliced average variance estimation[27], principal Hessian direction[25,28], directional regression[29], and likelihood-based SDR[30] have been developed to estimate the central subspace. Although great progress in SDR has been made the most current methods for dimension reduction still have serious limitation in their practice applications. The first limitation is that the current popular methods for the SDR require the inverse of the sample covariance matrix of the predictors which are often near singular for large number of predictors[31]. Second, the CS estimation needs to use all the original predictors. Therefore, the estimations are not efficient due to the presence of a lot of irrelevant and noise data and the results are difficult to interpret[32]. These methods cannot be used to identify genetic variants of clinical significance.

In the past several years, the variable selection methods have been explored in the SDR analysis to address these two problems. The variable selection methods for the SDR include model-free variable selection[33], the shrinkage sliced inverse regression[34], sparse SDR[35], constrained canonical correlation and variable filtering for dimension reduction[36]. The common feature of these methods is that the identification of CS and variable selection are separated. At the first step, the relevant SDR methods are used to estimate the CS. At the second step, the regularization methods are applied to the estimated basis vector forming the CS to select predictors. In addition, these methods still cannot solve the singularity problem of the covariance matrix of the predictors if at the first stage all predictors are used for SDR. Chen et al.[32] proposed coordinate-independent sparse sufficient dimension reduction and variable selection for SDR which can simultaneously reduce dimension and select variables. However, they need to partition all response variables into slices and still require the inverse of the sample covariance matrix of the predictors. As a consequence, their method can only solve SDR with a small number of predictors. To avoid the inverse of covariance of predictors, Wang and Zhu[31] used the optimal scoring interpretation for SIR and reformulated the sparse SIR problem into sparse optimal scoring-based SDR. However, their variable selection is carried out for each vector in the CS. The variables cannot simultaneously be selected for all vectors in the CS. Joint CS construction and variable selection cannot fully be implemented. Furthermore, the optimal scoring-based SDR still cannot allow solving large SDR problems with up to millions of predictors.

To overcome these limitations we formulate a clinically significant genetic variant discovery problem into a novel sparse SDR problem in which the variables are selected across all the vectors in the CS. The number of variables which the current sparse classification and regression problem can deal with cannot exceed 50,000 variables regardless of what computer is used. To systematically search millions or tens of millions of genetic variants that contain sufficient information for disease prediction across the genome, we resort to a split-and-conquer approach. We propose to split the whole genome into a number of sub-genomic regions. We show that the SDR for the whole genome can be partitioned into a number of sub-SDR problems defined for divided genomic regions. For each sub-genomic region, we solve the sparse SDR problem in which we perform joint dimension reduction and variable selection across the vectors in the CS. The results from each sub-genomic region are then combined to obtain an overall result that is equivalent to the global sparse SDR problem across the genome. Solving large sparse SDR problems requires heavy computation. A core part of the algorithm for sparse SDR is convex optimization. To speed up computation, we employ the alternating direction method of multipliers for convex optimization which can easily be implemented in parallel. To illustrate its application, the proposed method for genome-based disease risk prediction is applied to the CAD dataset from the WTCCC study, RA dataset from North American Rheumatoid Arthritis Consortium (NARAC) study and the EOMI exome sequence dataset which includes European origin from the NHLBI's Exome Sequencing Project. A program for implementing the developed sparse SDR algorithm to genome-wide search variants of clinical utility can be downloaded from our website http://www.sph.uth.tmc.edu/hgc/faculty/xiong/index.htm.

## Methods

### Sufficient Dimension Reduction

Let $Y$ be a univariate response variable (phenotype) which can be a continuous or discrete variable and $X$ be a $p$ dimensional vector of predictors (genotypes for genetic variants). Since dimension of genomic variation is extremely high, dimension reduction can be used to reduce the impact of noise and irrelevant predictors on risk prediction. Dimension reduction is to identify a linear or nonlinear combination of the original set of variables while preserving relevant information[37]. We have two classes of dimension reduction: unsupervised dimension reduction and supervised dimension reduction. Principal component analysis (PCA) is a typical method for unsupervised dimension reduction which projects predictor data onto a linear space without response variable information. Supervised dimension reduction is to discover the best subspace that maximally reduces the dimension of the input while preserving the information necessary to predict the response variable. The current popular supervised dimension reduction method is SDR which aims to find a linear subspace S such that the response Y is conditionally independent of the covariate vector X, given the projection of X on S:

$$Y \perp X \mid P_S X \,, \tag{1}$$

where $\perp$ indicates independence and $P_S$ represents a projection on $S$. In other words, all the information of $X$ about $Y$ is contained in the space $S$. The subspace S is referred to as a dimension reduction subspace. The subspace $S$ may not be unique. To uniquely describe dimension reduction subspace, we introduce central subspace (CS) that is defined as the intersections of all dimension reduction subspaces $S$, if it is also a dimension reduction subspace[24]. The CS is denoted by $S_{Y|X}$.

Many methods have been developed for identifying CS. These methods are based on inversion regression[26]. Let $\Sigma_x$ be a covariance matrix of $X$. We define the standardized predictors as

$$Z = \Sigma^{-1/2}(X - E(X)) \,. \tag{2}$$

The CS for the standardized predictors $Z$ is denoted by $S_{Y|Z}$. It is clear that $S_{Y|Z} = \Sigma^{1/2} S_{Y|X}$.

It is known that the eigenvectors $\{\gamma_1,...,\gamma_k\}$ satisfying the following eigenequation:

$$\mathrm{cov}(E(Z \mid Y))\gamma = \lambda_z \gamma \tag{3}$$

form a basis for the CS $S_{Y|Z}$ (See Supplemental Note A). Similarly, a basis for the CS $S_{Y|X}$ can be generated by the eigenvectors $\{\beta_1,...,\beta_k\}$ that satisfies the following eigenequation (Supplemental Note A):

$$\text{cov}(E(X - E(X)|Y))\beta = \lambda_x \Sigma_x \beta \,. \tag{4}$$

**Coordinate Hypothesis**

Although it is a powerful tool for dimension reduction, SDR also has serious limitations. SDR uses all of the original predictors to estimate the CS. The results are difficult to interpret and cannot be used for discovery of predictive variants. To overcome these limitations, we can remove irrelevant predictors while preserving all information about response variable. This idea can be formulated by introducing the concept of coordinate hypothesis which assumes that some coordinate variables (components) in the basis vectors for the CS are zero, i.e., their corresponding original variables will make no contribution to the projection of the original predictors on the CS. Let $H$ be a selected $r$ dimensional subspace of the predictor space that specifies the hypothesis of a set of components in the basis vectors being zero. We test the coordinate hypothesis of the form:

$$P_H S_{Y|X} = O_p \,, \tag{5}$$

where $O_p$ represents the origin in $R^p$. For example, by arranging the order of variables in the dataset, we can always partition the predictor dataset $X$ into two parts: $X = [X_1^T, X_2^T]^T$. The corresponding basis matrix for CS can also be partitioned as $B = [\beta_1^T, \beta_2^T]^T$. Let $H = \text{Span}((0, \text{I}_r)^T)$. Equation (5) implies

$$P_H S_{Y|X} = (0, I_r)\begin{bmatrix}\beta_1\\ \beta_2\end{bmatrix} = \beta_2 = 0\,.$$

Equation (5) provides a general framework for SDR-based variable selection. Since the number of variables involved in genome-based disease risk prediction may reach as high as millions or ten millions of variables, in practice, it is difficult to solve such large SDR-based variable selection problems. Fortunately, we can use a split-and-conquer approach to solve this problem. We can show that SDR for a whole genome can be partitioned into a number of small sub-SDR problems defined for divided small genomic regions (See Supplemental Note B). The combined sub-SDR solutions for genomic regions will globally solve the SDR for a whole genome.

**Reformulation of SIR for SDR as an Optimization Problem**

The SIR for estimation of the CS can be formulated as an eigenvalue problem as shown in Equation (4). The eigenvalue problem can also be formulated as a constrained optimization problem[31,38]:

$$\begin{aligned} &\min_{T_i, b_i} \quad E[(T_i(Y) - E(T_i(Y)) - (X - E[X])^T b_i)^2] \\ &\text{s.t.} \quad \operatorname{var}(T_i(Y)) = 1, \quad \operatorname{cov}(T_i(Y), T_j(Y)) = 0,\ j = 1,\dots,i-1, i = 1,\dots,d, \end{aligned} \qquad (6)$$

where $T_i(y)$ is a set of transformation function of the response variable $y$. The transformation function can be expanded in terms of basis functions:

$$T_i(y) - E(T_i(y)) = \sum_{k=1}^{K} \xi_{ik}\phi_k(y) = \xi_i^T \phi(y), i = 1,\dots,d\ , \qquad (7)$$

where $\phi_k(y)$ are known basis functions, $\xi_{ik}$ are coefficients of expansion,

$\phi(y) = [\phi_1(y),\dots,\phi_K(y)]^T$ and $\xi_i = [\xi_{i1},\dots,\xi_{ik}]^T$. For convenience, we assume that $\phi_1(y) \equiv 1$.

Assume that the predictors are centered. We then denote $x = X - E(X)$. The optimization problem (6) can be written in terms of expansion coefficients:

$$\begin{aligned} &\min_{\xi_i, b} \ \mathrm{E}[(\xi_i^{\mathrm{T}}\phi(y) - x^T b_i)^2] \\ &\text{s.t. } \xi_i^T \operatorname{cov}(\phi(y))\xi_i = 1, \xi_i^T \operatorname{cov}(\phi(y))\xi_j = 0, j = 1,\ldots,i-1, i = 1,\ldots,d. \end{aligned} \tag{8}$$

Assume that $(X_1, Z_1),\ldots,(X_n, Z_n)$ are sampled and that predictors $X_j$ are centered. Let

$Z = [\phi(z_1),\ldots,\phi(z_n)]^T$ and $X = [x_1,\ldots,x_n]^T$ .Then, the sampling formula for the expectation and covariance in problem (8) are given by

$$\mathrm{E}[(\xi_i^{\mathrm{T}}\phi(z) - x^T b_i)^2] \approx \frac{1}{n} \| Z\xi_i - Xb_i \|_2^2 \text{ and } \operatorname{cov}(\phi(z)) \approx \frac{1}{n} Z^T Z .$$

Therefore, the problem (b) can be approximated by

$$\begin{aligned} &\min_{\theta_i \in R^k, b_i \in R^P} \| Z\theta_i - X\beta_i \|^2 \\ &\text{s. t. } \quad \theta_i^T D\theta_i = 1, \theta_i^T D\theta_j = 0, j = 1,\ldots,i-1, i = 1,\ldots,d, \end{aligned} \tag{9}$$

where $D = \frac{Z^T Z}{n}$, $\theta_i = \xi_i$ and $\beta_i = \frac{1}{\sqrt{n}} b_i$.

In other words, the SIR is reformulated as a bi-convex optimization problem, or an optimal scoring problem[31,39].

**Solve Sparse SDR by Alternative Direction Method of Multipliers**

To systematically search the genetic variants of prediction value across the genome based on SDR, penalized techniques should be used to solve the optimal scoring problem (9). Furthermore, in the genome-based disease risk prediction, the number of genetic variants is much larger than the number of sampled individuals. As a consequence, the sample covariance of matrix of genetic variants is singular and its inverse does not exist. Finding solutions to the optimal scoring is problematic. The sparse optimal scoring or SDR algorithms are needed.

Let $B=[\beta_1,...,\beta_i]=[\beta_1^*,...,\beta_p^*]^T, i=1,...,d$ be a $p \times i$ matrix which forms the basis matrix of the CS. For a simplified discussion, $\beta_{ij}, i=1,2,...,d$ is referred to as the $j$-th coordinate in the CS. In the sparse optimal scoring formulation of SDR by Wang and Zhu[31], the penalty is imposed separately for each vector in the CS. Consequently, a coordinate in some vectors in the CS will be penalized toward zero, but the same coordinate in other vectors in the CS may not be penalized to zero. Therefore, it is difficult to use their sparse optimal scoring formulation of SDR for variable screening. To overcome this limitation and develop a sparse SDR that can simultaneously reduce the dimension and the number of predictors, we introduce a coordinated-independent penalty function. We introduce a coordinate-independent penalty function to penalize the coordinate in all reduction directions (vectors forming the CS) toward zero[32]:

$$\rho(B)=\sum_{l=1}^{p}\lambda_l\sqrt{\beta_l^{*T}\beta_l^*}=\sum_{l=1}^{p}\lambda_l \|\beta_l^*\|_2 .$$

For simplicity of computation, we define $\lambda_l=\lambda\|\beta_l^*\|_2^{-r}$, $r \ge 0$ is a pre-specified parameter. Thus, the penalty function can be simplified to

$$\rho(B)=\lambda\sum_{l=1}^{p}\|\beta_l^*\|_2^{1-r} \tag{10}$$

where $\lambda$ is a penalty parameter.

After introducing the penalty function, the sparse version of optimal scoring problem (9) for penalizing the variable can be defined as

$$\min_{\theta_i \in R^k, \beta_i \in R^P} \| Z\theta_i - X\beta_i \|_2^2 + \lambda \sum_{l=1}^{p} \| \beta_l^* \|_2^{1-r}$$
$$\text{s. t.} \quad \theta_i^T D\theta_i = 1, \theta_i^T D\theta_j = 0, j < i, i = 1,\ldots,d \tag{11}$$

The problem (11) is a bi-convex problem. It is convex in $\theta$ for each $\beta$ and convex in $\beta$ for each $\theta$. It can be solved by a simple iterative algorithm. The iterative process consists of two steps: (1) for fixed $\theta_i$ we optimize with respect to $\beta_i$ and for fixed $\beta_i$ we optimize with respect to $\theta_i$. The algorithms are given bellow.

Step 1: Initialization.

Let $D = \frac{Z^T Z}{n}$ and $Q_1 = [1,1,\ldots,1]^T$. We first initialize for $\theta_i^{(0)}, i = 1,\ldots,d$

$$\tilde{\theta}_i^{(0)} = (I - Q_i Q_i^T D)\theta_*,\ \theta_i^{(0)} = \frac{\tilde{\theta}_i^{(0)}}{\sqrt{\tilde{\theta}_i^{T(0)} D \tilde{\theta}_i^{(0)}}},\ Q_{i+1} = [Q_i : \theta_i],$$

where $\theta_*$ is a random $k$ - vector.

Step 2: Iterate between $\theta^{(s)}$ and $\beta^{(s)}$ until convergence or until a specified maximum number of iterations (s=1, 2,…) is reached:

Step A: For fixed $\theta_i^{(s-1)}, i = 1,\ldots,d,$ we solve the following minimization problem:

$$\min_{\beta_i^{(s)} \in R^P} \sum_{\mathrm{i=1}}^{\mathrm{d}} \| \mathrm{Z}\theta_i^{(s-1)} - X\beta_i^{(s)} \|_2^2 + \lambda \left( (1-\delta) \sum_{j=1}^{p} \| \beta_j^{*(s)} \|_2^2 + \delta \sum_{j=1}^{p} \| \beta_j^{*(s)} \|_2^{1-r} \right) \tag{12a}$$

where $B^{(s)} = [\beta_1^{(s)},\ldots,\beta_d^{(s)}] = [\beta_1^{*(s)},\ldots,\beta_p^{*(s)}]^T$.

Step B: For fixed $\beta_i^{(s)},\ i = 1,\ldots,d,$ we seek $\theta_i^{(s)},\ i = 1,\ldots,d,$ which solve the following unconstrained optimization problem:

$$\min_{\theta_i^{(s)} \in R^k} \| Z\theta_i^{(s)} - X\beta_i^{(s)} \|_2^2$$
$$\text{s. t.} \quad \theta_i^{(s)T} D\theta_i^{(s)} = 1, \theta_i^{(s)T} D\theta_j^{(s)} = 0, \ j = 1,\ldots, i-1. \tag{12b}$$

Solution to the above optimization leads to a nonlinear equation (Supplemental Note C):

$$\theta_i^{(s)} = \frac{(I - Q_i^{(s)} Q_i^{(s)T} D) D^{-1} Z^T X \beta_i^{(s)}}{\theta_i^{T(s)} Z^T X \beta_i^{(s)}}, \tag{13}$$

where $Q_1^{(0)} = Q_1^{(1)} = \ldots = Q_1^{(s)} = [1,1,\ldots,1]^T$.

By Newton's method or simple iteration, we obtain a solution $\tilde{\theta}_i^{(s)}$ to Equation 12b. Set

$$\theta_i^{(s)} = \frac{\tilde{\theta}_i^{(s)}}{\sqrt{\tilde{\theta}_i^{T(s)} D \tilde{\theta}_i^{(s)}}}, \qquad Q_{i+1} = [Q_i : \theta_i], i = 1,\ldots, d.$$

If $\| \theta_i^{(s+1)} - \theta_i^{(s)} \|_2 < \varepsilon, \ \|| \beta_i^{(s+1)} - \beta_i^{(s)} \|_2 < \varepsilon, \ i = 1,\ldots, d,$ then stop; otherwise, $s := s+1;$ go to step A.

Now we study how to use ADMM to solve the optimization problem (12a). The algorithm for ADMM to solve optimization problem (12a) is given below (See Supplemental Note C):

Initial value ($m = 0$):

$$\begin{aligned} \beta_j^{(s)(0)} &= (X^T X + \frac{\rho}{2} I)^{-1} X^T Z \theta_j^{(s)} \\ \alpha_j^{(s)(0)} &= \beta_k^{(s)(0)} \\ u_j^{(s)(0)} &= 0, \ j = 1,\ldots, i. \end{aligned} \tag{13}$$

For fixed $\theta_j^{(s)}, j = 1,\ldots, i,$ iterate with $m$ until convergence:

Step (i): $m := m+1.$

Step (ii): $\beta_j^{(s)(m+1)} = (X^T X + \frac{\rho}{2} I)^{-1} [X^T Z \theta_j^{(s)} + \frac{\rho}{2} (\alpha_j^{(m)} - u_j^{(m)})], \ j = 1,\ldots, d$ . (14a)

Step (iii): Let

$B = [\beta_1^{(s)(m+1)}, ..., \beta_i^{(s)(m+1)}] = [\beta_1^{*(s)}, \cdots, \beta_p^{*(s)}]^T$, $\alpha^{(s)(m)} = [\alpha_1^{(s)(m)}, ..., \alpha_i^{(s)(m)}] = [\alpha_1^{*(s)(m)}, \cdots, \alpha_p^{*(s)(m)}]^T$ and $u^{(s)(m)} = [u_1^{(s)(m)}, ..., u_i^{(s)(m)}] = [u_1^{*(s)(m)}, \cdots, u_p^{*(s)(m)}]^T$.

Then,

$$\alpha_i^{*(s)(m+1)} = (\beta_i^{*(s)(m+1)} + u_i^{*(s)(m)}) \left( \frac{\left( \frac{\| \beta_i^{*(s)(m+1)} + u_i^{*(s)(m)} \|_2^{(1+r)} - \frac{\lambda\delta(1-r^2)}{\rho}}{1 + \frac{2\lambda(1-\delta)(1+r)}{\rho}} \right)^{\frac{1}{1+r}}}{\| \beta_i^{*(s)(m+1)} + u_i^{*(s)(m)} \|_2} \right), \ l = 1, ..., p .$$

and

$$\alpha^{(s)(m+1)} = [\alpha_1^{*(s)(m+1)}, \cdots, \alpha_p^{*(s)(m+1)}]^T = [\alpha_1^{(s)(m+1)}, ..., \alpha_d^{(s)(m+1)}]. \quad (14b)$$

Step (iv): $u_j^{(s)(m+1)} = u_j^{(s)(m)} + \beta_j^{(s)(m+1)} - \alpha_j^{(s)(m+1)}$. (14c)

When $\| \beta_j^{(s)(m+1)} - \beta_j^{(s)(m)} \|_2 \le \varepsilon, \| \alpha_j^{(s)(m+1)} - \alpha_j^{(s)(m)} \|_2 \le \varepsilon, \| u_j^{(s)(m+1)} - u_j^{(s)(m)} \|_2 \le \varepsilon, j = 1, ..., i$, stop, go to step B; otherwise go to step (i).

Since the number of SNPs selected for risk prediction by the sparse SDR method is usually less than 50,000 SNPs, a split-and-conquer algorithm is used to search clinically valuable SNPs for disease risk prediction. Briefly, we first divide the whole genome into $K$ sub-genomic regions. The sparse SDR method is then applied to each sub-genomic region to search SNPs with some optimal criterion in the training dataset. Then, we collect all selected SNPs in each sub-genomic region to generate a new dataset for final classification. The sparse SDR method is again applied to the new generated dataset to search SNPs and predict disease. In the previous section, we proved that the proposed split-and-conquer algorithm can reach global optimal classification or disease risk prediction.

## Results

## Application to Real Data Examples

### Prediction of Coronary Artery Disease (CAD) Using the GWAS Dataset

The first clinical use of genetic variants is disease risk prediction that can discriminate between individuals who will develop the disease of interest and those who will not. To examine whether it can systematically search clinically valuable genetic variants for disease prediction, the proposed sparse SDR method was first applied to the GWAS data of the Wellcome Trust Case Control Consortium (WTCCC) for coronary artery disease (CAD) study where 1,929 cases and 2,938 controls were sampled and the total number of SNP markers is 393,473[40]. To reduce bias in the genotypes, we removed all SNPs that were excluded in the original WTCCC CAD study[40]. To unbiased evaluate the performance of the sparse SDR method for disease risk prediction, a 10 fold cross validation (CV) was used to calculate average sensitivity, specificity and classification accuracy. Specifically, the original sample is randomly partitioned into 10 equal size subsamples. Of the 10 subsamples, a single subsample is retained as the test dataset, and the remaining 9 subsamples are used as a training dataset. The cross-validation process is then repeated 10 times (the folds), with each of the 10 subsamples used exactly once as the test dataset. The 10 results from the folds can then be averaged to produce a single estimation of sensitivity, specificity and classification accuracy.

The whole genome was portioned into 20 sub-genomic regions, each with 20 genomic regions which included 19,674 SNPs. To evaluate its performance for disease risk prediction without bias, the sparse SDR was applied to each sub-genomic region in the training dataset. After convergence of the iteration process in the optimal scoring algorithm, the top 2,000 SNPs with the largest contribution to the classification (the largest absolution values of $\beta$ in Equation (12a) were selected for first variable screening. The results from each of the 20 sub-genomic regions were then combined. We totally selected 40,000 SNPs that were used for the second variable screening. We then split the 40,000 SNPs into 4 equal numbers of SNPs sub-datasets. Again, the sparse SDR was applied to each of the 8 sub-datasets. The top 1,500 SNPs with the largest contribution to the classification were then selected for the third (final) variable screening and classification accuracy calculation. Variable screening was performed by the sparse SDR method for each of the ten sets of training datasets. The finalselected SNPs were then used to learn a predictive discriminative model on training individuals which was in turn used to classify CAD on the test dataset.

The final classification performance results of the sparse SDR method was compared with the software implementing sparse logistic regression[41] and the SNP ranking method based on the GWAS P-values of the $\chi^2$ association test in each training dataset where top associated SNPs were selected as input to k-nearest neighbor classifiers[42] such that the selected SNPs can reach the highest classification accuracy in each training dataset. The results were summarized in Tables 1-3 where CV1-CV10 represent ten partitioned datasets in 10 fold cross validations. Several remarkable features emerge from these results. The first, we observed that all sensitivity, specificity and accuracy of classifying CAD using the proposed sparse SDR method in the test dataset were much higher than that from using the sparse logistic regression and SNP ranking method. Using the sparse SDR method compared to the sparse logistic regression and SNP ranking significantly increased classification accuracy at least by ~ 17%. The second, Table 4 listed finally selected SNPs for disease prediction which were shared among at least 2 training folds. Notably, most selected SNPs were in linkage equilibrium or in weak linkage disequilibrium (LD) (data not shown). The P-values of the selected SNPs for testing their association with CAD range from $8.10 \times 10^{-26}$ to 0.947. We observed that not all selected SNPs were significantly associated with CAD. Among 39 selected SNPs in Table 4, there were 20 SNPs with P-values > 0.1. The large P-values of selected best SNPs for CAD classification indicated that they were not associated with CAD by the $\chi^2$ test. This showed that the SNP ranking method based on P-values may not be the best strategy for SNP screening.

**Prediction of Rheumatoid Arthritis (RA) Using the GWAS Dataset**

To further evaluate its disease risk prediction, the sparse SDR method was applied to RA data from GWAS of the North American Rheumatoid Arthritis Consortium (NARAC) where 545,080 SNPs were typed for 866 RA patients and 1,194 controls[43,44]. A total of 490,613 markers from chromosomes 1 to 22, were selected in this study. The genotype data were imputed using the package of BEAGLE-3.3.2[45]. Since the number of sampled individuals in the dataset was not large, the 5 fold cross validation procedure was used to evaluate the classification accuracy.

Again, the split-and-conquer strategy was used for sparse SDR and sparse logistic regression. The whole genome was portioned into 25 sub-genomic regions, each of 20 genomic regions included 19,625 SNPs. At the first step, the sparse SDR was applied to each sub-genomic region in the training dataset. After convergence of the iteration process in the optimal scoring algorithm, the top 2,000 SNPs with the largest contribution to the classification were selected for each sub-genomic region, which led to a total of 50,000 SNP across the whole genome being merged into a dataset for further classification. At the second step, 50,000 SNPs were divided into 4 groups, each of which contained 12,500 SNPs. Again applying the feature selection and classification algorithms to each of the 4 groups, we selected the top 1,500 SNPs from each of the 4 groups which were then merged into one final analysis group. At the third step, a total of 6,000 SNPs in the final analysis group were used for variable screening and classification. The results were summarized in Tables 5-7. The pattern of RA classification using three methods was similar to that of CAD. The sparse SDR had the highest classification accuracy among the three methods. The sensitivity and specificity of classification in both training and test datasets were balanced. We observed that the classification accuracy in the test dataset was similar to that in the training dataset occurred. We also observed that no significant differences in sensitivity, specificity and classification accuracy among 5 folds in both training and test datasets. This indicates that the sparse SDR and its selected SNPs have an excellent generalization property.

**Prediction of Early-onset Myocardial Infarction (EOMI)**

To further evaluate their performance, the proposed method for sparse sufficient dimension reduction was applied to the early-onset myocardial infarction (EOMI) exome sequence data from the NHLBI's Exome Sequencing Project (ESP) (that can be downloaded from dbGaP) where a total of 905 (467 cases and 438 controls) were exome sequenced. Since sparse logistic regression failed to analyze more than 2,000 SNPs, the split and conquer strategies for the sparse SDR and sparse logistic regression were different. For the sparse SDR algorithm, a total of 143,331 SNPs that were included in the analysis was divided into 7 subregions, each subregion having 20,478 SNPs. First, the sparse SDR was applied to each sub-genomic region in the training dataset. From each subregion we selected 800 top SNPs with the largest $\beta$ coefficients. Then, we merged the selected 5,600 SNPs from seven subregions into a final analysis dataset. For the sparse logistic regression, the whole SNP dataset was divided into 200 subregions. From each subregion, we used sparse logistic regression to select the SNPs with non-zero coefficients. A total of 5,519 selected SNPs were further divided into three subregions. We used sparse logistic regression to screen the SNPs in each subregion. The final selected 184 SNPs from three subregions were used as a final analysis group. The results were summarized in Tables 9-11. Again, on average, the sparse SDR had the highest classification accuracy in both the training and test dataset. The total number of selected SNPs in five folds were 981 SNPs that included 117 rare variants and 864 common variants.

The area under the operating characteristic curve (AUC) which measures the ability of an algorithm to discriminate between individuals with disease and control is a widely used performance index for classification. For the convenience of comparisons with other methods, we presented Table 12 and Supplementary Tables S2-S4. Table 12 strongly demonstrated that the AUC values of using genetic information and sparse SDR to classify CAD and RA can reach as high as 0.852 and 0.830, respectively. These AUC values are higher or at least close to the AUC of Framingham Risk Score for coronary heart diseases ($AUC \approx 0.8$) which is considered clinically useful[46]. Even if we used only genetic variants from Exome sequencing the AUC can be as high as 0.714.

## Discussion

The rapid advances in sequencing technologies are producing unprecedentedly massive and highly-dimensional genomic data. There have been bitter debates on whether the genomic variant information has the potential for disease risk prediction. The current paradigm for genomic risk prediction is to evaluate the potential classification accuracy of the genetic variants that are ranked by P-values in the association tests through GWAS. These overly simplified search algorithms failed to identify a set of genetic variants with high discriminatory ability and develop efficient risk prediction models. A fundamental question is how to efficiently extract genomic variants of clinical utility and to develop novel unified approaches for classification analysis of genomic and clinical data. To address these central themes and critical barriers in genomic risk prediction, we shift the paradigm of feature selection from P-value and risk score ranking to optimal genome-wide searching. To systematically search the variants of clinical utility we have addressed several critical issues in this paper.

The first issue we addressed is data reduction. The popular approach to data reduction is unsupervised data reduction. However, unsupervised dimension reduction in disease risk prediction will lose individual disease status information. We propose a supervised dimension reduction method in which both genetic variant information and disease status information will be used. The proposed method for supervised dimension reduction is SDR which replaces the high dimensional original predictor vector with its projection onto a low dimensional space of the predictor space, a central subspace (CS) (minimal set of linear combinations of the original predictors), and without loss of information on phenotypes.

The second issue we addressed is feature selection. The most popular traditional paradigm for genetic variant selection is by P-value or risk score ranking. The association tests are used to calculate the P-value for each variant. The variant selection is by P-value ranking. The top variants with the smallest P-value are selected for risk prediction. However, assessing the clinical utility of the genetic variants by identifying disease association has limitations. The significantly associated genetic variants may not have the high discriminatory power to substantially improve the accuracy of risk prediction. To overcome the limitation of overly simplified P-value or score-based variant selection algorithms, we developed a sparse SDR algorithm for variant selection in which we formulated the variant selection problem into a penalized optimal score problem, a convex optimization problem. Solving the convex optimization problem through genome-wide searches we select a set of genetic variants with the highest discriminatory power. Using three real data examples we demonstrated that the set of selected SNPs by the sparse SDR algorithm had a higher level of classification accuracy than that by P-value ranking algorithms. Some researchers claim that " a risk factor must have a much stronger association with the disease outcome than we ordinarily see in etiologic research if it is to provide a basis for early diagnosis or prediction in individual patients"[47,48]. This statement may not be true. Our three real data examples showed that the selected SNPs for disease risk prediction may be significantly associated with the diseases or may not be associated with the diseases (Tables 4 and 8). Some selected SNPs were not associated with the disease at all. Using only significantly associated SNPs as predictors may increase sensitivity, but will reduce specificity. To balance the sensitivity and specificity, in addition to using strongly associated SNPs as predictors we also need to include weakly associated SNPs as predictors. The assumptions and arguments in the paper[47] seem biased. The sparse SDR algorithm provides a search engine to select a set of SNPs with high discrimination power. Selecting SNPs by simply P-value ranking is not good strategy.

The third issue we addressed is that the current sparse SDR requires the inverse of the sample covariance matrix of the predictors which is often near singular. The singularity problem of the sample covariance matrix is more severe in the presence of rare variants for NGS data, by formulating the sparse SDR problem as an optimal scoring problem that does not require the inverse of the covariance matrix of the variant predictors.

The fourth issue we addressed is the computational complexity of the algorithms raised by large genomic data analysis. NGS may produce million or even dozens of millions of genetic variants. Selection of variants of clinical utility from huge numbers of features poses a substantial computational challenge. If the number of genetic variants exceeds ten 10,000, the popular coordinate descent algorithms may fail to converge to optimal solutions. We used the ADMM algorithm to combine the merits of dual decomposition and augmented Lagrangian methods for constrained optimization and sparse SDR. The distributed ADMM algorithm allows for searching optimal genetic variants for classification through more than 20,000 variants. The ADMM coupled with the split-and-conquer strategy can search millions of features to achieve accurate classification.

The fifth issue we addressed is the generalization power of the selected genetic variants and their ability to balance the sensitivity and specificity of classification. Unlike sparse logistic regression and the P-value rank method, our real data analysis strongly demonstrated that the sparse SDR can balance the sensitivity and specificity of the classification and achieve high generalization power.

Many investigators question the clinical value of genetic variation information on disease risk prediction. The common consensus is that any methods that are of clinical utility should reach a level of prediction with AUC value as high as 0.8. Question is whether using genomic variation information alone we can reach AUC values higher than 0.8. Few of the previous genomic risk prediction studies can achieve AUC values close to 0.8. However, our analysis of GWAS data of CAD and RA strongly demonstrates that using sparse SDR algorithm we can select sets of SNPs which can reach AUC values higher than 0.8. A key issue is how to develop algorithms to systematically search SNPs that are of clinical utility. In this paper by developing sparse SDR search algorithms and their application to CAD and RA genetic data we showed that genomic information has great potential of clinically use.

Next-generation sequencing technologies will identify ten millions of genetic variants across the human genome. The rich genetic variation information provides powerful resources for disease risk prediction. However, more and more researches realize that genomic risk prediction is much difficult than the association analysis. In association tests we analyze a SNP, or a gene or a pathway at a time. However, genomic risk prediction we need to simultaneously search millions or even tens of millions of genetic variants. This poses a great challenge to us. The proposed sparse SDR method and split and conquer strategy only partially solve the extremely big genomic risk prediction problem. The results in this paper are preliminary. The purpose of this paper is to stimulate further discussions about how to develop optimal strategies and algorithms for genomic risk prediction.

**References**

1. Roberts, N.J., Vogelstein, J.T., Parmigiani, G., Kinzler, K.W., Vogelstein, B., and Velculescu, V.E. (2012). The predictive capacity of personal genome sequencing. Science translational medicine 4, 133ra158.

2. Begg, C.B., and Pike, M.C. (2012). Comment on "the predictive capacity of personal genome sequencing". Science translational medicine 4, 135le133; author reply 135lr133.

3. Golan, D., and Rosset, S. (2012). Comment on "the predictive capacity of personal genome sequencing". Science translational medicine 4, 135le134; author reply 135lr133.

4. Topol, E.J. (2012). Comment on "the predictive capacity of personal genome sequencing". Science translational medicine 4, 135le135; author reply 135lr133.

5. Manor, O., and Segal, E. (2013). Predicting Disease Risk Using Bootstrap Ranking and Classification Algorithms PLoS Computational Biology 9.

6. Wendelsdorf, K. (2013). Gene testing revolution: Disease prediction results skyrocket for whole genome and whole exome sequencing.

7. Fugger, L., McVean, G., and Bell, J.I. (2012). Genomewide association studies and common disease--realizing clinical utility. N Engl J Med 367, 2370-2371.

8. Evans, D.M., Visscher, P.M., and Wray, N.R. (2009). Harnessing the information contained within genome-wide association studies to improve individual prediction of complex disease risk. Hum Mol Genet 18, 3525-3531.

9. Kooperberg, C., LeBlanc, M., and Obenchain, V. (2010). Risk prediction using genome-wide association studies. Genetic epidemiology 34, 643-652.

10. Kraft, P., and Hunter, D.J. (2009). Genetic risk prediction--are we there yet? N Engl J Med 360, 1701-1703.

11. Janssens, A.C., and van Duijn, C.M. (2008). Genome-based prediction of common diseases: advances and prospects. Hum Mol Genet 17, R166-173.

12. Kullo, I.J., and Cooper, L.T. (2010). Early identification of cardiovascular risk using genomics and proteomics. Nat Rev Cardiol 7, 309-317.

13. Wray, N.R., Goddard, M.E., and Visscher, P.M. (2008). Prediction of individual genetic risk of complex disease. Curr Opin Genet Dev 18, 257-263.

14. Wei, Z., Wang, K., Qu, H.Q., Zhang, H., Bradfield, J., Kim, C., Frackleton, E., Hou, C., Glessner, J.T., Chiavacci, R., et al. (2009). From disease association to risk assessment: an optimistic view from genome-wide association studies on type 1 diabetes. PLoS Genet 5, e1000678.

15. Frank, I.E., and Friedman, J.H. (1993). A Statistical View of Some Chemometrics Regression Tools. Technometrics 35, 109-135.

16. Tibshirani, R. (1996). Regression Shrinkage and Selection via the Lasso. Journal of the Royal Statistical Society Series B (Methodological) 58, 267-288.

17. Fan, J., and Li, R. (2001). Variable selection via nonconcave penalized likelihood and its oracle properties. Journal of the American Statistical Association 96, 1348-1360.

18. E, C., and T, T. (2007). The Dantzig selector: statistical estimation when p is much larger than n. Ann Stat 35, 2313.

19. H, Z., and R, L. (2008). One-step sparse estimates in nonconcave penalized likelihood models. Ann Stat 36, 1509.

20. Austin, E., Pan, W., and Shen, X. (2013). Penalized Regression and Risk Prediction in Genome-Wide Association Studies. Statistical analysis and data mining 6.

21. Fukumizu, K., Bach, F.R., and Jordan, M.I. (2004). Dimensionality reduction for supervised learning with reproducing kernel Hilbert spaces. The Journal of Machine Learning Research 5, 73-99.

22. Cook, R.D. (2004). Testing predictor contributions in sufficient dimension reduction. The Annals of Statistics 32, 1062-1092.

23. Fukumizu, K., Bach, F.R., and Jordan, M.I. (2006). Kernel dimension reduction in regression.

24. Cook, R.D. (1994). On the Interpretation of Regression Plots. Journal of the American Statistical Association 89, 177-189.

25. Cook, R.D. (1998). Principal Hessian Directions Revisited. Journal of the American Statistical Association 93, 84-94.

26. Li, K.-C. (1991). Sliced inverse regression for dimension reduction. Journal of the American Statistical Association 86, 316-327.

27. Cook, R.D., and Weisberg, S. (1991). Sliced Inverse Regression for Dimension Reduction - Comment. Journal of the American Statistical Association 86, 328-332.

28. Li, K.-C. (1992). On Principal Hessian Directions for Data Visualization and Dimension Reduction: Another Application of Stein's Lemma. Journal of the American Statistical Association 87, 1025-1039.

29. Li, B., and Wang, S. (2007). On Directional Regression for Dimension Reduction. Journal of the American Statistical Association 102, 997-1008.

30. Cook, R.D. (2007). Fisher lecture: Dimension reduction in regression. Statistical Science 22, 1-26.

31. Wang, T., and Zhu, L. (2013). Sparse sufficient dimension reduction using optimal scoring. Computational Statistics & Data Analysis.

32. Chen, X., Zou, C., and Cook, R.D. (2010). Coordinate-independent sparse sufficient dimension reduction and variable selection. The Annals of Statistics, 3696-3723.

33. Li, L., Dennis Cook, R., and Nachtsheim, C.J. (2005). Model free variable selection. Journal of the Royal Statistical Society: Series B (Statistical Methodology) 67, 285-299.

34. Ni, L., Cook, R.D., and Tsai, C.-L. (2005). A note on shrinkage sliced inverse regression. Biometrika 92, 242-247.

35. Li, L. (2007). Sparse sufficient dimension reduction. Biometrika 94, 603-613.

36. Zhou, J., and He, X. (2008). Dimension reduction based on constrained canonical correlation and variable filtering. The Annals of Statistics 36, 1649-1668.

37. Nilsson, J., Sha, F., and Jordan, M.I. (2007). Regression on manifolds using kernel dimension reduction. In Proceedings of the 24th international conference on Machine learning. (ACM), pp 697-704.

38. Chen, C.-H., and Li, K.-C. (1998). Can SIR be as popular as multiple linear regression? Statistica Sinica 8, 289-316.

39. Clemmensen, L., Hastie, T., Witten, D., and Ersbøll, B. (2011). Sparse discriminant analysis. Technometrics 53.

40. Frayling, T.M., Timpson, N.J., Weedon, M.N., Zeggini, E., Freathy, R.M., Lindgren, C.M., Perry, J.R., Elliott, K.S., Lango, H., and Rayner, N.W. (2007). A common variant in the FTO gene is associated with body mass index and predisposes to childhood and adult obesity. Science 316, 889-894.

41. Friedman, J., Hastie, T., and Tibshirani, R. (2010). Regularization Paths for Generalized Linear Models via Coordinate Descent. J Stat Softw 33, 1-22.

42. Venables, W.N., and Ripley, B.D. (2002). Modern Applied Statistics with S. Fourth Edition.

43. Plenge, R.M., Seielstad, M., Padyukov, L., Lee, A.T., Remmers, E.F., Ding, B., Liew, A., Khalili, H., Chandrasekaran, A., Davies, L.R., et al. (2007). TRAF1-C5 as a risk locus for rheumatoid arthritis--a genomewide study. N Engl J Med 357, 1199-1209.

44. Luo, L., Peng, G., Zhu, Y., Dong, H., Amos, C.I., and Xiong, M. (2010). Genome-wide gene and pathway analysis. European Journal of Human Genetics 18, 1045-1053.

45. Browning, B.L., and Browning, S.R. (2011). A fast, powerful method for detecting identity by descent. The American Journal of Human Genetics 88, 173-182.

46. De Jager, P.L., Chibnik, L.B., Cui, J., Reischl, J., Lehr, S., Simon, K.C., Aubin, C., Bauer, D., Heubach, J.F., Sandbrink, R., et al. (2009). Integration of genetic risk factors into a clinical algorithm for multiple sclerosis susceptibility: a weighted genetic risk score. The Lancet Neurology 8, 1111-1119.

47. Ware, J.H. (2006). The limitations of risk factors as prognostic tools. N Engl J Med 355, 2615-2617.

48. Ripatti, S., Tikkanen, E., Orho-Melander, M., Havulinna, A.S., Silander, K., Sharma, A., Guiducci, C., Perola, M., Jula, A., Sinisalo, J., et al. (2010). A multilocus genetic risk score for coronary heart disease: case-control and prospective cohort analyses. Lancet 376, 1393-1400.

**Table 1**. Accuracy of the CAD classifications using the sparse SDR method.

| Sample Folder | Training Dataset | | | Test Dataset | | | #Selected SNPs |
|---|---|---|---|---|---|---|---|
| | Sensitivity | Specificity | Accuracy | Sensitivity | Specificity | Accuracy | |
| CV – 1 | 0.6332 | 0.9131 | 0.8024 | 0.6502 | 0.9043 | 0.8024 | 30 |
| CV – 2 | 0.4752 | 0.9644 | 0.7702 | 0.4842 | 0.9502 | 0.7699 | 13 |
| CV – 3 | 0.6450 | 0.9198 | 0.8103 | 0.6387 | 0.9122 | 0.8098 | 27 |
| CV – 4 | 0.6628 | 0.9074 | 0.8117 | 0.6893 | 0.9038 | 0.8090 | 71 |
| CV – 5 | 0.6590 | 0.9149 | 0.8133 | 0.6837 | 0.8958 | 0.8131 | 62 |
| CV – 6 | 0.6158 | 0.9238 | 0.8020 | 0.6298 | 0.9167 | 0.8019 | 38 |
| CV – 7 | 0.6600 | 0.9129 | 0.8121 | 0.6136 | 0.9226 | 0.8076 | 36 |
| CV – 8 | 0.6592 | 0.9055 | 0.8078 | 0.6724 | 0.8934 | 0.8072 | 31 |
| CV – 9 | 0.6378 | 0.9024 | 0.7990 | 0.6452 | 0.9194 | 0.7980 | 30 |
| CV – 10 | 0.6616 | 0.8990 | 0.8041 | 0.6485 | 0.8912 | 0.8039 | 95 |
| Average | 0.6310 | 0.9163 | 0.8033 | 0.6356 | 0.9109 | 0.8023 | |

**Table 2**. Accuracy of the CAD classifications using the sparse logistic regression.

| Sample Folder | Training Dataset | | | Test Dataset | | | #Selected SNPs |
|---|---|---|---|---|---|---|---|
| | Sensitivity | Specificity | Accuracy | Sensitivity | Specificity | Accuracy | |
| CV – 1 | 0.6873 | 0.7192 | 0.7102 | 0.5726 | 0.6545 | 0.6344 | 34 |
| CV – 2 | 0.6734 | 0.7121 | 0.7011 | 0.5694 | 0.6888 | 0.6538 | 36 |
| CV – 3 | 0.6682 | 0.7151 | 0.7012 | 0.5369 | 0.6925 | 0.6471 | 38 |
| CV – 4 | 0.6841 | 0.7172 | 0.7083 | 0.6691 | 0.6515 | 0.6567 | 40 |
| CV – 5 | 0.6741 | 0.7045 | 0.6964 | 0.6103 | 0.6921 | 0.6700 | 29 |
| CV – 6 | 0.6796 | 0.7116 | 0.7028 | 0.6129 | 0.6904 | 0.6673 | 32 |
| CV – 7 | 0.6809 | 0.7096 | 0.7017 | 0.5868 | 0.7017 | 0.6723 | 31 |
| CV – 8 | 0.6809 | 0.7142 | 0.7048 | 0.6436 | 0.6841 | 0.6749 | 36 |
| CV – 9 | 0.6833 | 0.7227 | 0.7117 | 0.6398 | 0.6535 | 0.6490 | 44 |
| CV – 10 | 0.6742 | 0.7077 | 0.6983 | 0.6316 | 0.7304 | 0.7059 | 38 |
| Average | 0.6786 | 0.7134 | 0.7036 | 0.6073 | 0.6839 | 0.6631 | |

**Table 3**. Accuracy of the CAD classifications using SNP ranking

| Sample Folder | Training Dataset | | | Test Dataset | | | #Selected SNPs |
|---|---|---|---|---|---|---|---|
| | Sensitivity | Specificity | Accuracy | Sensitivity | Specificity | Accuracy | |
| CV – 1 | 0.4109 | 0.8657 | 0.6859 | 0.3842 | 0.8845 | 0.6838 | 30 |
| CV – 2 | 0.4038 | 0.8695 | 0.6847 | 0.3579 | 0.8771 | 0.6762 | 30 |
| CV – 3 | 0.4075 | 0.8709 | 0.6863 | 0.3927 | 0.8683 | 0.6902 | 30 |
| CV – 4 | 0.4070 | 0.8753 | 0.6921 | 0.4029 | 0.8731 | 0.6652 | 30 |
| CV – 5 | 0.4017 | 0.8715 | 0.6852 | 0.3827 | 0.8632 | 0.6759 | 30 |
| CV – 6 | 0.4115 | 0.8709 | 0.6892 | 0.4327 | 0.8237 | 0.6673 | 30 |
| CV – 7 | 0.4086 | 0.8747 | 0.6889 | 0.3693 | 0.8081 | 0.6448 | 30 |
| CV – 8 | 0.4178 | 0.8695 | 0.6904 | 0.3621 | 0.8640 | 0.6682 | 30 |
| CV – 9 | 0.4014 | 0.8732 | 0.6888 | 0.3871 | 0.8498 | 0.6449 | 30 |
| CV – 10 | 0.3992 | 0.8627 | 0.6774 | 0.4000 | 0.8776 | 0.7059 | 30 |
| Average | 0.4069 | 0.8704 | 0.6869 | 0.3872 | 0.8589 | 0.6722 | |

**Table 4**. Number of CV folders where the selected SNPs were shared in the training datasets (CAD).

| Chromosome | SNP | Gene Name | P-value | Number of Shared Folders |
|---|---|---|---|---|
| Chr10 | rs7906587 | PNLIPRP3 | 8.10E-26 | 10 |
| Chr18 | rs4799934 | CELF4 | 3.37E-23 | 10 |
| Chr18 | rs1595963 | | 7.97E-16 | 10 |
| Chr5 | rs2416472 | | 1.06E-10 | 10 |
| Chr22 | rs4819660 | FLJ41941 | 6.13E-07 | 10 |
| Chr22 | rs4819661 | FLJ41941 | 3.45E-01 | 10 |
| Chr5 | rs17411921 | | 3.58E-01 | 10 |
| Chr4 | rs1553460 | | 3.40E-19 | 9 |
| Chr4 | rs890447 | LINC00499 | 1.05E-11 | 9 |
| Chr4 | rs6531531 | | 1.05E-08 | 9 |
| Chr16 | rs11640295 | | 1.88E-08 | 9 |
| Chr16 | rs16955238 | | 7.89E-08 | 9 |
| Chr4 | rs11100890 | LINC00499 | 3.35E-01 | 9 |
| Chr18 | rs1786776 | CELF4 | 7.40E-01 | 9 |
| Chr1 | rs691531 | | 7.11E-05 | 8 |
| Chr18 | rs16947249 | | 2.84E-01 | 8 |
| Chr16 | rs16955255 | | 9.47E-01 | 8 |
| Chr3 | rs326296 | | 1.00E-05 | 7 |
| Chr10 | rs11197703 | | 3.95E-01 | 7 |
| Chr4 | rs1503882 | | 8.85E-01 | 6 |
| Chr4 | rs10022638 | | 8.78E-01 | 5 |
| Chr16 | rs1559394 | PRM2 | 4.34E-01 | 4 |
| Chr4 | rs17523800 | | 4.78E-01 | 4 |
| Chr3 | rs9874467 | | 5.12E-01 | 4 |
| Chr4 | rs6848027 | | 8.70E-01 | 4 |
| Chr16 | rs7187741 | | 9.13E-01 | 4 |
| Chr5 | rs17076079 | | 1.84E-17 | 3 |
| Chr16 | rs16870039 | | 7.01E-06 | 3 |
| Chr3 | rs804980 | | 1.89E-01 | 3 |
| Chr10 | rs10749223 | | 3.40E-01 | 3 |
| Chr4 | rs1039539 | | 5.21E-01 | 3 |
| Chr19 | rs11671119 | MEF2BNB-MEF2B | 1.45E-17 | 2 |
| Chr16 | rs8058964 | CDH13 | 1.36E-06 | 2 |
| Chr3 | rs11924705 | | 3.21E-04 | 2 |
| Chr3 | rs7653441 | FNDC3B | 7.89E-04 | 2 |
| Chr11 | rs12287340 | SPON1 | 2.38E-01 | 2 |
| Chr3 | rs804974 | HGD | 4.48E-01 | 2 |
| Chr4 | rs9884478 | NPFFR2 | 4.99E-01 | 2 |
| Chr3 | rs10934513 | | 5.11E-01 | 2 |

**Table 5**. Accuracy of the RA classifications using the sparse SDR method.

| Sample Folder | Training Dataset | | | Test Dataset | | | #Selected SNPs |
|---|---|---|---|---|---|---|---|
| | Sensitivity | Specificity | Accuracy | Sensitivity | Specificity | Accuracy | |
| CV – 1 | 0.7587 | 0.7906 | 0.7771 | 0.7193 | 0.8112 | 0.7738 | 16 |
| CV – 2 | 0.7752 | 0.7701 | 0.7722 | 0.8046 | 0.7456 | 0.7711 | 12 |
| CV – 3 | 0.7551 | 0.7753 | 0.7668 | 0.7514 | 0.7679 | 0.7610 | 13 |
| CV – 4 | 0.7742 | 0.7722 | 0.7730 | 0.7401 | 0.7975 | 0.7729 | 9 |
| CV – 5 | 0.7655 | 0.7739 | 0.7704 | 0.7683 | 0.7650 | 0.7663 | 8 |
| Average | 0.7657 | 0.7764 | 0.7719 | 0.7567 | 0.7774 | 0.7690 | |

**Table 6**. Accuracy of the RA classifications using the sparse logistic regression.

| Sample Folder | Training Dataset | | | Test Dataset | | | #Selected SNPs |
|---|---|---|---|---|---|---|---|
| | Sensitivity | Specificity | Accuracy | Sensitivity | Specificity | Accuracy | |
| CV – 1 | 0.5363 | 0.8793 | 0.7340 | 0.4971 | 0.8996 | 0.7357 | 5 |
| CV – 2 | 0.6832 | 0.8182 | 0.7619 | 0.7069 | 0.7456 | 0.7289 | 5 |
| CV – 3 | 0.6851 | 0.8122 | 0.7589 | 0.6705 | 0.7975 | 0.7439 | 4 |
| CV – 4 | 0.6261 | 0.8228 | 0.7405 | 0.5989 | 0.8523 | 0.7440 | 7 |
| CV – 5 | 0.6489 | 0.8170 | 0.7461 | 0.6159 | 0.7949 | 0.7211 | 5 |
| Average | 0.6359 | 0.8299 | 0.7483 | 0.6178 | 0.8180 | 0.7347 | |

**Table 7**. Accuracy of the RA classifications using SNP ranking.

| Sample Folder | Training Dataset | | | Test Dataset | | | #Selected SNPs |
|---|---|---|---|---|---|---|---|
| | Sensitivity | Specificity | Accuracy | Sensitivity | Specificity | Accuracy | |
| CV – 1 | 0.8852 | 0.6122 | 0.7278 | 0.8830 | 0.6386 | 0.7381 | 20 |
| CV – 2 | 0.8657 | 0.6635 | 0.7479 | 0.8793 | 0.6272 | 0.7363 | 20 |
| CV – 3 | 0.8571 | 0.6508 | 0.7375 | 0.8497 | 0.6414 | 0.7293 | 20 |
| CV – 4 | 0.8563 | 0.6498 | 0.7362 | 0.8588 | 0.6582 | 0.7440 | 20 |
| CV – 5 | 0.8820 | 0.6372 | 0.7406 | 0.8415 | 0.6667 | 0.7387 | 20 |
| Average | 0.8693 | 0.6427 | 0.7380 | 0.8625 | 0.6464 | 0.7373 | |

**Table 8**. Number of CV folders where the selected SNPs were shared in the training datasets (RA).

| Chromosome | SNP | Gene Name | P-value | Number of Shared Folders |
|---|---|---|---|---|
| Chr6 | rs2395175 | | 7.37E-117 | 5 |
| Chr6 | rs660895 | | 1.92E-112 | 5 |
| Chr6 | rs6457617 | | 1.03E-79 | 5 |
| Chr6 | rs9275555 | | 2.16E-63 | 5 |
| Chr6 | rs6903608 | | 3.17E-48 | 5 |
| Chr6 | rs9275224 | | 8.04E-80 | 4 |
| Chr6 | rs9275595 | | 1.13E-60 | 2 |
| Chr9 | rs2900180 | | 5.28E-09 | 2 |
| Chr14 | rs8017455 | LOC101928431 | 3.08E-03 | 2 |
| Chr6 | rs6910071 | C6orf10 | 4.78E-97 | 1 |
| Chr6 | rs2395163 | | 1.04E-93 | 1 |
| Chr6 | rs2395185 | | 3.50E-68 | 1 |
| Chr6 | rs7745656 | | 1.68E-36 | 1 |
| Chr6 | rs9275601 | | 1.17E-14 | 1 |
| Chr6 | rs10947262 | BTNL2 | 9.45E-14 | 1 |
| Chr20 | rs1182531 | PHACTR3 | 1.70E-07 | 1 |
| Chr5 | rs16883129 | | 5.26E-07 | 1 |
| Chr14 | rs12885166 | RIN3 | 8.76E-07 | 1 |
| Chr6 | rs10948693 | | 1.51E-06 | 1 |
| Chr16 | rs7202398 | TANGO6 | 6.88E-06 | 1 |
| Chr20 | rs2024946 | | 2.14E-05 | 1 |
| Chr10 | rs12246473 | | 2.23E-05 | 1 |
| Chr6 | rs17828521 | TINAG | 1.62E-04 | 1 |
| Chr3 | rs4682062 | PHLDB2 | 3.56E-04 | 1 |
| Chr3 | rs4686162 | | 4.15E-04 | 1 |
| Chr15 | rs8040193 | | 7.74E-04 | 1 |
| Chr3 | rs5015055 | CACNA2D3 | 9.11E-04 | 1 |
| Chr11 | rs12365956 | TMPRSS13 | 1.91E-03 | 1 |
| Chr3 | rs6769555 | SLC9A9 | 2.36E-03 | 1 |
| Chr9 | rs10759014 | PTPRD | 2.80E-03 | 1 |
| Chr8 | rs13275996 | CSMD1 | 3.19E-03 | 1 |
| Chr18 | rs10502447 | | 3.32E-03 | 1 |

**Table 9**. Accuracy of the EOMI classifications using the sparse SDR method.

| Sample Folder | Training Dataset | | | Test Dataset | | | #Selected SNPs |
|---|---|---|---|---|---|---|---|
| | Sensitivity | Specificity | Accuracy | Sensitivity | Specificity | Accuracy | |
| CV – 1 | 0.9888 | 0.9842 | 0.9864 | 0.7654 | 0.7159 | 0.7396 | 275 |
| CV – 2 | 0.9971 | 0.9893 | 0.9930 | 0.6739 | 0.6915 | 0.6828 | 298 |
| CV – 3 | 0.9971 | 0.9919 | 0.9944 | 0.6517 | 0.6837 | 0.6684 | 300 |
| CV – 4 | 0.9769 | 0.9565 | 0.9664 | 0.7065 | 0.6970 | 0.7016 | 193 |
| CV – 5 | 0.9689 | 0.9604 | 0.9645 | 0.6310 | 0.7386 | 0.6860 | 160 |
| Average | 0.9858 | 0.9765 | 0.9810 | 0.6857 | 0.7053 | 0.6957 | |

**Table 10**. Accuracy of the EOMI classifications using the sparse logistic regression.

| Sample Folder | Training Dataset | | | Test Dataset | | | #Selected SNPs |
|---|---|---|---|---|---|---|---|
| | Sensitivity | Specificity | Accuracy | Sensitivity | Specificity | Accuracy | |
| CV – 1 | 0.9720 | 0.9631 | 0.9674 | 0.6914 | 0.6591 | 0.6746 | 168 |
| CV – 2 | 0.9682 | 0.9598 | 0.9638 | 0.6304 | 0.6277 | 0.6290 | 132 |
| CV – 3 | 0.9828 | 0.9458 | 0.9638 | 0.5843 | 0.6531 | 0.6203 | 150 |
| CV – 4 | 0.9855 | 0.9620 | 0.9734 | 0.6413 | 0.6364 | 0.6387 | 155 |
| CV – 5 | 0.9689 | 0.9683 | 0.9686 | 0.6548 | 0.7159 | 0.6860 | 158 |
| Average | 0.9755 | 0.9598 | 0.9674 | 0.6404 | 0.6584 | 0.6497 | |

**Table 11**. Accuracy of the EOMI classifications using SNP ranking.

| Sample Folder | Training Dataset | | | Test Dataset | | | #Selected SNPs |
|---|---|---|---|---|---|---|---|
| | Sensitivity | Specificity | Accuracy | Sensitivity | Specificity | Accuracy | |
| CV – 1 | 0.8850 | 0.8564 | 0.8696 | 0.6575 | 0.6563 | 0.6568 | 100 |
| CV – 2 | 0.9057 | 0.8554 | 0.8776 | 0.6282 | 0.6019 | 0.6129 | 100 |
| CV – 3 | 0.8938 | 0.8417 | 0.8649 | 0.6081 | 0.6106 | 0.6096 | 100 |
| CV – 4 | 0.9091 | 0.8582 | 0.8810 | 0.6790 | 0.6636 | 0.6702 | 100 |
| CV – 5 | 0.8944 | 0.8394 | 0.8636 | 0.7015 | 0.6476 | 0.6686 | 100 |
| Average | 0.8976 | 0.8502 | 0.8713 | 0.6549 | 0.6360 | 0.6436 | |

**Table 12**. Average AUC values of three methods for classifying CAD, RA and EOMI.

| Disease | Training Dataset | | | Test Dataset | | |
|---|---|---|---|---|---|---|
| | Sparse SDR | Sparse Logistic | P-value Rank | Sparse SDR | Sparse Logistic | P-value Rank |
| CAD | 0.8660 | 0.7310 | 0.6776 | 0.8520 | 0.6723 | 0.6595 |
| RA | 0.8560 | 0.8299 | 0.8356 | 0.8300 | 0.8268 | 0.8288 |
| EOMI | 1.0000 | 1.0000 | 0.9492 | 0.7140 | 0.6480 | 0.6846 |

**Supplemental Note A**

**Eigenequation for estimation of a basis of the central space**

To use inversion regression to estimate the CS we make the following assumptions[1]:

(1) Linearity condition: $E(Z \mid P_{S_{Y|Z}} Z) = P_{S_{Y|Z}} Z$. Let a $p \times k$ dimensional matrix $\gamma = [\gamma_1, \ldots, \gamma_k]$ form basis matrix for the CS $S_{Y|Z}$. Then, assumption (1) requires $E(Z \mid \gamma^T Z)$ be a linear function of $\gamma^T Z$.

(2) Coverage condition: Span{ $E(Z \mid Y = y) \mid y = 1, \ldots, h\} = S_{Y|Z}$, where we assume that $Y$ is discretized into $h$ slices. In other words, the space spanned by the inverse conditional mean defines the CS. Therefore, for each $y$ of $Y$, the inverse conditional mean $E(Z \mid Y = y)$ is a linear combination of bases of $S_{Y|X}$:

$$E(Z \mid Y = y) = \gamma \rho_y, \tag{A1}$$

where $\gamma$ is the basis matrix for $S_{Y|Z}$ and $\rho_y \in R^k$ is a $k$ dimensional vector.

(3) Constant covariance condition: Var $(Z \mid P_{S_{Y|Z}} Z) = Q_{S_{Y|Z}}$, where $Q_{S_{Y|Z}} = I_p - P_{S_{Y|Z}}$. In other words, $\mathrm{Var}(Z \mid P_{S_{Y|Z}} Z)$ should be a nonrandom matrix.

To characterize the basis in the CS for $Z$, we define a linear combination of variables in $E(Z \mid Y)$ as $\gamma_0^T E(Z \mid Y)$. To maximally employ information in the CS $S_{Y|Z}$, we maximize the variance of $\gamma_0^T E(Z \mid Y)$:

$$\mathrm{cov}(\gamma_0^T E(Z \mid Y)) = \gamma_0^T \mathrm{cov}(E(Z \mid Y))\gamma_0, \tag{A2a}$$

under the constraints $\gamma_0^T \gamma_0 = 1$ or

$$\begin{aligned}\text{cov}(\gamma_0^T E(Z \mid Y)) &= \gamma_0^T \text{cov}(E(Z \mid Y))\gamma_0 \\ &= (\Sigma^{-1/2}\gamma_0)^T \text{cov}(E(X - E(X) \mid Y))\Sigma^{-1/2}\gamma_0 \\ &= \beta^T \text{cov}(E(X - E(X) \mid Y)\beta,\end{aligned} \qquad \text{(A2b)}$$

under the constraints

$$\gamma_0^T \gamma_0 = 1 \text{ or } \beta^T \Sigma \beta = 1,$$

where $\beta_1,...,\beta_k$ is the basis for CS $S_{Y|X}$. Solving the optimization problem (A2a) or (A2b) leads to the following eigenequation:

$$\text{cov}(E(Z \mid Y))\gamma = \lambda_z \gamma, \qquad \text{(A3a)}$$

or

$$\text{cov}(E(X - E(X) \mid Y))\beta = \lambda_x \Sigma_x \beta, \qquad \text{(A3b)}$$

where $\lambda_z$ and $\lambda_x$ are eigenvalues, and $\gamma$ and $\beta$ is an eigenvector, respectively. Solutions to eigenequation (A3a) and (A3b)yields the basis matrices $\eta = [\gamma_1,...,\gamma_d]$ for $S_{Y|Z}$ and $B = [\beta_1,...,\beta_k]$ for $S_{Y|X}$, respectively

**Supplemental Note B**

**Partition of Global SDR for Whole Genome into a Number of Small SDR for Small**

Suppose that genome is divided into $d$ genomic regions. For the $j$-the genomic region, we assume that some components of the basis vector are zero. The basis vector in the $j$-the genomic region can be denoted by

$$\gamma_j = \begin{bmatrix} \xi_j \\ 0 \end{bmatrix}.$$

From equation (A1), we have

$$E\left\{ \begin{bmatrix} Z_1^j \\ Z_2^j \end{bmatrix} \middle| Y = y \right\} = \begin{bmatrix} \xi_j \\ 0 \end{bmatrix} \rho_y, \quad j = 1,2,...,d. \qquad \text{(B1)}$$

By arranging the order of variables, the vector $Z$ and $\gamma$ can be written as

$$Z = \begin{bmatrix} Z_1^1 \\ \vdots \\ Z_d^1 \\ 0 \end{bmatrix} = \begin{bmatrix} Z_1 \\ Z_2 \end{bmatrix} \text{ and } \gamma = \begin{bmatrix} \xi_1 \\ \vdots \\ \xi_d \\ 0 \end{bmatrix} = \begin{bmatrix} \xi \\ 0 \end{bmatrix} \qquad \text{(B2)}$$

where $Z_1 = \begin{bmatrix} Z_1^1 \\ \vdots \\ Z_d^1 \end{bmatrix}$ and $\xi = \begin{bmatrix} \xi_1 \\ \vdots \\ \xi_d \end{bmatrix}$.

Combining equations (A1) and (B2) we obtain

$$\begin{aligned} E[Z_1 \mid Y = y] &= \xi \rho_y \\ E[Z_2 \mid Y = y] &= 0, \end{aligned} \qquad \text{(B3)}$$

But,

$$\operatorname{cov}(E(Z \mid Y)) = \begin{bmatrix} \operatorname{cov}(E(Z_1 \mid Y)) & 0 \\ 0 & 0 \end{bmatrix}. \quad \text{(B4)}$$

It follows from equations (3) and (B4) that

$$\operatorname{cov}(E(Z_1 \mid Y))\xi = \lambda_z \xi, \quad \text{(B5)}$$

which implies that zero components can be removed by solving eigenequation for each genomic region.

Recall that

$$Z = \Sigma^{-1/2}(X - E(X)).$$

It follows from equation (A1) that

$$E(X - E(X) \mid Y) = \Sigma_x \beta \rho_y . \quad \text{(B6).}$$

For the $j$-the genomic region, we assume that some components of the basis vector for $S_{Y|X}$ are zero. The basis vector in the $j$-the genomic region can be denoted by

$$\beta_j = \begin{bmatrix} B_j \\ 0 \end{bmatrix}.$$

From equation (B6), we have

$$E\left\{ \begin{matrix} X_1^j - E(X_1^j) \\ X_2^j - E(X_2^j) \end{matrix} \middle| Y = y \right\} = \begin{bmatrix} \Sigma_{11}^j & \Sigma_{12}^j \\ \Sigma_{21}^j & \Sigma_{22}^j \end{bmatrix} \begin{bmatrix} B_j \\ 0 \end{bmatrix} \rho_y , \ j = 1,2,...,m.$$

By arranging the order of variables, the vector $X$ and $\beta$ can be written as

$$X = \begin{bmatrix} X_1^1 \\ \vdots \\ X_m^1 \\ X_1^2 \\ \vdots \\ X_m^2 \end{bmatrix} = \begin{bmatrix} X_1 \\ X_2 \end{bmatrix} \text{ and } \beta = \begin{bmatrix} B_1 \\ \vdots \\ B_m \\ 0 \end{bmatrix} = \begin{bmatrix} B \\ 0 \end{bmatrix} \quad \text{(B7)}$$

where $X_1 = \begin{bmatrix} X_1^1 \\ \vdots \\ X_m^1 \end{bmatrix}, X_2 = \begin{bmatrix} X_1^2 \\ \vdots \\ X_m^2 \end{bmatrix}$ and $B = \begin{bmatrix} b_1 \\ \vdots \\ b_m \end{bmatrix}$.

Combining equations (B6) and (B7) we obtain

$$\begin{bmatrix} E(X_1 - E(X_1) \mid Y = y) \\ E(X_2 - E(X_2) \mid Y = y) \end{bmatrix} = \begin{bmatrix} \Sigma_{11} & \Sigma_{12} \\ \Sigma_{21} & \Sigma_{22} \end{bmatrix} \begin{bmatrix} B \\ 0 \end{bmatrix} \rho_y . \quad \text{(B8)}$$

From equation (B8), we can obtain the following covariance matrix $\operatorname{cov}(E(X - E(X) \mid Y))$:

$$\operatorname{cov}(E(X - E(X) \mid Y)) = \begin{bmatrix} \Sigma_{11} B E(\rho_y \rho_y^T) B^T \Sigma_{11} & \Sigma_{11} B E(\rho_y \rho_y^T) B^T \Sigma_{12} \\ \Sigma_{21} B E(\rho_y \rho_y^T) B^T \Sigma_{11} & \Sigma_{21} B E(\rho_y \rho_y^T) B^T \Sigma_{12} \end{bmatrix} . \quad \text{(B9)}$$

Then, we have

$$\operatorname{cov}(E(X_1 - E(X_1) \mid Y) = \Sigma_{11} B E(\rho_y \rho_y^T) B^T \Sigma_{11} .$$

Let $\beta = \begin{bmatrix} e \\ 0 \end{bmatrix}$. The eigenequation (4) for the vector of variable $X_1$ is given by

$$\operatorname{cov}(E(X_1 - E(X_1) \mid Y) e = \Sigma_{11} B E(\rho_y \rho_y^T) B^T \Sigma_{11} e = \lambda_x \Sigma_{11} e, \quad \text{(B10)}$$

which implies that

$$\Sigma_{21} B E(\rho_y \rho_y^T) B^T \Sigma_{11} e = \lambda_x \Sigma_{21} e . \quad \text{(B11)}$$

It follows from equation (B9) that

$$\operatorname{cov}(E(X - E(X)\,|\,Y))\beta = \begin{bmatrix} \Sigma_{11} B E(\rho_y \rho_y^T) B^T \Sigma_{11} & \Sigma_{11} B E(\rho_y \rho_y^T) B^T \Sigma_{12} \\ \Sigma_{21} B E(\rho_y \rho_y^T) B^T \Sigma_{11} & \Sigma_{21} B E(\rho_y \rho_y^T) B^T \Sigma_{12} \end{bmatrix} \begin{bmatrix} e \\ 0 \end{bmatrix} = \begin{bmatrix} \Sigma_{11} B E(\rho_y \rho_y^T) B^T \Sigma_{11} e \\ \Sigma_{21} B E(\rho_y \rho_y^T) B^T \Sigma_{11} e \end{bmatrix} \quad \text{(B12)}$$

Combining equations (B10)-(B12) leads to

$$\operatorname{cov}(E(X - E(X)\,|\,Y))\beta = \lambda_x \begin{bmatrix} \Sigma_{11} e \\ \Sigma_{21} e \end{bmatrix} = \lambda_x \begin{bmatrix} \Sigma_{11} & \Sigma_{12} \\ \Sigma_{21} & \Sigma_{22} \end{bmatrix} \begin{bmatrix} e \\ 0 \end{bmatrix} = \lambda_x \Sigma\beta, \quad \text{(B13)}$$

which implies that solution to the eigenequation for sub-vector of predictors $X_1$ also satisfies the eigenequation (B13) for whole vector of predictors $X$ .

**Supplemental Note C**

**Optimal Scoring and Alternative Direction Methods of Multipliers (ADMM) Algorithms**

We first introduce alternative direction methods of multipliers (ADMM) for solving the constrained convex optimization problem[2]. We consider the following general optimization problem:

$$\begin{array}{ll} \text{minimize} & f(x)+g(z) \\ \text{s.t.} & Ax+Bz=c \end{array}. \tag{C1}$$

We form the augmented Lagrangian

$$L\rho(x,z,y)= f(x)+g(z)+y^T(Ax+Bz-c)+\frac{\rho}{2}\|Ax+Bz-c\|_2^2. \tag{C2}$$

ADMM algorithm is given by

$$\begin{aligned} x^{k+1} &= \arg\min_x L_\rho(x,z^k,y^k) \\ z^{k+1} &= \arg\min_z L\rho(x^{k+1},z,y^k) \\ y^{k+1} &= y^k+\rho(Ax^{k+1}+Bz^{k+1}-c). \end{aligned} \tag{C3}$$

Combining the linear and quadratic terms in the augmented Lagrangian and scaling the dual variable yields

$$y^T(Ax+Bz-c)+\frac{\rho}{2}\|Ax+Bz-c\|_2^2=\frac{\rho}{2}\|r+u\|_2^2-\frac{\rho}{2}\|u\|_2^2, \tag{C4}$$

where $r=Ax+Bz-c, u=\frac{1}{\rho}y$ is the scaled dual variable.

The scaled form of ADMM in algorithm (C3) can be expressed as

$$x^{k+1}=\arg\min_x\,(f(x)+\frac{\rho}{2}\|Ax+Bz^k-c+u^k\|_2^2)$$

$$z^{k+1}=\arg\min_z\,(g(z)+\frac{\rho}{2}\|Ax^{k+1}+Bz-c+u^k\|_2^2)$$

$$u^{k+1} = u^k + Ax^{k+1} + Bz^{k+1} - c. \quad \text{(C5)}$$

Let $f_j(\beta_j^{(s)}) = \| Z\theta_j^{(s-1)} - X\beta_j^{(s)} \|_2^2$ , $B^{(s)} = [\beta_1^{(s)},...,\beta_d^{(s)}] = [\beta_1^{*(s)},...,\beta_p^{*(s)}]^T$ and

$$g(\alpha_1^{(s)},...,\alpha_d^{(s)}) = g(\alpha_1^{*(s)},...,\alpha_p^{*(s)}) = \lambda\left((1-\delta)\sum_{i=1}^{p}\| \alpha_i^{*(s)} \|_2^2 + \delta\sum_{i=1}^{p}\| \alpha_i^{*(s)} \|_2^{1-r}\right),$$ where

$\alpha^{(s)} = [\alpha_1^{(s)},...,\alpha_d^{(s)}] = [\alpha_1^{*(s)},...,\alpha_p^{*(s)}]^T$ . The problem (12a) in the text can be rewritten as

$$\min \sum_{j=1}^{d} f_j(\beta_j^{(s)}) + g(\alpha_1^{(s)},...,\alpha_d^{(s)}) \quad \text{(C6)}$$

s.t. $\beta_j^{(s)} - \alpha_j^{(s)} = 0, j = 1,...,d$ .

The scaled ADMM for solving problem (C6) is given by

$$\begin{aligned}
\beta_j^{(s)(m+1)} &= \arg\min_{\beta} \ (f_j(\beta_j^{(s)})) + \frac{\rho}{2}\| \beta_j^{(s)} - \alpha_j^{(s)(m)} + u_j^{(s)(m)} \|_2^2) \\
&= (X^T X + \frac{\rho}{2} I)^{-1}[X^T Z\theta_j^{(s)} + \frac{\rho}{2}(\alpha_j^{(s)(m)} - u_j^{(s)(m)})] \\
&= \frac{2}{\rho}\left[I + \frac{2}{\rho}X^T\left(-I - \frac{2}{\rho}XX^T\right)^{-1} X\right]\left[X^T Z\theta_j^{(s)} + \frac{\rho}{2}(\alpha_j^{(s)(m)} - u_j^{(s)(m)})\right] \\
&= \frac{2}{\rho}\left[X^T Z\theta_j^{(s)} + \frac{\rho}{2}(\alpha_j^{(s)(m)} - u_j^{(s)(m)})\right] + \frac{4}{\rho^2}X^T\left(-I - \frac{2}{\rho}XX^T\right)^{-1} X\left[X^T Z\theta_j^{(s)} + \frac{\rho}{2}(\alpha_j^{(s)(m)} - u_j^{(s)(m)})\right] \\
&= \frac{2}{\rho}\left[X^T Z\theta_j^{(s)} + \frac{\rho}{2}(\alpha_j^{(s)(m)} - u_j^{(s)(m)})\right] + \left[\frac{4}{\rho^2}X^T\left(-I - \frac{2}{\rho}XX^T\right)^{-1}\right]X\left[X^T Z\theta_j^{(s)} + \frac{\rho}{2}(\alpha_j^{(s)(m)} - u_j^{(s)(m)})\right]
\end{aligned}$$

(C7)

$$\alpha^{*(s)(m+1)} = \arg\min_{\alpha^{(s)}} \ (g(\alpha^{(s)}) + \frac{\rho}{2}\sum_{j=1}^{d}\| \alpha_j^{(s)} - \beta_j^{(s)(m+1)} - u_j^{(s)(m)} \|_2^2) \quad \text{(C8)}$$

$$u_j^{(s)(m+1)} = u_j^{(s)(m)} + \beta_j^{(s)(m+1)} - \alpha_j^{(s)(m+1)} . \quad \text{(C9)}$$

Now we study how to solve the non-smooth optimization problem (C8).

First, we assume that $\|\alpha_i^{*(s)}\|_2 \neq 0, i=1,\ldots,p.$ By definition, we have

$$\frac{\partial g(\alpha)}{\partial \alpha_j^{(s)}} = \lambda\left(2(1-\delta)\alpha_j + \delta(1-r)[\frac{\alpha_{1j}^{(s)}}{\|\alpha_1^{*(s)}\|_2^{1+r}},\cdots,\frac{\alpha_{pj}^{(s)}}{\|\alpha_p^{*(s)}\|_2^{1+r}}]^T\right). \qquad \text{(C10)}$$

$$\alpha_j^{(s)} = \begin{bmatrix} \alpha_{1j}^{(s)} \\ \\ \alpha_{pj}^{(s)} \end{bmatrix}$$

Minimization of the problem (C8) can be solved by the following equation:

$$\frac{\partial g(\alpha)}{\partial \alpha_j^{(s)}} + \rho(\alpha_j^{(s)} - \beta_j^{(s)(m+1)} - u_j^{(s)(m)}) = 0,\ \ \mathrm{j}=1,\ldots,\mathrm{d}. \qquad \text{(C11)}$$

Let $B^{(s)} = [\beta_1^{(s)},\ldots,\beta_d^{(s)}] = [\beta_1^{*(s)},\cdots,\beta_p^{*(s)}]^T$, $\alpha^{(s)} = [\alpha_1^{(s)},\ldots,\alpha_d^{(s)}] = [\alpha_1^{*(s)},\cdots,\alpha_p^{*(s)}]^T$, and

$u^{(s)(m)} = [u_1^{(s)(m)},\ldots,u_d^{(s)(m)}] = [u_1^{*(s)(m)},\cdots,u_p^{*(s)(m)}]^T$.

Changing column vectors in equation (C11) to row vectors, we obtain

$$\lambda\left(2(1-\delta)\alpha_i^{*(s)} + \delta(1-r)\frac{\alpha_i^{*(s)}}{\|\alpha_i^{*(s)}\|_2^{1+r}}\right) + \rho(\alpha_i^{*(s)} - \beta_i^{*(s)(m+1)} - u_i^{*(s)(m)}) = 0, \mathrm{i}=1,\ldots,\mathrm{p}$$

which implies that

$$\left(\frac{\lambda\delta(1-r)}{\|\alpha_i^{*(s)}\|_2^{1+r}} + \rho + 2\lambda(1-\delta)\right)\alpha_i^{*(s)} = \rho(\beta_i^{*(s)(m+1} + u_i^{*(s)(m)}),\ i=1,\ldots,p. \qquad \text{(C12)}$$

Dividing equation (C12) by $\rho$, we obtain

$$\left(1 + \frac{2\lambda(1-\delta)}{\rho} + \frac{\lambda\delta(1-r)}{\rho\|\alpha_i^{*(s)}\|_2^{1+r}}\right)\alpha_i^{*(s)} = \beta_i^{*(s)(m+1)} + u_i^{*(s)(m)}. \qquad \text{(C13)}$$

Taking norm $\|.\|_2$ on both sides of equation (C13), we have

$$\left(1+\frac{\lambda(1-\delta)}{\rho}+\frac{\lambda\delta(1-r)}{\rho\|\alpha_i^{*(s)}\|_2^{1+r}}\right)^{(1+r)}\|\alpha_i^{*(s)}\|_2^{1+r}=\|\beta_i^{*(s)(m+1)}+u_i^{*(s)(m)}\|_2^{(1+r)}. \quad \text{(C14)}$$

By approximation, we have

$$\left(1+\frac{2\lambda(1-\delta)}{\rho}+\frac{\lambda\delta(1-r)}{\rho\|\alpha_i^{*(s)}\|_2^{1+r}}\right)^{(1+r)}=1+(1+r)\left(\frac{2\lambda(1-\delta)}{\rho}+\frac{\lambda\delta(1-r)}{\rho\|\alpha_i^{*(s)}\|_2^{1+r}}\right).$$

Substituting the above equation into equation (C14), we obtain

$$\|\alpha_i^{*(s)}\|_2^{1+r}=\left(\frac{\|\beta_i^{*(s)(m+1)}+u_i^{*(s)(m)}\|_2^{(1+r)}-\dfrac{\lambda\delta(1-r^2)}{\rho}}{1+\dfrac{2\lambda(1-\delta)(1+r)}{\rho}}\right). \quad \text{(C15)}$$

Therefore, we have

$$\|\alpha_i^{*(s)}\|_2=\left(\frac{\|\beta_i^{*(s)(m+1)}+u_i^{*(s)(m)}\|_2^{(1+r)}-\dfrac{\lambda\delta(1-r^2)}{\rho}}{1+\dfrac{2\lambda(1-\delta)(1+r)}{\rho}}\right)^{\frac{1}{1+r}}. \quad \text{(C16)}$$

Substituting equation (C16) into equation (C14) results in

$$\begin{aligned}1+\frac{2\lambda(1-\delta)}{\rho}+\frac{\lambda\delta(1-r)}{\rho\|\alpha_i^{*(s)}\|_2^{1+r}}&=\frac{\|\beta_i^{*(s)(m+1)}+u_i^{*(s)(m)}\|_2}{\|\alpha_i^{*(s)}\|_2}\\&=\frac{\|\beta_i^{*(s)(m+1)}+u_i^{*(s)(m)}\|_2}{\left(\dfrac{\|\beta_i^{*(s)(m+1)}+u_i^{*(s)(m)}\|_2^{(1+r)}-\dfrac{\lambda\delta(1-r^2)}{\rho}}{1+\dfrac{2\lambda(1-\delta)(1+r)}{\rho}}\right)^{\frac{1}{1+r}}}.\end{aligned} \quad \text{(C17)}$$

Substituting equation (C17) into equation (C13), we obtain

$$\alpha_i^{*(s)(m+1)} = \frac{\beta_i^{*(s)(m+1)} + u_i^{*(s)(m)}}{1 + \frac{2\lambda(1-\delta)}{\rho} + \frac{\lambda\delta(1-r)}{\rho \| \alpha_i^{*(s)} \|_2^{1+r}}}$$

$$= (\beta_i^{*(s)(m+1)} + u_i^{*(s)(m)}) \left( \frac{\left( \frac{\| \beta_i^{*(s)(m+1)} + u_i^{*(s)(m)} \|_2^{(1+r)} - \frac{\lambda\delta(1-r^2)}{\rho}}{1 + \frac{2\lambda(1-\delta)(1+r)}{\rho}} \right)^{\frac{1}{1+r}}}{\| \beta_i^{*(s)(m+1)} + u_i^{*(s)(m)} \|_2} \right) \quad \text{(C18)}$$

Let

$$\alpha^{(s)(m+1)} = [\alpha_1^{*(s)(m+1)}, \cdots, \alpha_p^{*(s)(m+1)}]^T = [\alpha_1^{(s)(m+1)}, \ldots, \alpha_d^{(s)(m+1)}]. \quad \text{(C19)}$$

Substituting equation (C19) into equation (C9), we can obtain $u_j^{(s)(m+1)}, j = 1, \ldots, d$.

Next we briefly discuss how to obtain solution (12b). Using the Lagrangian multiplier method, the constrained optimization problem (12b) can be formulated as the following unconstrained optimization problem:

$$L(\theta_i, \lambda_\theta, \mu) = (Z\theta_i^{(s)} - X\beta_i^{(s)})^T (Z\theta_i^{(s)} - X\beta_i^{(s)}) + \lambda_\theta (1 - \theta_i^{(s)T} D\theta_i^{(s)}) + \sum_{j=1}^{i-1} \mu_j \theta_i^{(s)T} D\theta_j^{(s)} . \quad \text{(C20)}$$

Differential $L(\theta_i, \lambda_\theta, \mu)$ with respect to $\theta_i^{(s)}$ and setting it to be equal to zero, we obtain

$$Z^T (Z\theta_i^{(s)} - X\beta_i^{(s)}) - \lambda_\theta D\theta_i^{(s)} + \sum_{j=1}^{i-1} \mu_j D\theta_j^{(s)} = 0, \quad \text{(C21)}$$

which implies that

$$\lambda_\theta = [\theta_i^{(s)T} Z^T Z\theta_i^{(s)} - \theta_i^{(s)T} Z^T X\beta_i^{(s)}] = n - \theta_i^{(s)T} Z^T X\beta_i^{(s)} \quad \text{(C22)}$$

and

$\mu_j = \theta_j^{(s)T} Z^T X \beta_i^{(s)}$, j=1,…,i-1. (C23)

Substituting equations (C22) and (C23) into equation (C21), we obtain

$$(n-\lambda_\theta) D\theta_i^{(s)} - Z^T X\beta_i^{(s)} + D\sum_{j=1}^{i-1} \theta_j^{(s)} \theta_j^{(s)T} Z^T X \beta_i^{(s)} = 0 . \quad \text{(C24)}$$

Let $Q_i^{(s)} = [\theta_1^{(s)}, ..., \theta_{i-1}^{(s)}]$, where $Q_1^{(s)} = [1,0,...,0]^T$ . Then, we have $Q_i^{(s)} Q_i^{(s)T} = \sum_{j=1}^{i-1} \theta_j^{(s)} \theta_j^{(s)T}$ . The solution to equation (C24) is given by

$$\theta_i^{(s)} = a(I - Q_i^{(s)} Q_i^{(s)T} D) D^{-1} Z^T X \beta_i^{(s)} \text{, where } a = \frac{1}{n-\lambda_\theta} . \quad \text{(C25)}$$

We can use the Newton's method to solve the nonlinear Equation (C25).

The recursive formula for solving it is given by

$$\theta_i^{(s)(m+1)} = \theta_i^{(s)(m)} - [(Z^T X\beta) \otimes \theta_i^{(s)(m)}]^{-1} [\theta_i^{T(s)(m)} Z^T X \beta_i^{(s)} \theta_i^{(s)(m)} - (I - Q_i^{(s)} Q_i^{(s)T} D) D^{-1} Z^T X \beta_i^{(s)}].$$

We can also use iterative algorithm to equation (C25). Just iteratively,

$$\theta_i^{(s)(m+1)} = \frac{(I - Q_i^{(s)} Q_i^{(s)T} D) D^{-1} Z^T X \beta_i^{(s)}}{\theta_i^{(s)T(m)} Z^T X \beta_i^{(s)}} . \quad \text{(C26)}$$

Finally, we set $\theta_i^{(s)} = \frac{\theta_i^{(s)(m+1)}}{\sqrt{\theta_i^{(s)T(m+1)} D \theta_i^{(s)(m+1)}}}$ , $Q_{i+1} = [Q_i : \theta_i], i = 1,...,d.$ (C27)

.

**Table S2**. AUC values of three methods for classifying RA.

| Sample Fold | Training Dataset | | | Test Dataset | | |
|---|---|---|---|---|---|---|
| | Sparse SDR | Sparse Logistic | P-value Rank | Sparse SDR | Sparse Logistic | P-value Rank |
| CV - 1 | 0.8650 | 0.8099 | 0.8334 | 0.8469 | 0.8235 | 0.8418 |
| CV - 2 | 0.8607 | 0.8319 | 0.8372 | 0.8447 | 0.8365 | 0.8344 |
| CV - 3 | 0.8506 | 0.8339 | 0.8358 | 0.8203 | 0.8309 | 0.8248 |
| CV - 4 | 0.8499 | 0.8364 | 0.8322 | 0.8199 | 0.8291 | 0.8252 |
| CV - 5 | 0.8547 | 0.8374 | 0.8393 | 0.8172 | 0.8142 | 0.8178 |
| Average | 0.8562 | 0.8299 | 0.8356 | 0.8298 | 0.8268 | 0.8288 |

**Table S3**. AUC value of three methods for classifying EOMI.

| Sample Fold | Training Dataset | | | Test Dataset | | |
|---|---|---|---|---|---|---|
| | Sparse SDR | Sparse Logistic | P-value Rank | Sparse SDR | Sparse Logistic | P-value Rank |
| CV - 1 | 1.0000 | 1.0000 | 0.9531 | 0.7740 | 0.6361 | 0.7379 |
| CV - 2 | 1.0000 | 1.0000 | 0.9524 | 0.6900 | 0.6493 | 0.6196 |
| CV - 3 | 1.0000 | 1.0000 | 0.9462 | 0.7060 | 0.6148 | 0.6621 |
| CV - 4 | 1.0000 | 1.0000 | 0.9471 | 0.7240 | 0.6377 | 0.7006 |
| CV - 5 | 1.0000 | 1.0000 | 0.9472 | 0.6790 | 0.7022 | 0.7028 |
| Average | 1.0000 | 1.0000 | 0.9492 | 0.7140 | 0.6480 | 0.6846 |

**References**

1. Cook, R.D. (2004). Testing predictor contributions in sufficient dimension reduction. The Annals of Statistics 32, 1062-1092.

2. Boyd, S., Parikh, N., Chu, E., Peleato, B., and Eckstein, J. (2011). Distributed optimization and statistical learning via the alternating direction method of multipliers. Foundations and Trends® in Machine Learning 3, 1-122.

Table S1. Selected best SNPs for risk prediction of EOMI.

| CHR | Position | Gene | RS | P-value |
|---|---|---|---|---|
| SNP16 | 66432424 | CDH5 | rs3826229 | 1.63E-17 |
| SNP2 | 102968211 | IL1RL1 | rs10192036 | 4.50E-17 |
| SNP2 | 102968211 | IL1RL1 | rs10204137 | 4.50E-17 |
| SNP4 | 100443784 | C4orf17 | rs61732380 | 1.03E-13 |
| SNP4 | 100443784 | C4orf17 | rs36110345 | 1.03E-13 |
| SNP8 | 18729818 | PSD3 | rs7003060 | 1.37E-13 |
| SNP3 | 112648125 | CD200R1 | rs6438117 | 1.53E-12 |
| SNP3 | 112648125 | CD200R1 | rs71625219 | 1.53E-12 |
| SNP3 | 112648125 | CD200R1 | rs113267312 | 1.53E-12 |
| SNP20 | 6065731 | FERMT1 | rs2232073 | 1.27E-10 |
| SNP11 | 118221350 | CD3G | rs71469175 | 6.72E-10 |
| SNP11 | 118221350 | CD3G | rs3753058 | 6.72E-10 |
| SNP3 | 46007825 | FYCO1 | rs13059238 | 2.11E-09 |
| SNP3 | 46007825 | FYCO1 | rs71622515 | 2.11E-09 |
| SNP16 | 1544301 | TELO2 | rs2667660 | 3.38E-09 |
| SNP16 | 1544301 | TELO2 | rs35400877 | 3.38E-09 |
| SNP16 | 1544301 | TELO2 | rs2667661 | 3.38E-09 |
| SNP4 | 175688141 | GLRA3 | rs12651268 | 1.03E-08 |
| SNP8 | 27925204 | C8orf80 | rs4732620 | 1.09E-06 |
| SNP13 | 36229873 | NBEA | rs2274550 | 1.18E-06 |
| SNP11 | 2329995 | TSPAN32 | rs11554947 | 5.18E-06 |
| SNP11 | 94322353 | PIWIL4 | rs11020846 | 6.26E-06 |
| SNP8 | 8234113 | AC068353.1 | rs2921005 | 1.25E-05 |
| SNP1 | 233398713 | PCNXL2 | rs1033325 | 1.26E-05 |
| SNP9 | 113169631 | SVEP1 | rs71492888 | 2.46E-05 |
| SNP9 | 113169631 | SVEP1 | rs7030192 | 2.46E-05 |
| SNP19 | 15226970 | ILVBL | rs2074262 | 3.26E-05 |
| SNP6 | 46672943 | PLA2G7 | rs1051931 | 3.49E-05 |
| SNP19 | 15295134 | NOTCH3 | rs1043996 | 4.26E-05 |
| SNP8 | 21550800 | GFRA2 | rs1128397 | 4.77E-05 |
| SNP10 | 47087371 | PPYR1 | rs1048156 | 5.80E-05 |
| SNP12 | 62894636 | MON2 | rs17120341 | 6.00E-05 |
| SNP6 | 32629868 | HLA-DQB1 | rs1049088 | 6.66E-05 |
| SNP6 | 32629868 | HLA-DQB1 | rs36233042 | 6.66E-05 |
| SNP6 | 32629868 | HLA-DQB1 | rs76163881 | 6.66E-05 |
| SNP6 | 32629868 | HLA-DQB1 | rs79326791 | 6.66E-05 |
| SNP6 | 32629868 | HLA-DQB1 | rs9273970 | 6.66E-05 |
| SNP10 | 72500863 | ADAMTS14 | rs10999500 | 7.37E-05 |
| SNP6 | 32609126 | HLA-DQA1 | rs1071630 | 7.97E-05 |
| SNP6 | 32609126 | HLA-DQA1 | rs80027061 | 7.97E-05 |
| SNP5 | 140736689 |  | rs4329068 | 8.34E-05 |
| SNP5 | 140767991 | PCDHGB4 | rs77505166 | 8.34E-05 |
| SNP5 | 140789819 | PCDHGB6 | rs3749768 | 8.34E-05 |
| SNP6 | 47847179 | C6orf138 | rs3799276 | 8.72E-05 |
| SNP6 | 47847179 | C6orf138 | rs74553560 | 8.72E-05 |
| SNP1 | 107599918 | PRMT6 | rs2232016 | 8.81E-05 |
| SNP5 | 16673975 | MYO10 | rs25901 | 9.28E-05 |
| SNP20 | 25011423 | ACSS1 | rs6050259 | 9.28E-05 |
| SNP7 | 142247213 | | rs361360 | 9.52E-05 |
| SNP5 | 134782510 | C5orf20 | rs12520809 | 0.000107896 |
| SNP11 | 34938310 | PDHX | rs1049307 | 0.000118896 |
| SNP17 | 73969835 | ACOX1 | rs3744032 | 0.000119662 |
| SNP5 | 121356189 | SRFBP1 | rs61734326 | 0.000132781 |
| SNP20 | 744415 | C20orf54 | rs3746804 | 0.00013338 |
| SNP3 | 174951756 | NAALADL2 | rs4371530 | 0.000136081 |
| SNP8 | 98943446 | MATN2 | rs2290470 | 0.000136294 |
| SNP13 | 22275394 | FGF9 | rs9509841 | 0.000137687 |
| SNP1 | 163313539 | NUF2 | rs11802875 | 0.000139896 |
| SNP4 | 84230619 | HPSE | rs11099592 | 0.000158398 |
| SNP17 | 3635740 | ITGAE | rs3744679 | 0.000161535 |
| SNP6 | 32610495 | HLA-DQA1 | rs9272793 | 0.000166793 |
| SNP6 | 32610495 | HLA-DQA1 | rs77143934 | 0.000166793 |
| SNP4 | 169083694 | ANXA10 | rs6836994 | 0.000187538 |
| SNP13 | 87045827 | | rs9513670 | 0.000197578 |
| SNP6 | 165713961 | C6orf118 | rs510579 | 0.000213693 |
| SNP6 | 41739173 | FRS3 | rs35310379 | 0.000219718 |
| SNP6 | 41739173 | FRS3 | rs3747747 | 0.000219718 |
| SNP1 | 203024767 | PPFIA4 | rs3736314 | 0.000239554 |
| SNP7 | 139415943 | HIPK2 | rs4074826 | 0.000242369 |
| SNP8 | 30700598 | TEX15 | rs61740968 | 0.000243818 |
| SNP11 | 4929061 | OR51A7 | rs10500627 | 0.000250944 |
| SNP6 | 132966279 | TAAR1 | rs8192620 | 0.000268658 |
| SNP21 | 37617630 | DOPEY2 | rs4817788 | 0.000270773 |
| SNP21 | 37617630 | DOPEY2 | rs145337645 | 0.000270773 |
| SNP1 | 163297322 | NUF2 | rs16852612 | 0.000274439 |
| SNP1 | 245704130 | KIF26B | rs61754898 | 0.000306782 |
| SNP11 | 33564123 | C11orf41 | rs2076623 | 0.000312153 |
| SNP4 | 15569018 | CC2D2A | rs73125627 | 0.000322019 |
| SNP14 | 99182626 | C14orf177 | rs4905757 | 0.000330498 |
| SNP16 | 30999198 | HSD3B7 | rs17849880 | 0.00033063 |
| SNP16 | 11362729 | TNP2 | rs11640138 | 0.000332421 |
| SNP2 | 49510173 | | rs7578654 | 0.000333252 |
| SNP11 | 92088177 | FAT3 | rs7479732 | 0.000342513 |
| SNP8 | 25222165 | DOCK5 | rs2271111 | 0.000373628 |
| SNP17 | 34340284 | CCL23 | rs1003645 | 0.000374423 |
| SNP9 | 126134542 | CRB2 | rs61740213 | 0.000394884 |
| SNP9 | 132580901 | TOR1A | rs1801968 | 0.000411334 |
| SNP9 | 132580901 | TOR1A | rs149375117 | 0.000411334 |
| SNP1 | 19027239 | PAX7 | rs2743201 | 0.000427295 |
| SNP20 | 57571763 | CTSZ | rs9760 | 0.000446305 |
| SNP10 | 64564934 | ADO | rs10995311 | 0.000511897 |
| SNP10 | 64564934 | ADO | rs3088257 | 0.000511897 |
| SNP4 | 7717012 | SORCS2 | rs2285781 | 0.00053854 |
| SNP2 | 86272758 | POLR1A | rs1561328 | 0.000543653 |
| SNP12 | 121622304 | P2RX7 | rs3751143 | 0.000560399 |
| SNP7 | 6063283 | AIMP2 | rs4560 | 0.000567636 |
| SNP10 | 30317826 | KIAA1462 | rs7920682 | 0.0005957 |
| SNP10 | 30317838 | KIAA1462 | rs7920686 | 0.0005957 |
| SNP2 | 223423362 | SGPP2 | rs4674664 | 0.000598681 |
| SNP11 | 106408071 | | rs1791468 | 0.000598748 |
| SNP2 | 167056337 | SCN9A | rs149207258 | 0.000607194 |
| SNP1 | 223954080 | CAPN2 | rs17599 | 0.000613954 |
| SNP3 | 47162661 | SETD2 | rs6767907 | 0.000623037 |
| SNP17 | 29855661 | RAB11FIP4 | rs2074149 | 0.000638447 |
| SNP15 | 86807761 | AGBL1 | rs10520618 | 0.000640316 |
| SNP17 | 78157995 | CARD14 | rs4889990 | 0.000643195 |
| SNP1 | 17257840 | CROCC | rs143501240 | 0.000646957 |
| SNP17 | 72937605 | OTOP3 | rs7210616 | 0.0006512 |
| SNP19 | 46181392 | GIPR | rs1800437 | 0.00067412 |
| SNP14 | 54417522 | BMP4 | rs17563 | 0.000683884 |
| SNP10 | 28378758 | MPP7 | rs2997211 | 0.000686523 |
| SNP22 | 18300240 | XXbac-B461K10.4 | rs5992854 | 0.00069583 |
| SNP5 | 180057293 | FLT4 | rs34221241 | 0.000708739 |
| SNP11 | 8246181 | LMO1 | rs1042359 | 0.000721245 |
| SNP22 | 40054948 | CACNA1I | rs5757761 | 0.000725376 |
| SNP12 | 111884608 | SH2B3 | rs3184504 | 0.000744898 |
| SNP10 | 69959242 | MYPN | rs7079481 | 0.000763317 |
| SNP1 | 58971831 | OMA1 | rs3087585 | 0.000785281 |
| SNP9 | 140262426 | EXD3 | rs11533158 | 0.000789769 |
| SNP15 | 58838038 | LIPC | rs6084 | 0.000853223 |
| SNP5 | 36143392 | LMBRD2 | rs267766 | 0.000868087 |
| SNP10 | 27317840 | ANKRD26 | rs10829163 | 0.000880815 |
| SNP2 | 131103635 | IMP4 | rs11542415 | 0.000895635 |
| SNP17 | 72938100 | OTOP3 | rs1542752 | 0.000924513 |
| SNP19 | 38702950 | DPF1 | rs11547760 | 0.000929175 |
| SNP5 | 5460569 | KIAA0947 | rs2578565 | 0.000941433 |
| SNP1 | 204589101 | LRRN2 | rs3789044 | 0.000944673 |
| SNP6 | 112457390 | LAMA4 | rs2032567 | 0.000946845 |
| SNP7 | 100420155 | EPHB4 | rs56173078 | 0.00095857 |
| SNP7 | 45140914 | TBRG4 | rs1042984 | 0.000995737 |
| SNP17 | 46262171 | SKAP1 | rs2278868 | 0.001016362 |
| SNP22 | 36537763 | APOL3 | rs61741884 | 0.001045407 |
| SNP17 | 33998904 | AP2B1 | rs17670584 | 0.001051024 |
| SNP7 | 129813738 | TMEM209 | rs3823482 | 0.001069436 |
| SNP20 | 19634732 | SLC24A3 | rs3790278 | 0.001102902 |
| SNP20 | 19634747 | SLC24A3 | rs3790279 | 0.001102902 |
| SNP12 | 121691096 | CAMKK2 | rs1132780 | 0.001160903 |
| SNP19 | 49377873 | PPP1R15A | rs35023389 | 0.001167378 |
| SNP17 | 3981290 | ZZEF1 | rs78806449 | 0.001235475 |
| SNP1 | 159785425 | FCRL6 | rs4443889 | 0.001238524 |
| SNP17 | 48356260 | TMEM92 | rs6504642 | 0.001267065 |
| SNP11 | 43940644 | ALKBH3 | rs1048928 | 0.00127668 |
| SNP19 | 15198631 | OR1I1 | rs8105737 | 0.001289778 |
| SNP5 | 181762 | PLEKHG4B | rs11133847 | 0.001298398 |
| SNP5 | 140057535 | HARS | rs2230361 | 0.001299713 |
| SNP9 | 112900199 | PALM2-AKAP2 | rs914358 | 0.001359177 |
| SNP11 | 113103996 | NCAM1 | rs584427 | 0.00137137 |
| SNP20 | 44417593 | WFDC3 | rs73122754 | 0.00141909 |
| SNP1 | 207881557 | CR1L | rs6683902 | 0.001441213 |
| SNP6 | 133015271 | VNN1 | rs2272996 | 0.001452823 |
| SNP22 | 22869649 | ZNF280A | rs361721 | 0.001468635 |
| SNP11 | 68748003 | MRGPRD | rs34847539 | 0.001473019 |
| SNP4 | 3006043 | GRK4 | rs1024323 | 0.001478199 |
| SNP15 | 101922323 | PCSK6 | rs1058260 | 0.001518848 |
| SNP20 | 33657126 | TRPC4AP | rs1998233 | 0.001545519 |
| SNP3 | 10251340 | IRAK2 | rs464286 | 0.001566101 |
| SNP11 | 62766431 | SLC22A8 | rs2276299 | 0.001569886 |
| SNP16 | 11002904 | CIITA | rs2228238 | 0.001612024 |
| SNP8 | 17612846 | MTUS1 | rs3739408 | 0.001672816 |
| SNP6 | 46684222 | PLA2G7 | rs1805017 | 0.001682981 |
| SNP6 | 46684222 | PLA2G7 | rs189781341 | 0.001682981 |
| SNP22 | 23627369 | BCR | rs140504 | 0.001702807 |
| SNP12 | 52435691 | NR4A1 | rs7316491 | 0.001813017 |
| SNP14 | 77304292 | C14orf166B | rs4903497 | 0.001830151 |
| SNP5 | 176516631 | FGFR4 | rs1966265 | 0.001838477 |
| SNP19 | 53770764 | VN1R6P | rs74429916 | 0.001839087 |
| SNP19 | 53770764 | VN1R6P | rs112711591 | 0.001839087 |
| SNP7 | 6449496 | DAGLB | rs2303361 | 0.001839087 |
| SNP14 | 59112677 | DACT1 | rs34015825 | 0.001841754 |
| SNP15 | 83481880 | WHAMM | rs35270670 | 0.001853455 |
| SNP4 | 101109161 | DDIT4L | rs3749604 | 0.001853455 |
| SNP11 | 6023818 | OR56A4 | rs10839221 | 0.001911988 |
| SNP1 | 234546245 | TARBP1 | rs1141264 | 0.001925138 |
| SNP1 | 234546245 | TARBP1 | rs188908188 | 0.001925138 |
| SNP6 | 111695268 | REV3L | rs455732 | 0.001970038 |
| SNP12 | 53682986 | ESPL1 | rs3214023 | 0.002005133 |
| SNP1 | 197896728 | LHX9 | rs12046958 | 0.002020414 |
| SNP19 | 35616316 | LGI4 | rs36102542 | 0.002120022 |
| SNP19 | 35616316 | LGI4 | rs12610234 | 0.002120022 |
| SNP14 | 93199080 | LGMN | rs2236264 | 0.002123567 |
| SNP6 | 37623626 | MDGA1 | rs13204070 | 0.002136242 |
| SNP16 | 89350178 | ANKRD11 | rs2279349 | 0.002144931 |
| SNP16 | 61687701 | CDH8 | rs16963768 | 0.002172345 |
| SNP18 | 47801800 | MBD1 | rs140690 | 0.002195131 |
| SNP5 | 75003678 | POC5 | rs2307111 | 0.002207158 |
| SNP2 | 26536299 | GPR113 | rs11886746 | 0.002223166 |
| SNP4 | 122683007 | TMEM155 | rs4370153 | 0.002262339 |
| SNP2 | 25655772 | DTNB | rs7583475 | 0.002283815 |
| SNP9 | 106896809 | SMC2 | rs7872034 | 0.002414487 |
| SNP8 | 62003453 | | rs1367972 | 0.002427899 |
| SNP16 | 1306642 | TPSD1 | rs61739908 | 0.00247017 |
| SNP1 | 247769217 | OR2G3 | rs61750030 | 0.00249081 |
| SNP14 | 60582053 | C14orf135 | rs150688 | 0.002505119 |
| SNP14 | 75537381 | FAM164C | rs11546525 | 0.002519914 |
| SNP20 | 3686436 | SIGLEC1 | rs6037651 | 0.002535009 |
| SNP20 | 3686436 | SIGLEC1 | rs142977227 | 0.002535009 |
| SNP1 | 166905927 | ILDR2 | rs33958744 | 0.002571962 |
| SNP11 | 125788678 | DDX25 | rs683155 | 0.00259469 |
| SNP1 | 180905448 | KIAA1614 | rs3795504 | 0.002608144 |
| SNP9 | 135538050 | DDX31 | rs35918594 | 0.002668691 |
| SNP11 | 6340525 | PRKCDBP | rs12570 | 0.002680747 |
| SNP19 | 50926265 | SPIB | rs113934432 | 0.002717404 |
| SNP19 | 50926265 | SPIB | rs11546996 | 0.002717404 |
| SNP7 | 151074296 | NUB1 | rs2159158 | 0.002718886 |
| SNP19 | 53454543 | ZNF816 | rs11084210 | 0.002721225 |
| SNP19 | 33655144 | WDR88 | rs1981827 | 0.0027489 |
| SNP2 | 152573981 | NEB | rs4611637 | 0.002836538 |
| SNP14 | 20404614 | OR4K1 | rs34608158 | 0.002837025 |
| SNP20 | 3682149 | SIGLEC1 | rs13037869 | 0.002885972 |
| SNP11 | 25098938 | LUZP2 | rs7930185 | 0.002892847 |
| SNP4 | 143067054 | INPP4B | rs35390852 | 0.002894568 |
| SNP9 | 95887320 | NINJ1 | rs2275848 | 0.002902858 |
| SNP20 | 3672836 | SIGLEC1 | rs910653 | 0.002945635 |
| SNP11 | 6790028 | OR2AG2 | rs10839616 | 0.002949935 |
| SNP16 | 82033810 | SDR42E1 | rs11542462 | 0.002983261 |
| SNP20 | 210061 | DEFB129 | rs13045643 | 0.003038182 |
| SNP5 | 121488506 | ZNF474 | rs35262183 | 0.003092279 |
| SNP19 | 6713262 | C3 | rs1047286 | 0.003094577 |
| SNP9 | 115950156 | FKBP15 | rs10465129 | 0.003129506 |
| SNP10 | 127724778 | ADAM12 | rs1044122 | 0.003156829 |
| SNP20 | 3285126 | C20orf194 | rs2254916 | 0.003231891 |
| SNP7 | 121080957 | | rs7795121 | 0.00329861 |
| SNP9 | 7103816 | KDM4C | rs3763651 | 0.003336063 |
| SNP8 | 131124559 | ASAP1 | rs966185 | 0.003419196 |
| SNP6 | 56044578 | COL21A1 | rs2038149 | 0.003425879 |
| SNP14 | 95053890 | SERPINA5 | rs6115 | 0.003437628 |
| SNP14 | 95053890 | SERPINA5 | rs6111 | 0.003437628 |
| SNP17 | 39580559 | KRT37 | rs9916475 | 0.00345537 |
| SNP17 | 39580562 | KRT37 | rs9916484 | 0.00345537 |
| SNP9 | 100675816 | C9orf156 | rs1127703 | 0.003506947 |
| SNP13 | 49951142 | CAB39L | rs8002829 | 0.003601865 |
| SNP20 | 2996497 | PTPRA | rs1178016 | 0.003630255 |
| SNP11 | 85692181 | PICALM | rs76719109 | 0.003648551 |
| SNP6 | 130154667 | C6orf191 | rs9492393 | 0.00367718 |
| SNP17 | 72469958 | CD300A | rs1127737 | 0.003684653 |
| SNP16 | 11001743 | CIITA | rs2229320 | 0.003704085 |
| SNP11 | 64680819 | ATG2A | rs618006 | 0.003707783 |
| SNP15 | 59347929 | RNF111 | rs1446239 | 0.003733095 |
| SNP19 | 38573149 | SIPA1L3 | rs10405667 | 0.003740363 |
| SNP9 | 132400480 | ASB6 | rs3739851 | 0.00380061 |
| SNP6 | 29581041 | GABBR1 | rs29225 | 0.003917066 |
| SNP6 | 29581041 | GABBR1 | rs116607207 | 0.003917066 |
| SNP6 | 29581110 | GABBR1 | rs17854217 | 0.003917066 |
| SNP6 | 29581110 | GABBR1 | rs116075387 | 0.003917066 |
| SNP6 | 29581113 | GABBR1 | rs17854216 | 0.003917066 |
| SNP6 | 29581113 | GABBR1 | rs114502897 | 0.003917066 |
| SNP17 | 36926767 | PIP4K2B | rs228290 | 0.00393365 |
| SNP6 | 151894505 | C6orf97 | rs953767 | 0.003988587 |
| SNP14 | 74523869 | C14orf45 | rs3742809 | 0.003996385 |
| SNP22 | 24468386 | CABIN1 | rs17854874 | 0.004135768 |
| SNP2 | 242066314 | PASK | rs2240542 | 0.004188484 |
| SNP6 | 151917660 | C6orf97 | rs34430497 | 0.004198454 |
| SNP3 | 155485302 | C3orf33 | rs358733 | 0.004273417 |
| SNP1 | 236645670 | EDARADD | rs604070 | 0.004325992 |
| SNP1 | 236645670 | EDARADD | rs147501905 | 0.004325992 |
| SNP5 | 6600150 | NSUN2 | rs3822434 | 0.004328579 |
| SNP4 | 186097045 | KIAA1430 | rs6855305 | 0.004350087 |
| SNP20 | 61040453 | GATA5 | rs6061243 | 0.00438493 |
| SNP11 | 7847472 | OR5P3 | rs1482793 | 0.004393382 |
| SNP15 | 28386626 | HERC2 | rs11636232 | 0.004412321 |
| SNP1 | 166958710 | MAEL | rs11578336 | 0.004435166 |
| SNP9 | 136328657 | C9orf7 | rs3124765 | 0.004446322 |
| SNP6 | 87970301 | ZNF292 | rs3812132 | 0.004549724 |
| SNP6 | 87994572 | GJB7 | rs4707358 | 0.004549724 |
| SNP11 | 125889526 | CDON | rs3740909 | 0.004617482 |
| SNP16 | 57762401 | CCDC135 | rs2923147 | 0.004626068 |
| SNP12 | 121416622 | HNF1A | rs1169289 | 0.004656457 |
| SNP11 | 66834252 | RHOD | rs2282502 | 0.004696561 |
| SNP22 | 25145453 | PIWIL3 | rs1892723 | 0.004771187 |
| SNP9 | 77761642 | OSTF1 | rs2146044 | 0.004824313 |
| SNP5 | 57755703 | PLK2 | rs3211270 | 0.0049139 |
| SNP14 | 77793207 | GSTZ1 | rs7975 | 0.004931472 |
| SNP22 | 22277571 | PPM1F | rs2070507 | 0.004946297 |
| SNP16 | 80718879 | CDYL2 | rs8049284 | 0.004990456 |
| SNP10 | 126517989 | FAM175B | rs2303611 | 0.005036086 |
| SNP9 | 36110063 | RECK | rs10972727 | 0.005094535 |
| SNP9 | 36110063 | RECK | rs149420244 | 0.005094535 |
| SNP1 | 86488232 | COL24A1 | rs641712 | 0.005136536 |
| SNP5 | 10239261 | FAM173B | rs2438652 | 0.005138186 |
| SNP12 | 7803646 | APOBEC1 | rs10431309 | 0.005150389 |
| SNP6 | 32363816 | BTNL2 | rs2076530 | 0.005153276 |
| SNP6 | 32363816 | BTNL2 | rs115611791 | 0.005153276 |
| SNP10 | 91066075 | IFIT2 | rs2070845 | 0.005189279 |
| SNP19 | 36530343 | THAP8 | rs10421966 | 0.005230784 |
| SNP7 | 150648789 | KCNH2 | rs1805121 | 0.005234589 |
| SNP7 | 150648789 | KCNH2 | rs33959111 | 0.005234589 |
| SNP2 | 133541575 | NCKAP5 | rs12611515 | 0.005271453 |
| SNP1 | 181725110 | CACNA1E | rs4652678 | 0.005276545 |
| SNP16 | 24801979 | TNRC6A | rs13336754 | 0.005281637 |
| SNP9 | 101340301 | GABBR2 | rs2808536 | 0.005292186 |
| SNP3 | 44692564 | ZNF35 | rs2272044 | 0.005385013 |
| SNP21 | 48022230 | S100B | rs1051169 | 0.005401526 |
| SNP17 | 15496727 | RP11-385D13.1 | rs79385100 | 0.00542131 |
| SNP2 | 39050141 | DHX57 | rs3770681 | 0.005498357 |
| SNP14 | 100808845 | WARS | rs9453 | 0.005518986 |
| SNP15 | 71125093 | LARP6 | rs3825970 | 0.005529281 |
| SNP10 | 115350597 | NRAP | rs1885434 | 0.005594473 |
| SNP11 | 801092 | LRDD | rs11539604 | 0.00562725 |
| SNP14 | 73412635 | DCAF4 | rs17856582 | 0.005682136 |
| SNP18 | 77067000 | ATP9B | rs3760541 | 0.005682487 |
| SNP18 | 77067000 | ATP9B | rs59800543 | 0.005682487 |
| SNP13 | 95264604 | GPR180 | rs9524559 | 0.005754591 |
| SNP16 | 11981487 | GSPT1 | rs3752426 | 0.005769807 |
| SNP8 | 2909992 | CSMD1 | rs6558702 | 0.005807132 |
| SNP8 | 145111979 | OPLAH | rs72695427 | 0.005828208 |
| SNP19 | 45868309 | ERCC2 | rs238406 | 0.005910642 |
| SNP17 | 48761105 | ABCC3 | rs2277624 | 0.005914725 |
| SNP1 | 242030151 | EXO1 | rs735943 | 0.006002421 |
| SNP15 | 40477831 | BUB1B | rs1801376 | 0.006028288 |
| SNP3 | 187089031 | RTP4 | rs1003995 | 0.00604319 |
| SNP18 | 334994 | COLEC12 | rs2305025 | 0.00607478 |
| SNP2 | 122216419 | CLASP1 | rs2304560 | 0.006115062 |
| SNP22 | 26193982 | MYO18B | rs738642 | 0.006120938 |
| SNP3 | 11887991 | C3orf31 | rs408600 | 0.006195941 |
| SNP12 | 406292 | KDM5A | rs2229351 | 0.006208818 |
| SNP12 | 427575 | KDM5A | rs11062385 | 0.006208818 |
| SNP9 | 32634467 | TAF1L | rs17219559 | 0.006211258 |
| SNP17 | 38640744 | TNS4 | rs2290207 | 0.006223269 |
| SNP2 | 18113623 | KCNS3 | rs4832524 | 0.006241331 |
| SNP2 | 190430177 | SLC40A1 | rs2304704 | 0.006242765 |
| SNP19 | 6709704 | C3 | rs2230205 | 0.006295225 |
| SNP6 | 7393452 | RIOK1 | rs2274212 | 0.006340462 |
| SNP13 | 31891746 | B3GALTL | rs1041073 | 0.006400305 |
| SNP16 | 56377748 | GNAO1 | rs1799917 | 0.006405128 |
| SNP17 | 4856580 | ENO3 | rs238239 | 0.006454721 |
| SNP16 | 27374400 | IL4R | rs1801275 | 0.006491844 |
| SNP15 | 59368167 | RNF111 | rs7178935 | 0.006508408 |
| SNP17 | 45451894 | C17orf57 | rs4968318 | 0.006621363 |
| SNP15 | 45968435 | SQRDL | rs1044032 | 0.006632858 |
| SNP6 | 74533192 | CD109 | rs2917887 | 0.00677447 |
| SNP22 | 26157068 | MYO18B | rs61734946 | 0.006775124 |
| SNP5 | 121355961 | SRFBP1 | rs55708726 | 0.006831666 |
| SNP4 | 175237408 | KIAA1712 | rs1553669 | 0.00685494 |
| SNP19 | 17628587 | PGLS | rs6743 | 0.006883122 |
| SNP17 | 8416901 | MYH10 | rs11374 | 0.006900759 |
| SNP4 | 37836302 | PGM2 | rs3752683 | 0.006949014 |
| SNP12 | 121881848 | KDM2B | rs10849885 | 0.006989908 |
| SNP6 | 132061420 | ENPP3 | rs17601580 | 0.007005649 |
| SNP4 | 144918712 | GYPB | rs1132783 | 0.007069757 |
| SNP4 | 144918712 | GYPB | rs184961047 | 0.007069757 |
| SNP9 | 97349666 | FBP2 | rs573212 | 0.007128858 |
| SNP4 | 186381165 | CCDC110 | rs7699724 | 0.007189703 |
| SNP16 | 89350038 | ANKRD11 | rs2279348 | 0.007279134 |
| SNP19 | 15905002 | OR10H5 | rs4808379 | 0.007306529 |
| SNP19 | 15905002 | OR10H5 | rs56548474 | 0.007306529 |
| SNP12 | 10999708 | PRR4 | rs1047699 | 0.007347502 |
| SNP4 | 143043340 | INPP4B | rs2270658 | 0.007391269 |
| SNP1 | 18149566 | ACTL8 | rs2296035 | 0.007397084 |
| SNP20 | 17600357 | RRBP1 | rs11960 | 0.007410143 |
| SNP15 | 93510603 | CHD2 | rs4777755 | 0.007456998 |
| SNP8 | 107782395 | ABRA | rs11996466 | 0.007599639 |
| SNP20 | 62328480 | RTEL1 | rs55765053 | 0.007625899 |
| SNP3 | 197579466 | LRCH3 | rs36078463 | 0.007625899 |
| SNP17 | 72927123 | OTOP2 | rs6501741 | 0.007632469 |
| SNP1 | 15790013 | CELA2A | rs3181471 | 0.007793714 |
| SNP1 | 15790013 | CELA2A | rs61745446 | 0.007793714 |
| SNP14 | 21557021 | ARHGEF40 | rs74584322 | 0.007813471 |
| SNP17 | 76178748 | TK1 | rs1143697 | 0.007902766 |
| SNP18 | 22020543 | IMPACT | rs677688 | 0.008004677 |
| SNP1 | 202304868 | UBE2T | rs14451 | 0.008217102 |
| SNP19 | 9058624 | MUC16 | rs11878666 | 0.008346101 |
| SNP11 | 48510777 | OR4A47 | rs7103992 | 0.008363343 |
| SNP19 | 58439306 | ZNF418 | rs7253514 | 0.008389573 |
| SNP3 | 81698130 | GBE1 | rs2229519 | 0.008428248 |
| SNP1 | 200959302 | KIF21B | rs2297911 | 0.008467606 |
| SNP19 | 57065189 | ZFP28 | rs145011 | 0.008491088 |
| SNP17 | 18880268 | SLC5A10 | rs2472715 | 0.008491427 |
| SNP19 | 52497943 | ZNF615 | rs10500311 | 0.008509155 |
| SNP8 | 1616718 | DLGAP2 | rs2235112 | 0.008524428 |
| SNP11 | 128781339 | KCNJ5 | rs6590357 | 0.008668459 |
| SNP9 | 113308525 | SVEP1 | rs41305611 | 0.00868478 |
| SNP1 | 110233147 | GSTM1 | rs1056806 | 0.008723802 |
| SNP10 | 27434483 | YME1L1 | rs2274634 | 0.008754563 |
| SNP15 | 101464915 | LRRK1 | rs11630691 | 0.008809312 |
| SNP11 | 44296946 | ALX4 | rs11037928 | 0.008811472 |
| SNP11 | 44296946 | ALX4 | rs145166164 | 0.008811472 |
| SNP16 | 28997997 | LAT | rs1131543 | 0.0089579 |
| SNP1 | 38149076 | C1orf109 | rs34250208 | 0.008959237 |
| SNP4 | 1244879 | C4orf42 | rs1564508 | 0.008986849 |
| SNP11 | 72408055 | ARAP1 | rs56200889 | 0.009044609 |
| SNP5 | 96139595 | ERAP1 | rs72773968 | 0.009084142 |
| SNP12 | 10131939 | CLEC12A | rs536947 | 0.00909932 |
| SNP1 | 167095765 | DUSP27 | rs6668826 | 0.009194533 |
| SNP19 | 53100323 | | rs7250630 | 0.009209459 |
| SNP16 | 28513403 | IL27 | rs181206 | 0.009341966 |
| SNP17 | 64222164 | APOH | rs1801692 | 0.009481613 |
| SNP15 | 42565588 | TMEM87A | rs2277533 | 0.009487245 |
| SNP9 | 136907005 | BRD3 | rs467387 | 0.009696923 |
| SNP7 | 24903139 | OSBPL3 | rs17851558 | 0.009741022 |
| SNP21 | 46058060 | KRTAP10-10 | rs13051517 | 0.009751475 |
| SNP21 | 46058060 | KRTAP10-10 | rs74695406 | 0.009751475 |
| SNP1 | 203194186 | CHIT1 | rs2297950 | 0.009784441 |
| SNP20 | 278806 | ZCCHC3 | rs1057189 | 0.009809861 |
| SNP16 | 18839362 | SMG1 | rs12445870 | 0.009900558 |
| SNP10 | 64415184 | ZNF365 | rs7076156 | 0.009906283 |
| SNP1 | 9011722 | CA6 | rs3765966 | 0.009982279 |
| SNP2 | 182374534 | ITGA4 | rs1143674 | 0.010011343 |
| SNP11 | 68704028 | IGHMBP2 | rs2236654 | 0.010029929 |
| SNP20 | 13695680 | ESF1 | rs6110019 | 0.010048485 |
| SNP11 | 60893235 | CD5 | rs2229177 | 0.010059846 |
| SNP13 | 111098226 | COL4A2 | rs4103 | 0.010070856 |
| SNP1 | 25570081 | C1orf63 | rs1043879 | 0.010090782 |
| SNP4 | 76878716 | SDAD1 | rs2242471 | 0.010110981 |
| SNP8 | 2071174 | MYOM2 | rs17854780 | 0.010217498 |
| SNP5 | 7743787 | ADCY2 | rs62342477 | 0.010221027 |
| SNP14 | 105196365 | ADSSL1 | rs33958252 | 0.010231527 |
| SNP22 | 39440149 | APOBEC3F | rs5750728 | 0.010231527 |
| SNP7 | 21640361 | DNAH11 | rs10269582 | 0.010290801 |
| SNP12 | 43769228 | ADAMTS20 | rs10506226 | 0.010319794 |
| SNP12 | 43769276 | ADAMTS20 | rs10880473 | 0.010319794 |
| SNP8 | 17396380 | SLC7A2 | rs13259948 | 0.010319794 |
| SNP15 | 77241542 | RCN2 | rs15939 | 0.010356752 |
| SNP1 | 5935162 | | rs1287637 | 0.010496788 |
| SNP11 | 26587394 | MUC15 | rs11029619 | 0.010505872 |
| SNP10 | 71391538 | C10orf35 | rs1381932 | 0.010522617 |
| SNP19 | 7083629 | ZNF557 | rs966591 | 0.010596718 |
| SNP21 | 19506651 | | rs2824646 | 0.010618238 |
| SNP11 | 74705696 | NEU3 | rs544115 | 0.010666373 |
| SNP21 | 46101993 | KRTAP12-1 | rs55881656 | 0.010673292 |
| SNP6 | 84303342 | SNAP91 | rs1033655 | 0.010698417 |
| SNP5 | 149495395 | PDGFRB | rs246388 | 0.010702508 |
| SNP17 | 29857460 | RAB11FIP4 | rs61731602 | 0.010835876 |
| SNP9 | 136599146 | SARDH | rs573904 | 0.010877394 |
| SNP1 | 183105705 | LAMC1 | rs20561 | 0.010947716 |
| SNP17 | 34261831 | LYZL6 | rs9754 | 0.011142891 |
| SNP1 | 165389129 | RXRG | rs1128977 | 0.011174668 |
| SNP8 | 143427178 | TSNARE1 | rs33970858 | 0.01121255 |
| SNP8 | 143427178 | TSNARE1 | rs143680041 | 0.01121255 |
| SNP5 | 101816104 | SLCO6A1 | rs6884141 | 0.011475106 |
| SNP8 | 131921956 | ADCY8 | rs12547243 | 0.011483511 |
| SNP2 | 37241050 | HEATR5B | rs17020125 | 0.011497076 |
| SNP7 | 86542455 | KIAA1324L | rs35895874 | 0.011502922 |
| SNP19 | 49339098 | HSD17B14 | rs8110220 | 0.011588866 |
| SNP1 | 245861571 | KIF26B | rs3205034 | 0.011649991 |
| SNP5 | 135692575 | TRPC7 | rs2546661 | 0.011665511 |
| SNP11 | 26568966 | ANO3 | rs2663168 | 0.011687728 |
| SNP4 | 944210 | TMEM175 | rs34884217 | 0.011743436 |
| SNP1 | 196887457 | CFHR4 | rs10494745 | 0.011867036 |
| SNP19 | 52937339 | ZNF534 | rs17780173 | 0.01188682 |
| SNP11 | 5529152 | UBQLN3 | rs2234455 | 0.011917242 |
| SNP11 | 116692334 | APOA4 | rs5104 | 0.011997619 |
| SNP16 | 31004169 | STX1B | rs17855121 | 0.012105718 |
| SNP5 | 16478200 | FAM134B | rs162848 | 0.012108797 |
| SNP20 | 39832628 | ZHX3 | rs17265513 | 0.012184911 |
| SNP4 | 77817874 | ANKRD56 | rs2703130 | 0.012244887 |
| SNP3 | 20113830 | KAT2B | rs3021408 | 0.012290625 |
| SNP10 | 104596981 | CYP17A1 | rs6162 | 0.012379783 |
| SNP11 | 1246941 | MUC5B | rs2672785 | 0.012425878 |
| SNP8 | 92033502 | TMEM55A | rs34503257 | 0.012486642 |
| SNP3 | 179137273 | GNB4 | rs1362650 | 0.012489997 |
| SNP14 | 106963101 | | rs7141669 | 0.01269184 |
| SNP5 | 35876274 | IL7R | rs3194051 | 0.01272551 |
| SNP5 | 34824555 | RAI14 | rs1048944 | 0.012727404 |
| SNP16 | 778024 | HAGHL | rs1406814 | 0.012756958 |
| SNP16 | 1250389 | CACNA1H | rs36117280 | 0.012839331 |
| SNP1 | 183909717 | GLT25D2 | rs2296713 | 0.01299474 |
| SNP4 | 69962610 | UGT2B7 | rs28365063 | 0.013162921 |
| SNP11 | 4945196 | OR51G1 | rs1378739 | 0.01329499 |
| SNP19 | 1119963 | SBNO2 | rs2074921 | 0.013327056 |
| SNP9 | 22029445 | | rs10965215 | 0.013380626 |
| SNP6 | 84234144 | PRSS35 | rs3812141 | 0.013473356 |
| SNP14 | 88651962 | KCNK10 | rs17762463 | 0.013474295 |
| SNP14 | 88651962 | KCNK10 | rs75132782 | 0.013474295 |
| SNP8 | 21984650 | HR | rs12675745 | 0.013480182 |
| SNP10 | 126715629 | CTBP2 | rs3781409 | 0.013480396 |
| SNP2 | 128934400 | UGGT1 | rs2290111 | 0.013554135 |
| SNP15 | 50279662 | ATP8B4 | rs16963151 | 0.013578214 |
| SNP9 | 18777368 | ADAMTSL1 | rs7033684 | 0.01364648 |
| SNP4 | 951947 | TMEM175 | rs34311866 | 0.013739875 |
| SNP9 | 114359624 | PTGR1 | rs1053959 | 0.013978747 |
| SNP11 | 125871715 | CDON | rs12274923 | 0.013997882 |
| SNP19 | 11224265 | LDLR | rs5930 | 0.014009687 |
| SNP19 | 49513273 | RUVBL2 | rs1062708 | 0.01404655 |
| SNP17 | 6683684 | FBXO39 | rs16956264 | 0.014166655 |
| SNP4 | 1988193 | | rs1077020 | 0.01417821 |
| SNP9 | 134136248 | FAM78A | rs9966 | 0.014216425 |
| SNP3 | 56763525 | ARHGEF3 | rs1009118 | 0.014246286 |
| SNP3 | 56763619 | ARHGEF3 | rs6765444 | 0.014246286 |
| SNP17 | 7951819 | ALOX15B | rs4792147 | 0.014281237 |
| SNP2 | 216229692 | FN1 | rs11651 | 0.014361291 |
| SNP16 | 4748825 | ANKS3 | rs841214 | 0.014409515 |
| SNP4 | 9909923 | SLC2A9 | rs2280205 | 0.014610656 |
| SNP12 | 124979789 | NCOR2 | rs872225 | 0.014654987 |
| SNP3 | 108639384 | GUCA1C | rs6804162 | 0.014670894 |
| SNP7 | 4213877 | SDK1 | rs601424 | 0.01471174 |
| SNP7 | 4213877 | SDK1 | rs190314623 | 0.01471174 |
| SNP17 | 71232990 | C17orf80 | rs61729639 | 0.014731335 |
| SNP19 | 18375815 | KIAA1683 | rs8103177 | 0.014757229 |
| SNP19 | 18375846 | KIAA1683 | rs2277921 | 0.014757229 |
| SNP17 | 7462969 | TNFSF13 | rs3803800 | 0.014788987 |
| SNP19 | 20003109 | ZNF253 | rs8106559 | 0.014836861 |
| SNP9 | 130984809 | DNM1 | rs35048348 | 0.014990554 |
| SNP18 | 74962645 | GALR1 | rs5374 | 0.015113535 |
| SNP14 | 21928383 | | rs1045811 | 0.015134059 |
| SNP12 | 70928616 | PTPRB | rs3752703 | 0.015183501 |
| SNP19 | 12774208 | MAN2B1 | rs1054486 | 0.015204686 |
| SNP19 | 12774208 | MAN2B1 | rs138005511 | 0.015204686 |
| SNP17 | 64881075 | CACNG5 | rs2286678 | 0.015482615 |
| SNP10 | 78944590 | KCNMA1 | rs1131824 | 0.015543939 |
| SNP16 | 3339435 | ZNF263 | rs220379 | 0.01566307 |
| SNP20 | 62200860 | RP4-697K14.7 | rs1757752 | 0.015736444 |
| SNP15 | 94841691 | MCTP2 | rs61737195 | 0.015773255 |
| SNP17 | 18874720 | FAM83G | rs916823 | 0.015791012 |
| SNP13 | 28009851 | MTIF3 | rs7669 | 0.015810577 |
| SNP10 | 29812602 | SVIL | rs7070678 | 0.015894594 |
| SNP13 | 41373254 | SLC25A15 | rs41396747 | 0.01598341 |
| SNP13 | 41373254 | SLC25A15 | rs55943532 | 0.01598341 |
| SNP1 | 3301721 | PRDM16 | rs2282198 | 0.016038227 |
| SNP19 | 5866724 | FUT5 | rs4807054 | 0.016151029 |
| SNP5 | 13754394 | DNAH5 | rs2401809 | 0.016192503 |
| SNP10 | 96066341 | PLCE1 | rs2274223 | 0.01631179 |
| SNP1 | 227182575 | CDC42BPA | rs1045247 | 0.01633809 |
| SNP6 | 158735087 | TULP4 | rs9364951 | 0.016347375 |
| SNP6 | 158735090 | TULP4 | rs7756620 | 0.016347375 |
| SNP22 | 29837537 | RFPL1 | rs3804076 | 0.016408608 |
| SNP10 | 95454681 | FRA10AC1 | rs2275438 | 0.016453294 |
| SNP10 | 95454681 | FRA10AC1 | rs137880326 | 0.016453294 |
| SNP3 | 45637439 | LIMD1 | rs267236 | 0.01654474 |
| SNP3 | 45637439 | LIMD1 | rs145508440 | 0.01654474 |
| SNP4 | 42639186 | | rs898500 | 0.016564857 |
| SNP17 | 78319380 | RNF213 | rs4890012 | 0.016567432 |
| SNP19 | 56735015 | ZSCAN5A | rs4801692 | 0.016603154 |
| SNP1 | 20945055 | CDA | rs1048977 | 0.016649871 |
| SNP11 | 17496516 | ABCC8 | rs1048099 | 0.016655027 |
| SNP9 | 21816758 | MTAP | rs7023954 | 0.016718507 |
| SNP2 | 3673648 | COLEC11 | rs10170348 | 0.016754337 |
| SNP16 | 85100976 | KIAA0513 | rs4783121 | 0.016806689 |
| SNP20 | 56190634 | ZBP1 | rs2073145 | 0.016817511 |
| SNP1 | 40150156 | HPCAL4 | rs9662128 | 0.016871646 |
| SNP19 | 8028544 | ELAVL1 | rs14394 | 0.016915282 |
| SNP10 | 60588553 | BICC1 | rs4948550 | 0.017026173 |
| SNP10 | 60588553 | BICC1 | rs145960427 | 0.017026173 |
| SNP19 | 44047550 | XRCC1 | rs3547 | 0.017060668 |
| SNP1 | 19950062 | C1orf151 | rs1737428 | 0.017218364 |
| SNP7 | 2649777 | IQCE | rs1061566 | 0.017260751 |
| SNP5 | 132535046 | FSTL4 | rs3749817 | 0.017311808 |
| SNP21 | 15481365 | LIPI | rs7278737 | 0.017334994 |
| SNP1 | 46806550 | NSUN4 | rs41293277 | 0.017432745 |
| SNP4 | 177137988 | ASB5 | rs6827525 | 0.017528176 |
| SNP1 | 151110832 | SEMA6C | rs72708441 | 0.017618186 |
| SNP17 | 11833287 | DNAH9 | rs2286303 | 0.017622826 |
| SNP1 | 246930564 | SCCPDH | rs7779 | 0.017676327 |
| SNP19 | 2994921 | TLE6 | rs34551565 | 0.017801519 |
| SNP4 | 123814308 | NUDT6 | rs1048201 | 0.017889959 |
| SNP1 | 205819039 | PM20D1 | rs11540016 | 0.017985849 |
| SNP1 | 205819104 | PM20D1 | rs11540014 | 0.017985849 |
| SNP2 | 232952286 | DIS3L2 | rs11887184 | 0.018006433 |
| SNP22 | 44965320 | RP3-474I12.8 | rs138593 | 0.018021588 |
| SNP13 | 46103935 | COG3 | rs2274285 | 0.018132721 |
| SNP6 | 62390916 | KHDRBS2 | rs1204114 | 0.018277864 |
| SNP11 | 74419378 | CHRDL2 | rs7948433 | 0.018319289 |
| SNP11 | 74419378 | CHRDL2 | rs150623836 | 0.018319289 |
| SNP7 | 33282577 | | rs7793096 | 0.018379952 |
| SNP19 | 15807884 | CYP4F12 | rs593818 | 0.018385481 |
| SNP6 | 24596478 | KIAA0319 | rs4576240 | 0.018640807 |
| SNP19 | 54849942 | LILRA4 | rs2241384 | 0.018823666 |
| SNP11 | 75428958 | MOGAT2 | rs554202 | 0.018909584 |
| SNP11 | 75428958 | MOGAT2 | rs183052760 | 0.018909584 |
| SNP1 | 27676925 | SYTL1 | rs6702341 | 0.019031688 |
| SNP10 | 45496120 | C10orf25 | rs12269028 | 0.019088088 |
| SNP2 | 18113508 | KCNS3 | rs3747516 | 0.019141858 |
| SNP19 | 49376584 | PPP1R15A | rs3786734 | 0.019160752 |
| SNP21 | 22746187 | NCAM2 | rs232518 | 0.019167264 |
| SNP13 | 31729729 | HSPH1 | rs1047086 | 0.0191735 |
| SNP19 | 33608733 | GPATCH1 | rs2287681 | 0.019179955 |
| SNP3 | 13612936 | FBLN2 | rs3732666 | 0.01918863 |
| SNP2 | 84800605 | DNAH6 | rs11891970 | 0.019214343 |
| SNP19 | 2917612 | ZNF57 | rs10410539 | 0.019245125 |
| SNP19 | 2917612 | ZNF57 | rs184508498 | 0.019245125 |
| SNP1 | 222923351 | FAM177B | rs6683071 | 0.019261623 |
| SNP1 | 175092674 | TNN | rs2285215 | 0.0193247 |
| SNP1 | 175092674 | TNN | rs78387535 | 0.0193247 |
| SNP6 | 32411646 | HLA-DRA | rs58547911 | 0.019363797 |
| SNP6 | 32411646 | HLA-DRA | rs7192 | 0.019363797 |
| SNP6 | 32411646 | HLA-DRA | rs114578736 | 0.019363797 |
| SNP9 | 130165995 | SLC2A8 | rs1138740 | 0.019448949 |
| SNP7 | 43519337 | HECW1 | rs2304327 | 0.019479208 |
| SNP19 | 50484234 | VRK3 | rs16981617 | 0.019715328 |
| SNP2 | 182399097 | ITGA4 | rs7562325 | 0.019739211 |
| SNP2 | 182399097 | ITGA4 | rs188628120 | 0.019739211 |
| SNP7 | 27204732 | HOXA9 | rs35355140 | 0.019772763 |
| SNP6 | 130761957 | TMEM200A | rs3813359 | 0.019864065 |
| SNP2 | 139429068 | NXPH2 | rs10803570 | 0.019964215 |
| SNP3 | 13670481 | FBLN2 | rs2242023 | 0.02000571 |
| SNP22 | 29414001 | | rs2347790 | 0.02022573 |
| SNP18 | 9117867 | NDUFV2 | rs906807 | 0.020301635 |
| SNP21 | 45959195 | KRTAP10-1 | rs233316 | 0.02031471 |
| SNP13 | 50205025 | ARL11 | rs3803185 | 0.020382149 |
| SNP14 | 20852770 | TEP1 | rs1713458 | 0.020438684 |
| SNP12 | 132562126 | EP400 | rs73164912 | 0.020623142 |
| SNP17 | 9395231 | STX8 | rs9893664 | 0.020641581 |
| SNP7 | 48450157 | ABCA13 | rs17548783 | 0.02065968 |
| SNP4 | 146653620 | C4orf51 | rs10008599 | 0.02070254 |
| SNP11 | 60102384 | MS4A6E | rs2304935 | 0.020749184 |
| SNP11 | 60102396 | MS4A6E | rs2304934 | 0.020749184 |
| SNP11 | 60102507 | MS4A6E | rs2304933 | 0.020749184 |
| SNP12 | 72307616 | TBC1D15 | rs3759171 | 0.020901955 |
| SNP16 | 10862964 | | rs11716 | 0.020966375 |
| SNP16 | 10862964 | | rs2233543 | 0.020966375 |
| SNP12 | 69986788 | CCT2 | rs1043434 | 0.020994305 |
| SNP7 | 135048804 | CNOT4 | rs3812265 | 0.020996355 |
| SNP5 | 149907602 | NDST1 | rs2273234 | 0.021069044 |
| SNP14 | 74086415 | ACOT6 | rs17782052 | 0.021229463 |
| SNP9 | 113169630 | SVEP1 | rs7046213 | 0.021244266 |
| SNP9 | 113169630 | SVEP1 | rs71492888 | 0.021244266 |
| SNP9 | 113169630 | SVEP1 | rs7030192 | 0.021244266 |
| SNP8 | 27373865 | EPHX2 | rs751141 | 0.02129326 |
| SNP15 | 89450587 | MFGE8 | rs1878326 | 0.021369704 |
| SNP6 | 10989942 | ELOVL2 | rs12195587 | 0.021379712 |
| SNP3 | 151046308 | P2RY13 | rs1466684 | 0.021500031 |
| SNP3 | 38271881 | OXSR1 | rs6599079 | 0.021611207 |
| SNP12 | 547683 | CCDC77 | rs735295 | 0.021743298 |
| SNP12 | 547683 | CCDC77 | rs143410673 | 0.021743298 |
| SNP5 | 180218668 | MGAT1 | rs634501 | 0.021743298 |
| SNP14 | 68257352 | ZFYVE26 | rs17192170 | 0.021866534 |
| SNP17 | 3397702 | ASPA | rs12948217 | 0.02218904 |
| SNP5 | 95152313 | GLRX | rs4561 | 0.022217082 |
| SNP4 | 155507590 | FGA | rs6050 | 0.022251978 |
| SNP22 | 22569554 | | rs6001413 | 0.022294158 |
| SNP14 | 20528207 | OR4L1 | rs1958715 | 0.022341961 |
| SNP4 | 39116911 | KLHL5 | rs3733276 | 0.022341961 |
| SNP4 | 39116911 | KLHL5 | rs146066882 | 0.022341961 |
| SNP8 | 107754473 | AC090579.1 | rs1681904 | 0.022455197 |
| SNP4 | 115749005 | NDST4 | rs6843860 | 0.02248723 |
| SNP1 | 156526387 | IQGAP3 | rs11264498 | 0.022506269 |
| SNP19 | 36595436 | WDR62 | rs1008328 | 0.02252746 |
| SNP2 | 230668858 | TRIP12 | rs13018957 | 0.022533523 |
| SNP6 | 31496925 | MCCD1 | rs3093983 | 0.022548942 |
| SNP6 | 31496925 | MCCD1 | rs114800489 | 0.022548942 |
| SNP6 | 31506624 | DDX39B | rs1129640 | 0.022548942 |
| SNP6 | 31506624 | DDX39B | rs116828464 | 0.022548942 |
| SNP12 | 10206925 | CLEC9A | rs7315231 | 0.022640895 |
| SNP12 | 49176582 | ADCY6 | rs3730064 | 0.022703499 |
| SNP7 | 40314165 | C7orf10 | rs12669315 | 0.022854926 |
| SNP2 | 160808075 | PLA2R1 | rs3828323 | 0.023116878 |
| SNP2 | 160808075 | PLA2R1 | rs72954858 | 0.023116878 |
| SNP1 | 152552285 | LCE3D | rs512208 | 0.023160155 |
| SNP1 | 152552285 | LCE3D | rs61745411 | 0.023160155 |
| SNP19 | 44377669 | ZNF404 | rs12977303 | 0.023220458 |
| SNP6 | 39282806 | KCNK16 | rs11756091 | 0.023220458 |
| SNP2 | 107450589 | ST6GAL2 | rs3796111 | 0.023234857 |
| SNP2 | 107450589 | ST6GAL2 | rs139517922 | 0.023234857 |
| SNP19 | 16040292 | CYP4F11 | rs3765070 | 0.023339502 |
| SNP5 | 81549240 | ATG10 | rs1864182 | 0.023509416 |
| SNP8 | 21847855 | XPO7 | rs56062629 | 0.023561298 |
| SNP7 | 143093538 | EPHA1 | rs10952549 | 0.023576197 |
| SNP2 | 185802243 | ZNF804A | rs1366842 | 0.023659604 |
| SNP1 | 226019653 | EPHX1 | rs2292566 | 0.023788685 |
| SNP2 | 223559089 | MOGAT1 | rs1868024 | 0.023819828 |
| SNP17 | 73919521 | FBF1 | rs7219918 | 0.023859485 |
| SNP2 | 112686988 | MERTK | rs72376349 | 0.023952918 |
| SNP2 | 112686988 | MERTK | rs13027171 | 0.023952918 |
| SNP7 | 154738409 | PAXIP1 | rs3501 | 0.02396947 |
| SNP1 | 20020994 | TMCO4 | rs4515815 | 0.023991723 |
| SNP21 | 46842443 | | rs8131523 | 0.024011147 |
| SNP2 | 218674697 | TNS1 | rs918949 | 0.024118689 |
| SNP7 | 11581121 | THSD7A | rs47 | 0.024136691 |
| SNP6 | 132859609 | | rs2842899 | 0.024210779 |
| SNP17 | 3352309 | SPATA22 | rs11556563 | 0.024220658 |
| SNP10 | 134013906 | DPYSL4 | rs56326856 | 0.024303725 |
| SNP2 | 75915035 | C2orf3 | rs1803196 | 0.024334425 |
| SNP14 | 94642417 | PPP4R4 | rs34520961 | 0.024380899 |
| SNP2 | 26647277 | C2orf39 | rs7423300 | 0.024380899 |
| SNP17 | 27284443 | SEZ6 | rs12941884 | 0.024547534 |
| SNP1 | 201038687 | CACNA1S | rs7415038 | 0.024619217 |
| SNP17 | 76471366 | DNAH17 | rs61746438 | 0.024889062 |
| SNP6 | 155606325 | TFB1M | rs324356 | 0.024905707 |
| SNP11 | 8009241 | EIF3F | rs12421289 | 0.024987165 |
| SNP10 | 88414570 | OPN4 | rs11202106 | 0.025229247 |
| SNP12 | 50232169 | BCDIN3D | rs11169172 | 0.025319398 |
| SNP10 | 45478092 | RASSF4 | rs870957 | 0.025324177 |
| SNP1 | 226055595 | TMEM63A | rs2292564 | 0.025438493 |
| SNP9 | 334337 | DOCK8 | rs10970979 | 0.025573811 |
| SNP15 | 45981317 | SQRDL | rs10643 | 0.025914119 |
| SNP10 | 73111408 | SLC29A3 | rs780668 | 0.026077597 |
| SNP14 | 77492891 | IRF2BPL | rs879027 | 0.026202032 |
| SNP4 | 139140494 | SLC7A11 | rs6838248 | 0.026203593 |
| SNP19 | 53344701 | ZNF468 | rs10420793 | 0.026208514 |
| SNP16 | 23226787 | SCNN1G | rs5723 | 0.026347047 |
| SNP4 | 160277276 | RAPGEF2 | rs1135004 | 0.026499137 |
| SNP16 | 3304762 | MEFV | rs224225 | 0.026569696 |
| SNP10 | 126089434 | OAT | rs11461 | 0.026578071 |
| SNP1 | 17739586 | RCC2 | rs942457 | 0.026667115 |
| SNP1 | 159505101 | OR10J5 | rs35393723 | 0.026767765 |
| SNP1 | 204589066 | LRRN2 | rs36012907 | 0.026919577 |
| SNP1 | 204589066 | LRRN2 | rs114877011 | 0.026919577 |
| SNP1 | 204966428 | NFASC | rs2802808 | 0.027050826 |
| SNP19 | 3281298 | CELF5 | rs36030381 | 0.027062616 |
| SNP3 | 36779707 | DCLK3 | rs3796187 | 0.02709981 |
| SNP2 | 111907691 | BCL2L11 | rs724710 | 0.027192632 |
| SNP5 | 178294059 | ZNF354B | rs11952817 | 0.027199758 |
| SNP5 | 178294059 | ZNF354B | rs11955074 | 0.027199758 |
| SNP5 | 178294059 | ZNF354B | rs113633676 | 0.027199758 |
| SNP6 | 70071173 | BAI3 | rs913543 | 0.027306701 |
| SNP21 | 43863521 | UBASH3A | rs868092 | 0.027485964 |
| SNP15 | 33872177 | RYR3 | rs674155 | 0.027817195 |
| SNP2 | 242610738 | ATG4B | rs11538896 | 0.027933973 |
| SNP4 | 47560015 | ATP10D | rs34169638 | 0.028117764 |
| SNP6 | 31113052 | CCHCR1 | rs3094225 | 0.028437274 |
| SNP6 | 31113052 | CCHCR1 | rs113704463 | 0.028437274 |
| SNP7 | 5342474 | SLC29A4 | rs149588788 | 0.028479516 |
| SNP15 | 41829230 | RPAP1 | rs721772 | 0.028647715 |
| SNP7 | 30695202 | CRHR2 | rs2240403 | 0.028738665 |
| SNP6 | 143816859 | FUCA2 | rs8161 | 0.028917108 |
| SNP16 | 65022114 | CDH11 | rs28216 | 0.029030788 |
| SNP6 | 146125793 | FBXO30 | rs3811102 | 0.029339639 |
| SNP10 | 7759595 | ITIH2 | rs7072478 | 0.029548798 |
| SNP1 | 161476204 | FCGR2A | rs9427397 | 0.029598286 |
| SNP1 | 161476204 | FCGR2A | rs9427398 | 0.029598286 |
| SNP6 | 158611969 | | rs9459642 | 0.029598286 |
| SNP16 | 89256695 | CDH15 | rs72819366 | 0.029754118 |
| SNP15 | 22960868 | CYFIP1 | rs2289818 | 0.029938303 |
| SNP19 | 3753769 | APBA3 | rs3746120 | 0.030114455 |
| SNP6 | 70733547 | COL19A1 | rs2273426 | 0.030294729 |
| SNP12 | 21028208 | SLCO1B3 | rs60140950 | 0.030301463 |
| SNP12 | 130839165 | PIWIL1 | rs10848087 | 0.031208191 |
| SNP5 | 38955796 | RICTOR | rs2043112 | 0.031302401 |
| SNP1 | 228560700 | OBSCN | rs512253 | 0.031379657 |
| SNP1 | 158584091 | SPTA1 | rs952094 | 0.031555571 |
| SNP11 | 96116444 | CCDC82 | rs10831519 | 0.031589157 |
| SNP18 | 29625685 | RNF125 | rs61749945 | 0.031599228 |
| SNP5 | 21121275 | | rs248205 | 0.031635027 |
| SNP3 | 124802888 | SLC12A8 | rs2981482 | 0.031702266 |
| SNP13 | 42729441 | DGKH | rs34298263 | 0.03172728 |
| SNP13 | 47409034 | HTR2A | rs6314 | 0.031733783 |
| SNP3 | 187088926 | RTP4 | rs1533594 | 0.031877657 |
| SNP3 | 123634046 | CCDC14 | rs2700373 | 0.031898881 |
| SNP22 | 50470516 | TTLL8 | rs11101958 | 0.031910853 |
| SNP10 | 106022789 | GSTO1 | rs4925 | 0.031962775 |
| SNP10 | 106022789 | GSTO1 | rs111637087 | 0.031962775 |
| SNP19 | 15133787 | CCDC105 | rs10424547 | 0.032149156 |
| SNP20 | 23016970 | SSTR4 | rs3746726 | 0.032159315 |
| SNP20 | 23017044 | SSTR4 | rs3746728 | 0.032159315 |
| SNP11 | 82643760 | C11orf82 | rs7947780 | 0.032405872 |
| SNP17 | 18055229 | MYO15A | rs854772 | 0.032691633 |
| SNP16 | 4934564 | PPL | rs1049207 | 0.032750518 |
| SNP4 | 39216221 | WDR19 | rs2167494 | 0.032844163 |
| SNP22 | 23040767 | | rs5751516 | 0.032876992 |
| SNP15 | 100333075 | AC090825.1 | rs11634114 | 0.032935324 |
| SNP15 | 100333075 | AC090825.1 | rs77574659 | 0.032935324 |
| SNP15 | 100333075 | AC090825.1 | rs7178437 | 0.032935324 |
| SNP15 | 100333075 | AC090825.1 | rs52815341 | 0.032935324 |
| SNP1 | 47726087 | STIL | rs13376679 | 0.032938506 |
| SNP16 | 27373972 | IL4R | rs2234900 | 0.033457408 |
| SNP22 | 32589090 | RFPL2 | rs8135276 | 0.033718036 |
| SNP12 | 6562285 | TAPBPL | rs2041384 | 0.034003612 |
| SNP6 | 36976637 | FGD2 | rs831510 | 0.03409661 |
| SNP22 | 44368204 | SAMM50 | rs3177036 | 0.034171034 |
| SNP3 | 113320477 | SIDT1 | rs33990195 | 0.034273662 |
| SNP1 | 160011512 | KCNJ10 | rs1130183 | 0.034292693 |
| SNP8 | 26240646 | | rs1055479 | 0.034349587 |
| SNP2 | 182981968 | PPP1R1C | rs1882212 | 0.034439277 |
| SNP6 | 31473561 | MICB | rs1065075 | 0.034477118 |
| SNP6 | 31473561 | MICB | rs149741144 | 0.034477118 |
| SNP12 | 124957627 | NCOR2 | rs10846679 | 0.03454138 |
| SNP18 | 47812587 | CXXC1 | rs11555886 | 0.034655783 |
| SNP12 | 116403920 | MED13L | rs3088260 | 0.034672808 |
| SNP19 | 14273641 | LPHN1 | rs3745462 | 0.034885543 |
| SNP1 | 176809368 | PAPPA2 | rs12118034 | 0.034928208 |
| SNP19 | 55739689 | TMEM86B | rs10413828 | 0.035178748 |
| SNP3 | 187419816 | RTP2 | rs61754877 | 0.035287406 |
| SNP3 | 56763328 | ARHGEF3 | rs1565377 | 0.035289518 |
| SNP8 | 69046409 | PREX2 | rs3793379 | 0.035331621 |
| SNP16 | 71509685 | ZNF19 | rs2288489 | 0.035393547 |
| SNP4 | 5838513 | CRMP1 | rs12331 | 0.035717146 |
| SNP10 | 3207632 | PITRM1 | rs3814596 | 0.03571717 |
| SNP6 | 116575116 | TSPYL4 | rs17524614 | 0.035761971 |
| SNP15 | 78390909 | SH2D7 | rs12593575 | 0.035999141 |
| SNP11 | 72946204 | P2RY2 | rs1626154 | 0.036037295 |
| SNP11 | 72946204 | P2RY2 | rs113741989 | 0.036037295 |
| SNP4 | 946226 | TMEM175 | rs11552301 | 0.036327848 |
| SNP11 | 5629607 | TRIM6 | rs10769121 | 0.036345537 |
| SNP7 | 71134983 | WBSCR17 | rs35612272 | 0.03637682 |
| SNP1 | 29475394 | SRSF4 | rs2230677 | 0.036405197 |
| SNP18 | 5956238 | L3MBTL4 | rs3737353 | 0.03641053 |
| SNP19 | 11891003 | ZNF441 | rs799193 | 0.036443941 |
| SNP14 | 21215997 | EDDM3A | rs11847654 | 0.036699775 |
| SNP17 | 3631241 | ITGAE | rs2976230 | 0.036869574 |
| SNP18 | 2595443 | NDC80 | rs12456560 | 0.036871949 |
| SNP7 | 139724555 | PARP12 | rs3735352 | 0.036906909 |
| SNP7 | 139724555 | PARP12 | rs35360282 | 0.036906909 |
| SNP10 | 124610027 | FAM24B | rs1891110 | 0.036922644 |
| SNP19 | 756985 | C19orf21 | rs8110536 | 0.036958137 |
| SNP20 | 62194030 | RP4-697K14.7 | rs3810483 | 0.037242555 |
| SNP20 | 62194030 | RP4-697K14.7 | rs142656726 | 0.037242555 |
| SNP2 | 191161622 | HIBCH | rs1058180 | 0.037443012 |
| SNP20 | 55209257 | TFAP2C | rs35023929 | 0.037731434 |
| SNP6 | 168457966 | FRMD1 | rs3734899 | 0.037792698 |
| SNP16 | 88620195 | C16orf85 | rs2879897 | 0.037899589 |
| SNP11 | 6519642 | DNHD1 | rs11604149 | 0.03810221 |
| SNP11 | 6520015 | DNHD1 | rs11040899 | 0.03810221 |
| SNP5 | 137717213 | KDM3B | rs10073922 | 0.038160331 |
| SNP1 | 247769752 | OR2G3 | rs61730407 | 0.038210695 |
| SNP20 | 49620783 | KCNG1 | rs1054268 | 0.038797 |
| SNP16 | 88552370 | ZFPM1 | rs3751673 | 0.038815987 |
| SNP14 | 101507727 | | rs12894467 | 0.038957189 |
| SNP17 | 3657159 | ITGAE | rs2272606 | 0.039223784 |
| SNP2 | 65296798 | CEP68 | rs7572857 | 0.039357884 |
| SNP11 | 45832509 | SLC35C1 | rs7130656 | 0.039578621 |
| SNP19 | 2353148 | | rs10402284 | 0.039578621 |
| SNP1 | 2938265 | ACTRT2 | rs4576609 | 0.039742621 |
| SNP1 | 2938265 | ACTRT2 | rs141283101 | 0.039742621 |
| SNP1 | 2938697 | ACTRT2 | rs3795262 | 0.039742621 |
| SNP12 | 100657464 | DEPDC4 | rs7307415 | 0.03987881 |
| SNP18 | 2707619 | SMCHD1 | rs2276092 | 0.03992941 |
| SNP3 | 46713457 | ALS2CL | rs7625303 | 0.040040087 |
| SNP3 | 46713457 | ALS2CL | rs74585038 | 0.040040087 |
| SNP19 | 42083849 | CEACAM21 | rs714106 | 0.040561657 |
| SNP17 | 4689313 | VMO1 | rs2279961 | 0.040768744 |
| SNP2 | 68385097 | PNO1 | rs2044693 | 0.041243306 |
| SNP20 | 17475217 | BFSP1 | rs6136118 | 0.041411897 |
| SNP2 | 120060082 | C2orf76 | rs1052500 | 0.04142461 |
| SNP8 | 82606782 | SLC10A5 | rs2955006 | 0.041651754 |
| SNP17 | 19644472 | ALDH3A1 | rs2072330 | 0.041669693 |
| SNP17 | 9846521 | GAS7 | rs17339499 | 0.042086421 |
| SNP14 | 106805278 | | rs61739320 | 0.042097096 |
| SNP2 | 227892720 | COL4A4 | rs2229813 | 0.042479496 |
| SNP12 | 520947 | CCDC77 | rs4980895 | 0.04260731 |
| SNP2 | 47301029 | TTC7A | rs3739099 | 0.04260731 |
| SNP3 | 167217960 | WDR49 | rs13061106 | 0.042636222 |
| SNP17 | 79637367 | CCDC137 | rs7226091 | 0.043042991 |
| SNP12 | 12240199 | BCL2L14 | rs879732 | 0.043094016 |
| SNP4 | 103964529 | NHEDC2 | rs2276976 | 0.043341898 |
| SNP7 | 1976457 | MAD1L1 | rs1801368 | 0.043396197 |
| SNP7 | 100551106 | | rs62483697 | 0.04360685 |
| SNP2 | 241713646 | KIF1A | rs1063353 | 0.043651376 |
| SNP7 | 115897392 | TES | rs4710 | 0.0440866 |
| SNP6 | 151121915 | PLEKHG1 | rs2073061 | 0.044106713 |
| SNP16 | 1250559 | CACNA1H | rs8044363 | 0.044145447 |
| SNP20 | 10620275 | JAG1 | rs1051421 | 0.044166869 |
| SNP19 | 10659659 | ATG4D | rs2304165 | 0.044588993 |
| SNP4 | 123229132 | KIAA1109 | rs7688384 | 0.044873969 |
| SNP7 | 83634713 | SEMA3A | rs7804122 | 0.045357008 |
| SNP17 | 61607708 | KCNH6 | rs7221517 | 0.045463043 |
| SNP19 | 54080067 | ZNF331 | rs61744130 | 0.045524608 |
| SNP11 | 2188238 | TH | rs6357 | 0.046107526 |
| SNP10 | 103340056 | POLL | rs3730477 | 0.046197088 |
| SNP11 | 4928841 | OR51A7 | rs7108225 | 0.046313014 |
| SNP11 | 4928841 | OR51A7 | rs141919327 | 0.046313014 |
| SNP12 | 99640428 | ANKS1B | rs3751323 | 0.04636905 |
| SNP16 | 5064917 | SEC14L5 | rs2012649 | 0.046431463 |
| SNP2 | 226273689 | KIAA1486 | rs2048936 | 0.046567384 |
| SNP12 | 114282496 | RBM19 | rs2075387 | 0.04659618 |
| SNP12 | 114282496 | RBM19 | rs150329912 | 0.04659618 |
| SNP1 | 168698173 | DPT | rs1052591 | 0.046697906 |
| SNP11 | 17542439 | USH1C | rs2240487 | 0.046782839 |
| SNP13 | 37580139 | EXOSC8 | rs1127446 | 0.046786943 |
| SNP17 | 74383475 | SPHK1 | rs3744037 | 0.046801815 |
| SNP16 | 49670390 | ZNF423 | rs3803667 | 0.047192206 |
| SNP5 | 149583300 | SLC6A7 | rs2240793 | 0.047195781 |
| SNP5 | 169127097 | DOCK2 | rs2112703 | 0.047325261 |
| SNP4 | 66197804 | EPHA5 | rs7349683 | 0.047442598 |
| SNP5 | 34757666 | RAI14 | rs17521570 | 0.047611012 |
| SNP1 | 64097432 | PGM1 | rs1126728 | 0.047727505 |
| SNP9 | 116800 | FOXD4 | rs79220013 | 0.047875113 |
| SNP7 | 56149362 | PHKG1 | rs11238393 | 0.047900321 |
| SNP17 | 48917333 | WFIKKN2 | rs9675120 | 0.048486034 |
| SNP7 | 44047066 | SPDYE1 | rs78424385 | 0.048606943 |
| SNP7 | 44047066 | SPDYE1 | rs79154079 | 0.048606943 |
| SNP7 | 44047066 | SPDYE1 | rs149337537 | 0.048606943 |
| SNP2 | 219140288 | TMBIM1 | rs2292550 | 0.048648483 |
| SNP2 | 159954175 | TANC1 | rs34588551 | 0.048672419 |
| SNP16 | 71807232 | AP1G1 | rs113235758 | 0.048874583 |
| SNP16 | 71807232 | AP1G1 | rs143490004 | 0.048874583 |
| SNP14 | 93401178 | CHGA | rs9658671 | 0.04904235 |
| SNP14 | 93401178 | CHGA | rs941581 | 0.04904235 |
| SNP19 | 10226256 | EIF3G | rs7710 | 0.04920255 |
| SNP18 | 61323012 | SERPINB3 | rs3180227 | 0.049328001 |
| SNP18 | 61323012 | SERPINB3 | rs72132327 | 0.049328001 |
| SNP15 | 34032131 | RYR3 | rs41279212 | 0.049329337 |
| SNP7 | 144098992 | NOBOX | rs727714 | 0.049821973 |
| SNP18 | 28986333 | DSG4 | rs4799570 | 0.049900042 |
| SNP17 | 42475983 | GPATCH8 | rs936019 | 0.04991367 |
| SNP17 | 4802329 | CHRNE | rs33978919 | 0.04999028 |
| SNP15 | 75979782 | CSPG4 | rs8030131 | 0.050085161 |
| SNP15 | 75979782 | CSPG4 | rs147134901 | 0.050085161 |
| SNP10 | 129249676 | DOCK1 | rs2229604 | 0.050339376 |
| SNP15 | 85383145 | ALPK3 | rs3803403 | 0.050407947 |
| SNP11 | 10503756 | AMPD3 | rs16907852 | 0.050446529 |
| SNP17 | 55183203 | AKAP1 | rs2230772 | 0.050630849 |
| SNP5 | 150891733 | FAT2 | rs2304028 | 0.050738683 |
| SNP19 | 8398006 | KANK3 | rs7249069 | 0.050951445 |
| SNP19 | 47278859 | SLC1A5 | rs3027961 | 0.050981679 |
| SNP22 | 21167787 | PI4KA | rs165854 | 0.051066272 |
| SNP6 | 160201295 | | rs2295901 | 0.051399774 |
| SNP14 | 23236524 | OXA1L | rs8572 | 0.051473656 |
| SNP1 | 2488153 | TNFRSF14 | rs4870 | 0.051896976 |
| SNP2 | 42990336 | OXER1 | rs1992286 | 0.051956544 |
| SNP17 | 71239087 | C17orf80 | rs1566290 | 0.052286311 |
| SNP4 | 3444593 | HGFAC | rs2073504 | 0.052344927 |
| SNP15 | 45047573 | TRIM69 | rs3100139 | 0.052647965 |
| SNP20 | 1617069 | SIRPG | rs2277761 | 0.052752644 |
| SNP9 | 139371786 | SEC16A | rs6560632 | 0.052869029 |
| SNP7 | 45113170 | CCM2 | rs2289367 | 0.053128217 |
| SNP7 | 45113170 | CCM2 | rs11552375 | 0.053128217 |
| SNP8 | 144689146 | PYCRL | rs2242089 | 0.053432321 |
| SNP7 | 44579180 | NPC1L1 | rs2072183 | 0.053473275 |
| SNP14 | 106471388 | | rs34874585 | 0.053500469 |
| SNP19 | 45028169 | | rs7260180 | 0.053583999 |
| SNP4 | 47525054 | ATP10D | rs7683838 | 0.053606435 |
| SNP6 | 32427748 | | rs9268831 | 0.053617903 |
| SNP6 | 32427748 | | rs115543118 | 0.053617903 |
| SNP15 | 86287910 | | rs11073517 | 0.053715173 |
| SNP18 | 20577669 | RBBP8 | rs17852769 | 0.053923719 |
| SNP1 | 28282206 | SMPDL3B | rs3813804 | 0.054213025 |
| SNP1 | 65860687 | DNAJC6 | rs4582839 | 0.054439922 |
| SNP4 | 961373 | DGKQ | rs1377586 | 0.05458057 |
| SNP13 | 111340342 | CARS2 | rs2304767 | 0.055035665 |
| SNP10 | 75871735 | VCL | rs2131956 | 0.055054238 |
| SNP1 | 247588053 | NLRP3 | rs34298354 | 0.055128072 |
| SNP19 | 54848741 | LILRA4 | rs12985462 | 0.055295729 |
| SNP19 | 54848741 | LILRA4 | rs148423363 | 0.055295729 |
| SNP5 | 75469778 | | rs113058744 | 0.055430238 |
| SNP5 | 75469778 | | rs111577216 | 0.055430238 |
| SNP11 | 36595600 | RAG1 | rs3740955 | 0.055661933 |
| SNP1 | 3677933 | CCDC27 | rs1181883 | 0.055673298 |
| SNP1 | 27876482 | AHDC1 | rs2076457 | 0.05594528 |
| SNP2 | 235951819 | SH3BP4 | rs3795962 | 0.056006051 |
| SNP4 | 68442968 | STAP1 | rs11556614 | 0.056019545 |
| SNP4 | 88732692 | IBSP | rs1054627 | 0.05628984 |
| SNP2 | 209036712 | C2orf80 | rs10804166 | 0.056476028 |
| SNP16 | 67268008 | FHOD1 | rs77958237 | 0.056613403 |
| SNP1 | 162313735 | NOS1AP | rs3751284 | 0.05663637 |
| SNP20 | 61048549 | GATA5 | rs41305803 | 0.056740709 |
| SNP4 | 71232701 | SMR3A | rs6853742 | 0.056818381 |
| SNP10 | 323283 | DIP2C | rs3740304 | 0.05683721 |
| SNP11 | 73689104 | UCP2 | rs660339 | 0.056883558 |
| SNP3 | 149468530 | COMMD2 | rs9843784 | 0.057040474 |
| SNP5 | 148206473 | ADRB2 | rs1042714 | 0.057040474 |
| SNP19 | 33487071 | RHPN2 | rs12610600 | 0.057080191 |
| SNP19 | 33487071 | RHPN2 | rs139766778 | 0.057080191 |
| SNP1 | 206603535 | SRGAP2 | rs2297539 | 0.057413813 |
| SNP6 | 4087934 | C6orf201 | rs619483 | 0.057763073 |
| SNP12 | 4737042 | AKAP3 | rs7972737 | 0.057771216 |
| SNP8 | 10396054 | PRSS55 | rs6601483 | 0.058244841 |
| SNP13 | 113681243 | MCF2L | rs9549343 | 0.058434164 |
| SNP21 | 44846016 | SIK1 | rs3746951 | 0.058458299 |
| SNP21 | 31587793 | CLDN8 | rs686364 | 0.058465234 |
| SNP15 | 63889934 | FBXL22 | rs8035931 | 0.058500862 |
| SNP13 | 103449202 | KDELC1 | rs1047740 | 0.058537762 |
| SNP12 | 9317784 | PZP | rs2277413 | 0.058591723 |
| SNP10 | 105957714 | WDR96 | rs10883979 | 0.058633834 |
| SNP19 | 37117302 | ZNF382 | rs3108171 | 0.059239864 |
| SNP4 | 6325086 | PPP2R2C | rs3796403 | 0.059302218 |
| SNP1 | 34330067 | HMGB4 | rs10379 | 0.059472376 |
| SNP11 | 5718517 | TRIM22 | rs7935564 | 0.059934919 |
| SNP19 | 55672784 | C19orf51 | rs7260320 | 0.060057554 |
| SNP11 | 18737095 | IGSF22 | rs4424652 | 0.060099514 |
| SNP11 | 18738281 | IGSF22 | rs10766494 | 0.060099514 |
| SNP6 | 55639028 | BMP5 | rs41271330 | 0.06049379 |
| SNP15 | 99670518 | SYNM | rs1670227 | 0.061189281 |
| SNP3 | 10302172 | TATDN2 | rs394558 | 0.061973816 |
| SNP19 | 3738971 | TJP3 | rs600986 | 0.062084018 |
| SNP13 | 67802339 | PCDH9 | rs8000556 | 0.062147304 |
| SNP22 | 43558972 | TSPO | rs41371752 | 0.062183538 |
| SNP22 | 43558972 | TSPO | rs6972 | 0.062183538 |
| SNP2 | 179638238 | TTN | rs2291306 | 0.062431707 |
| SNP3 | 130743812 | ASTE1 | rs35558913 | 0.062506991 |
| SNP22 | 39387558 | APOBEC3B | rs1065184 | 0.062581591 |
| SNP22 | 39387558 | APOBEC3B | rs57217289 | 0.062581591 |
| SNP17 | 5042894 | USP6 | rs8073787 | 0.062874871 |
| SNP8 | 19190491 | SH2D4A | rs1574288 | 0.062920365 |
| SNP2 | 70188676 | ASPRV1 | rs3796097 | 0.063022499 |
| SNP11 | 233067 | SIRT3 | rs11246020 | 0.063132164 |
| SNP11 | 233067 | SIRT3 | rs61744899 | 0.063132164 |
| SNP11 | 233212 | SIRT3 | rs11555236 | 0.063132164 |
| SNP4 | 8229774 | AC104650.1 | rs1281149 | 0.063516487 |
| SNP6 | 42689755 | PRPH2 | rs7764439 | 0.063653549 |
| SNP19 | 40739513 | AKT2 | rs33933140 | 0.063940122 |
| SNP1 | 247614617 | OR2B11 | rs4925663 | 0.064084456 |
| SNP11 | 119244095 | USP2 | rs587985 | 0.064688069 |
| SNP2 | 3483205 | TTC15 | rs6767 | 0.064737688 |
| SNP18 | 32917644 | ZNF24 | rs2032729 | 0.064749265 |
| SNP19 | 38673298 | SIPA1L3 | rs3745945 | 0.064871451 |
| SNP1 | 169510380 | F5 | rs9287090 | 0.065141791 |
| SNP1 | 152957887 | SPRR1A | rs1611764 | 0.06527053 |
| SNP7 | 65425894 | GUSB | rs9530 | 0.065815169 |
| SNP18 | 8783835 | KIAA0802 | rs35739383 | 0.065822133 |
| SNP19 | 19625547 | TSSK6 | rs7250893 | 0.065969508 |
| SNP1 | 215793834 | KCTD3 | rs14137 | 0.065977348 |
| SNP11 | 63057925 | SLC22A10 | rs1790218 | 0.065982157 |
| SNP17 | 6981353 | CLEC10A | rs732828 | 0.066234738 |
| SNP19 | 4511955 | PLIN4 | rs7260518 | 0.066456238 |
| SNP10 | 73270906 | CDH23 | rs3802720 | 0.066770781 |
| SNP10 | 133761285 | PPP2R2D | rs34473884 | 0.067223336 |
| SNP3 | 170825920 | TNIK | rs2291900 | 0.067506328 |
| SNP1 | 64643277 | ROR1 | rs7527017 | 0.067578409 |
| SNP1 | 64643277 | ROR1 | rs80063252 | 0.067578409 |
| SNP6 | 150211123 | RAET1E | rs2151910 | 0.067832819 |
| SNP19 | 3162909 | GNA15 | rs1637656 | 0.067918565 |
| SNP10 | 126714714 | CTBP2 | rs2946994 | 0.068628228 |
| SNP10 | 126714714 | CTBP2 | rs149670486 | 0.068628228 |
| SNP16 | 57080528 | NLRC5 | rs289723 | 0.068749759 |
| SNP11 | 406043 | SIGIRR | rs7947 | 0.068839172 |
| SNP1 | 75214441 | TYW3 | rs1133891 | 0.068942568 |
| SNP9 | 4685008 | CDC37L1 | rs2295967 | 0.069053552 |
| SNP19 | 13445208 | CACNA1A | rs2248069 | 0.069161165 |
| SNP2 | 71351487 | MCEE | rs11541017 | 0.069164174 |
| SNP2 | 233321641 | ALPI | rs61732032 | 0.069348119 |
| SNP6 | 106999822 | AIM1 | rs2297970 | 0.069372155 |
| SNP17 | 2318550 | AC006435.1 | rs4613098 | 0.069630022 |
| SNP12 | 123200937 | GPR109B | rs1696352 | 0.069821353 |
| SNP5 | 107197420 | | rs61749621 | 0.069915813 |
| SNP5 | 107197502 | FBXL17 | rs34990078 | 0.069915813 |
| SNP1 | 25890189 | LDLRAP1 | rs28969504 | 0.070338121 |
| SNP15 | 42302437 | PLA2G4E | rs9920824 | 0.07040767 |
| SNP16 | 11154770 | CLEC16A | rs2286973 | 0.070561976 |
| SNP20 | 62896665 | PCMTD2 | rs17878941 | 0.070854833 |
| SNP5 | 77412011 | AP3B1 | rs42360 | 0.07101506 |
| SNP6 | 158517308 | SYNJ2 | rs2502601 | 0.07174732 |
| SNP14 | 94391699 | FAM181A | rs10141024 | 0.071794883 |
| SNP19 | 855966 | ELANE | rs17216649 | 0.071849593 |
| SNP19 | 8376431 | NDUFA7 | rs561 | 0.071992675 |
| SNP20 | 33874720 | FAM83C | rs2425049 | 0.072686065 |
| SNP19 | 54759361 | LILRB5 | rs12975366 | 0.073045195 |
| SNP5 | 94939193 | ARSK | rs17084927 | 0.073811914 |
| SNP14 | 92441066 | TRIP11 | rs1051340 | 0.07409547 |
| SNP17 | 3422032 | TRPV3 | rs7216486 | 0.07427913 |
| SNP9 | 91656963 | SHC3 | rs3750399 | 0.074308989 |
| SNP4 | 47408709 | GABRB1 | rs6289 | 0.074606525 |
| SNP12 | 70989977 | PTPRB | rs11178317 | 0.075076398 |
| SNP2 | 192160839 | MYO1B | rs4853574 | 0.075220916 |
| SNP2 | 192160839 | MYO1B | rs17854823 | 0.075220916 |
| SNP14 | 93170993 | LGMN | rs9791 | 0.075602218 |
| SNP14 | 93170993 | LGMN | rs61734480 | 0.075602218 |
| SNP13 | 25367301 | RNF17 | rs1158061 | 0.075768865 |
| SNP15 | 66625161 | DIS3L | rs11071885 | 0.075920776 |
| SNP16 | 929711 | LMF1 | rs2076425 | 0.075926403 |
| SNP2 | 39583446 | MAP4K3 | rs7422651 | 0.075926403 |
| SNP17 | 8172506 | PFAS | rs62637606 | 0.076038676 |
| SNP4 | 126336703 | FAT4 | rs17009618 | 0.076242768 |
| SNP10 | 122663585 | WDR11 | rs1652727 | 0.076315298 |
| SNP15 | 33357262 | FMN1 | rs2306277 | 0.076389876 |
| SNP1 | 41485902 | SLFNL1 | rs3738368 | 0.076530909 |
| SNP21 | 46086758 | KRTAP12-2 | rs35163632 | 0.076530909 |
| SNP21 | 46086758 | KRTAP12-2 | rs13046903 | 0.076530909 |
| SNP6 | 2623820 | C6orf195 | rs6902511 | 0.077021148 |
| SNP6 | 160328620 | MAS1 | rs220721 | 0.077163713 |
| SNP3 | 133494354 | TF | rs1049296 | 0.077221069 |
| SNP8 | 52733050 | PCMTD1 | rs12335014 | 0.077290641 |
| SNP17 | 6704071 | TEKT1 | rs17804647 | 0.077302884 |
| SNP1 | 75037845 | C1orf173 | rs12723334 | 0.077436016 |
| SNP1 | 31215364 | LAPTM5 | rs1050663 | 0.077822839 |
| SNP4 | 184931818 | STOX2 | rs4861597 | 0.078019309 |
| SNP17 | 27959903 | SSH2 | rs2289629 | 0.07828713 |
| SNP11 | 4661243 | OR51D1 | rs61740347 | 0.078434056 |
| SNP11 | 4661568 | OR51D1 | rs61746547 | 0.078434056 |
| SNP11 | 4661826 | OR51D1 | rs79020081 | 0.078434056 |
| SNP8 | 124710729 | ANXA13 | rs2294013 | 0.078733394 |
| SNP17 | 73552185 | LLGL2 | rs1671036 | 0.078854886 |
| SNP19 | 37676528 | ZNF585B | rs8111369 | 0.07892639 |
| SNP19 | 37676528 | ZNF585B | rs75662394 | 0.07892639 |
| SNP1 | 111730901 | DENND2D | rs608881 | 0.079085374 |
| SNP17 | 10304500 | MYH8 | rs3744553 | 0.079159808 |
| SNP10 | 12209752 | NUDT5 | rs6686 | 0.079219364 |
| SNP1 | 248059456 | OR2W3 | rs12135078 | 0.079473968 |
| SNP12 | 96266035 | CCDC38 | rs2117914 | 0.079782871 |
| SNP4 | 141598126 | TBC1D9 | rs2303911 | 0.079873216 |
| SNP19 | 731144 | PALM | rs1050457 | 0.080718264 |
| SNP9 | 35674053 | CA9 | rs2071676 | 0.080941818 |
| SNP6 | 66044927 | EYS | rs61753610 | 0.080962919 |
| SNP22 | 22453513 | | rs9611014 | 0.0811611 |
| SNP6 | 31323233 | HLA-B | rs709052 | 0.081173035 |
| SNP6 | 31323233 | HLA-B | rs145627832 | 0.081173035 |
| SNP6 | 31323233 | HLA-B | rs41557013 | 0.081173035 |
| SNP1 | 110300441 | EPS8L3 | rs3818562 | 0.081412154 |
| SNP3 | 121526204 | IQCB1 | rs4543051 | 0.081532413 |
| SNP5 | 56778103 | ACTBL2 | rs61737336 | 0.081618275 |
| SNP11 | 118895686 | TRAPPC4 | rs8192696 | 0.081636703 |
| SNP19 | 5783634 | PRR22 | rs2446210 | 0.081662785 |
| SNP1 | 162367103 | SH2D1B | rs34001279 | 0.08170039 |
| SNP15 | 23931507 | NDN | rs2192206 | 0.081711274 |
| SNP17 | 33690619 | SLFN11 | rs72825958 | 0.081887398 |
| SNP1 | 177902388 | AL359075.1 | rs3813647 | 0.081920741 |
| SNP16 | 19085298 | COQ7 | rs11074359 | 0.082077016 |
| SNP1 | 161495477 | HSPA6 | rs17853454 | 0.082224005 |
| SNP1 | 161495477 | HSPA6 | rs72633678 | 0.082224005 |
| SNP3 | 9867625 | TTLL3 | rs3732527 | 0.082357233 |
| SNP3 | 27478899 | SLC4A7 | rs2029618 | 0.082498098 |
| SNP1 | 2444414 | PANK4 | rs7535528 | 0.08296228 |
| SNP11 | 123894234 | OR10G9 | rs11219413 | 0.083046749 |
| SNP11 | 123894402 | OR10G9 | rs12221656 | 0.083046749 |
| SNP11 | 123894402 | OR10G9 | rs78197836 | 0.083046749 |
| SNP21 | 33956579 | TCP10L | rs2017816 | 0.083167056 |
| SNP1 | 77093180 | ST6GALNAC3 | rs1184626 | 0.083662436 |
| SNP10 | 27389197 | ANKRD26 | rs7897309 | 0.083668318 |
| SNP12 | 78400884 | NAV3 | rs34276383 | 0.083736836 |
| SNP16 | 71773190 | AP1G1 | rs904763 | 0.083775216 |
| SNP5 | 43175103 | ZNF131 | rs35397154 | 0.084045049 |
| SNP1 | 12082926 | MIIP | rs2295283 | 0.084119195 |
| SNP12 | 99640557 | ANKS1B | rs1552759 | 0.084834765 |
| SNP2 | 71906278 | DYSF | rs17718530 | 0.085096543 |
| SNP10 | 97388162 | ALDH18A1 | rs2275272 | 0.085117931 |
| SNP10 | 97388162 | ALDH18A1 | rs1063928 | 0.085117931 |
| SNP6 | 97058574 | FHL5 | rs2252816 | 0.085138969 |
| SNP11 | 1026048 | MUC6 | rs55965773 | 0.085147868 |
| SNP19 | 35648365 | FXYD5 | rs1688005 | 0.085787525 |
| SNP1 | 160136350 | ATP1A4 | rs7529215 | 0.085828903 |
| SNP1 | 160136350 | ATP1A4 | rs55710009 | 0.085828903 |
| SNP1 | 160136350 | ATP1A4 | rs143083729 | 0.085828903 |
| SNP11 | 73670645 | DNAJB13 | rs653263 | 0.086081277 |
| SNP19 | 17025292 | CPAMD8 | rs706761 | 0.086723539 |
| SNP22 | 23047290 | | rs9623875 | 0.08719579 |
| SNP20 | 52675188 | BCAS1 | rs394732 | 0.087619001 |
| SNP6 | 107076783 | RTN4IP1 | rs1987623 | 0.087955105 |
| SNP2 | 207174316 | ZDBF2 | rs3732084 | 0.088007727 |
| SNP8 | 103573001 | ODF1 | rs66466156 | 0.088070292 |
| SNP8 | 103573001 | ODF1 | rs3018445 | 0.088070292 |
| SNP16 | 89704365 | DPEP1 | rs1126464 | 0.088550678 |
| SNP22 | 23077555 | | rs4822296 | 0.088714053 |
| SNP15 | 74473739 | STRA6 | rs736118 | 0.089215731 |
| SNP9 | 35957669 | OR2S2 | rs2233564 | 0.089238648 |
| SNP9 | 35958047 | OR2S2 | rs2233558 | 0.089238648 |
| SNP2 | 108863650 | | rs9646947 | 0.08935338 |
| SNP1 | 118644430 | SPAG17 | rs10754367 | 0.089484247 |
| SNP7 | 107593989 | LAMB1 | rs20556 | 0.089727294 |
| SNP17 | 4337283 | SPNS3 | rs35751906 | 0.089879676 |
| SNP16 | 89613123 | SPG7 | rs2292954 | 0.090717763 |
| SNP9 | 109691676 | ZNF462 | rs3814538 | 0.0907485 |
| SNP10 | 61414011 | SLC16A9 | rs2242206 | 0.091315379 |
| SNP10 | 75673101 | PLAU | rs2227564 | 0.091513296 |
| SNP17 | 47588000 | NGFR | rs11466155 | 0.092006799 |
| SNP6 | 44117607 | TMEM63B | rs4714761 | 0.092253125 |
| SNP13 | 33703656 | STARD13 | rs495680 | 0.092313921 |
| SNP3 | 182925404 | MCF2L2 | rs6804951 | 0.092461648 |
| SNP3 | 184647413 | VPS8 | rs4643688 | 0.092461648 |
| SNP8 | 125035788 | FER1L6 | rs4358790 | 0.092542632 |
| SNP10 | 73537614 | CDH23 | rs17712523 | 0.092643084 |
| SNP3 | 196674879 | PIGZ | rs12636891 | 0.09268539 |
| SNP17 | 3839685 | ATP2A3 | rs1062683 | 0.092749743 |
| SNP19 | 58084930 | ZNF416 | rs3746222 | 0.09304643 |
| SNP19 | 45448465 | APOC4 | rs5167 | 0.093497896 |
| SNP13 | 32912299 | BRCA2 | rs543304 | 0.09354902 |
| SNP14 | 52186972 | FRMD6 | rs2277494 | 0.093707036 |
| SNP14 | 91671124 | C14orf159 | rs2295524 | 0.09418971 |
| SNP1 | 205306590 | KLHDC8A | rs3210952 | 0.094834725 |
| SNP7 | 133979795 | SLC35B4 | rs1421484 | 0.095025535 |
| SNP20 | 5753579 | C20orf196 | rs237422 | 0.095487577 |
| SNP8 | 133900252 | TG | rs180223 | 0.095627429 |
| SNP8 | 133909974 | TG | rs853326 | 0.095627429 |
| SNP8 | 142170884 | DENND3 | rs2289001 | 0.096059794 |
| SNP11 | 33053107 | DEPDC7 | rs966191 | 0.096095597 |
| SNP5 | 13719022 | DNAH5 | rs30169 | 0.096397731 |
| SNP15 | 90768271 | SEMA4B | rs11547966 | 0.096588839 |
| SNP2 | 44071743 | ABCG8 | rs4148211 | 0.09667394 |
| SNP17 | 58121453 | HEATR6 | rs16943991 | 0.096942842 |
| SNP19 | 44535999 | ZNF222 | rs7258517 | 0.096983184 |
| SNP1 | 9307138 | H6PD | rs11121350 | 0.097255835 |
| SNP19 | 3186085 | NCLN | rs11551095 | 0.097388577 |
| SNP9 | 116973273 | COL27A1 | rs34350265 | 0.098171051 |
| SNP18 | 20953720 | C18orf45 | rs8099409 | 0.098621673 |
| SNP10 | 13534851 | BEND7 | rs2251555 | 0.099087173 |
| SNP2 | 202082459 | CASP10 | rs13006529 | 0.099163542 |
| SNP12 | 60173406 | SLC16A7 | rs3763979 | 0.099634589 |
| SNP1 | 28209362 | C1orf38 | rs3766399 | 0.099905587 |
| SNP18 | 60562298 | PHLPP1 | rs624821 | 0.099905587 |
| SNP1 | 196682947 | CFH | rs2274700 | 0.099922802 |
| SNP2 | 160879259 | PLA2R1 | rs33985939 | 0.10034987 |
| SNP17 | 55950064 | CUEDC1 | rs35704289 | 0.100910616 |
| SNP14 | 73969610 | HEATR4 | rs11626122 | 0.101050207 |
| SNP1 | 38463504 | FHL3 | rs7366048 | 0.101129271 |
| SNP19 | 52825235 | ZNF480 | rs8102373 | 0.101366204 |
| SNP16 | 2818161 | SRRM2 | rs2301802 | 0.101670119 |
| SNP3 | 148727133 | GYG1 | rs4938 | 0.101758399 |
| SNP10 | 98380137 | PIK3AP1 | rs12784975 | 0.101820011 |
| SNP12 | 121442199 | C12orf43 | rs3751151 | 0.101984076 |
| SNP15 | 50474766 | SLC27A2 | rs1648348 | 0.102346975 |
| SNP20 | 3193978 | ITPA | rs8362 | 0.102621632 |
| SNP7 | 141752213 | MGAM | rs2961085 | 0.103297149 |
| SNP21 | 32253629 | KRTAP11-1 | rs71321355 | 0.103362572 |
| SNP22 | 22890792 | PRAME | rs17497547 | 0.104103209 |
| SNP1 | 28861636 | RCC1 | rs2066726 | 0.104494184 |
| SNP6 | 21065450 | CDKAL1 | rs56087852 | 0.104823991 |
| SNP3 | 418089 | | rs3773380 | 0.104864862 |
| SNP21 | 43531632 | UMODL1 | rs220129 | 0.105171404 |
| SNP3 | 105258861 | ALCAM | rs1044240 | 0.105200107 |
| SNP1 | 40980559 | DEM1 | rs1134586 | 0.105228771 |
| SNP20 | 57290347 | NPEPL1 | rs6026468 | 0.105234947 |
| SNP14 | 77951124 | ISM2 | rs3742728 | 0.105428277 |
| SNP21 | 43896143 | RSPH1 | rs117385282 | 0.105509939 |
| SNP21 | 43896143 | RSPH1 | rs142166675 | 0.105509939 |
| SNP9 | 77257346 | RORB | rs2273975 | 0.105564425 |
| SNP9 | 117668142 | TNFSF8 | rs3181195 | 0.105720406 |
| SNP1 | 162344102 | C1orf111 | rs2282397 | 0.106209916 |
| SNP7 | 37936530 | TXNDC3 | rs3213975 | 0.106450677 |
| SNP19 | 43763033 | PSG9 | rs3180651 | 0.106904347 |
| SNP22 | 50873497 | PPP6R2 | rs1134848 | 0.107114774 |
| SNP16 | 68857441 | CDH1 | rs1801552 | 0.107567813 |
| SNP5 | 140683265 | SLC25A2 | rs11952797 | 0.10766603 |
| SNP6 | 31079703 | C6orf15 | rs2233978 | 0.107881743 |
| SNP6 | 31079703 | C6orf15 | rs144759863 | 0.107881743 |
| SNP7 | 88423881 | C7orf62 | rs2293583 | 0.108009702 |
| SNP17 | 7154582 | CTDNEP1 | rs3744399 | 0.108283816 |
| SNP1 | 70820880 | HHLA3 | rs55863806 | 0.10865763 |
| SNP11 | 34991727 | PDHX | rs497582 | 0.108951143 |
| SNP3 | 8675539 | C3orf32 | rs2276800 | 0.108955726 |
| SNP7 | 150645534 | KCNH2 | rs1805123 | 0.109718279 |
| SNP20 | 2539387 | TMC2 | rs6050063 | 0.110105947 |
| SNP10 | 71010375 | HKDC1 | rs5030948 | 0.11051243 |
| SNP19 | 10106936 | COL5A3 | rs1559186 | 0.111146856 |
| SNP8 | 20038466 | SLC18A1 | rs2270641 | 0.111715049 |
| SNP5 | 143200053 | HMHB1 | rs161557 | 0.111827351 |
| SNP14 | 23598976 | SLC7A8 | rs17183863 | 0.112031338 |
| SNP2 | 105713697 | MRPS9 | rs15567 | 0.113395456 |
| SNP6 | 56917538 | KIAA1586 | rs6926980 | 0.113600523 |
| SNP7 | 143771408 | OR2A25 | rs59319753 | 0.114002908 |
| SNP12 | 106903321 | POLR3B | rs13561 | 0.114869692 |
| SNP14 | 96103099 | | rs4905366 | 0.114869692 |
| SNP9 | 101611335 | GALNT12 | rs2273846 | 0.115631512 |
| SNP18 | 56247600 | ALPK2 | rs9944810 | 0.115870599 |
| SNP19 | 23545004 | ZNF91 | rs296092 | 0.117328293 |
| SNP2 | 207603234 | MDH1B | rs2287631 | 0.11746031 |
| SNP19 | 8138054 | FBN3 | rs7257948 | 0.117690317 |
| SNP1 | 14105139 | PRDM2 | rs2076324 | 0.118191786 |
| SNP1 | 14105139 | PRDM2 | rs148625644 | 0.118191786 |
| SNP11 | 11977573 | USP47 | rs71734478 | 0.118597443 |
| SNP11 | 11977573 | USP47 | rs2307073 | 0.118597443 |
| SNP12 | 49580180 | TUBA1A | rs1056875 | 0.118765785 |
| SNP14 | 23044002 | | rs7621 | 0.11912907 |
| SNP3 | 10885920 | SLC6A11 | rs2304725 | 0.119438672 |
| SNP17 | 10427924 | MYH2 | rs1042236 | 0.11951675 |
| SNP17 | 10427924 | MYH2 | rs61730799 | 0.11951675 |
| SNP15 | 55722872 | DYX1C1 | rs77641439 | 0.120318396 |
| SNP19 | 53612055 | ZNF415 | rs10410030 | 0.120520882 |
| SNP20 | 44746982 | | rs1883832 | 0.120620024 |
| SNP3 | 169539990 | LRRIQ4 | rs61754874 | 0.120940059 |
| SNP19 | 1256998 | MIDN | rs9823 | 0.121411755 |
| SNP19 | 1256998 | MIDN | rs139717027 | 0.121411755 |
| SNP17 | 15554504 | TRIM16 | rs4792642 | 0.121441845 |
| SNP17 | 15554504 | TRIM16 | rs144568697 | 0.121441845 |
| SNP16 | 71411636 | CALB2 | rs11545954 | 0.121859505 |
| SNP7 | 94041937 | COL1A2 | rs412777 | 0.122663545 |
| SNP7 | 122303321 | CADPS2 | rs2251761 | 0.123150397 |
| SNP3 | 124646837 | MUC13 | rs4679394 | 0.12331026 |
| SNP12 | 132325239 | MMP17 | rs6598163 | 0.12346743 |
| SNP4 | 1244416 | C4orf42 | rs1732115 | 0.12379347 |
| SNP22 | 50297888 | ALG12 | rs3922872 | 0.123845702 |
| SNP7 | 129664312 | ZC3HC1 | rs1464890 | 0.124957121 |
| SNP12 | 7945559 | NANOG | rs4294629 | 0.12512076 |
| SNP10 | 29932918 | | rs2488689 | 0.125280218 |
| SNP3 | 39111140 | WDR48 | rs2293312 | 0.125310881 |
| SNP3 | 196674916 | PIGZ | rs1147240 | 0.125343876 |
| SNP1 | 151747970 | TDRKH | rs11204885 | 0.125703176 |
| SNP19 | 51983673 | CEACAM18 | rs61743859 | 0.125992721 |
| SNP17 | 1840468 | RTN4RL1 | rs9903800 | 0.126594011 |
| SNP18 | 43262359 | SLC14A2 | rs3745009 | 0.127361225 |
| SNP6 | 74155346 | MB21D1 | rs610913 | 0.128370748 |
| SNP16 | 20811681 | ERI2 | rs2301771 | 0.12871004 |
| SNP21 | 28338423 | ADAMTS5 | rs55933916 | 0.129105083 |
| SNP12 | 107395106 | CRY1 | rs8192440 | 0.129937835 |
| SNP20 | 31622083 | BPIL3 | rs2070317 | 0.129982516 |
| SNP1 | 112018657 | C1orf162 | rs6703267 | 0.130426939 |
| SNP4 | 40438576 | RBM47 | rs2307046 | 0.130429469 |
| SNP2 | 138420996 | THSD7B | rs10206850 | 0.130647992 |
| SNP2 | 277003 | ACP1 | rs79716074 | 0.130792146 |
| SNP9 | 36169723 | CCIN | rs34789048 | 0.131398691 |
| SNP19 | 46057081 | OPA3 | rs3826860 | 0.132259177 |
| SNP7 | 42004600 | GLI3 | rs34089404 | 0.132299778 |
| SNP8 | 12957475 | DLC1 | rs532841 | 0.132299778 |
| SNP2 | 152322095 | RIF1 | rs2444257 | 0.132824591 |
| SNP12 | 4388084 | CCND2 | rs3217805 | 0.133200198 |
| SNP3 | 9876987 | TTLL3 | rs1057281 | 0.133543223 |
| SNP3 | 195615376 | TNK2 | rs3747669 | 0.133726848 |
| SNP1 | 67861520 | IL12RB2 | rs2229546 | 0.133985986 |
| SNP2 | 98844674 | VWA3B | rs7601049 | 0.134005519 |
| SNP2 | 209010558 | CRYGB | rs2854723 | 0.134117433 |
| SNP15 | 57731573 | CGNL1 | rs7182648 | 0.134205141 |
| SNP8 | 142204326 | DENND3 | rs1045248 | 0.134223456 |
| SNP2 | 233899126 | NEU2 | rs2233391 | 0.134233187 |
| SNP9 | 114411945 | DNAJC25 | rs7019332 | 0.134452414 |
| SNP19 | 1004724 | GRIN3B | rs11880849 | 0.134824865 |
| SNP12 | 20852546 | SLCO1C1 | rs34243130 | 0.134942031 |
| SNP11 | 30033191 | KCNA4 | rs3802914 | 0.135181952 |
| SNP6 | 46803018 | MEP1A | rs2297020 | 0.135213023 |
| SNP6 | 46803018 | MEP1A | rs74742999 | 0.135213023 |
| SNP1 | 201012597 | CACNA1S | rs41267497 | 0.135318889 |
| SNP19 | 47124714 | PTGIR | rs2229129 | 0.135446359 |
| SNP16 | 633354 | PIGQ | rs71391136 | 0.135506012 |
| SNP16 | 633354 | PIGQ | rs710925 | 0.135506012 |
| SNP16 | 633354 | PIGQ | rs112194445 | 0.135506012 |
| SNP19 | 56733425 | ZSCAN5A | rs34187696 | 0.13611077 |
| SNP15 | 22939192 | CYFIP1 | rs11633474 | 0.136194327 |
| SNP14 | 105180565 | INF2 | rs4983535 | 0.136210858 |
| SNP4 | 56874517 | CEP135 | rs3214045 | 0.136412579 |
| SNP14 | 93118038 | RIN3 | rs3829947 | 0.136891184 |
| SNP3 | 58552997 | FAM107A | rs1043942 | 0.137004557 |
| SNP11 | 72946140 | P2RY2 | rs3741156 | 0.137515267 |
| SNP21 | 45970812 | KRTAP10-2 | rs2329834 | 0.137646631 |
| SNP3 | 42599091 | SEC22C | rs2271184 | 0.137806929 |
| SNP17 | 48539035 | ACSF2 | rs9674937 | 0.137904493 |
| SNP19 | 56114045 | ZNF524 | rs1077806 | 0.137996521 |
| SNP13 | 111870037 | ARHGEF7 | rs2296354 | 0.138110549 |
| SNP17 | 80350331 | C17orf101 | rs11903 | 0.138110549 |
| SNP1 | 60466814 | C1orf87 | rs626251 | 0.138428304 |
| SNP1 | 16265904 | SPEN | rs41269155 | 0.138572236 |
| SNP19 | 56416409 | NLRP13 | rs28506091 | 0.138572236 |
| SNP2 | 130925088 | SMPD4 | rs10909567 | 0.138585247 |
| SNP12 | 40688695 | LRRK2 | rs7966550 | 0.138633159 |
| SNP22 | 50754648 | FAM116B | rs11553141 | 0.138955659 |
| SNP6 | 49425521 | MUT | rs2229384 | 0.139856775 |
| SNP13 | 25009485 | PARP4 | rs1050110 | 0.139931628 |
| SNP13 | 25009485 | PARP4 | rs41506549 | 0.139931628 |
| SNP15 | 86125031 | AKAP13 | rs7178065 | 0.139931879 |
| SNP5 | 149776232 | TCOF1 | rs15251 | 0.14014999 |
| SNP11 | 62346131 | TUT1 | rs3888173 | 0.140552044 |
| SNP12 | 93202801 | EEA1 | rs7970286 | 0.140802265 |
| SNP11 | 8959370 | ASCL3 | rs2742505 | 0.141307674 |
| SNP6 | 37252210 | TBC1D22B | rs3818136 | 0.141559624 |
| SNP21 | 45502844 | TRAPPC10 | rs915877 | 0.142714083 |
| SNP17 | 80899281 | TBCD | rs9390 | 0.142752234 |
| SNP19 | 51871195 | CLDND2 | rs3745403 | 0.142784969 |
| SNP15 | 80191343 | ST20 | rs7257 | 0.142798505 |
| SNP1 | 215960167 | USH2A | rs10864198 | 0.14332529 |
| SNP16 | 16138322 | ABCC1 | rs246221 | 0.143790382 |
| SNP2 | 207833983 | CPO | rs7582305 | 0.1440239 |
| SNP12 | 6935977 | GPR162 | rs11612427 | 0.144377046 |
| SNP1 | 205554085 | MFSD4 | rs7526132 | 0.144496837 |
| SNP4 | 2341194 | ZFYVE28 | rs2071680 | 0.144701739 |
| SNP12 | 53045777 | KRT2 | rs11835758 | 0.144860464 |
| SNP4 | 149358014 | | rs2070951 | 0.145703117 |
| SNP8 | 121215991 | COL14A1 | rs2305600 | 0.145703117 |
| SNP3 | 122277268 | PARP9 | rs34006803 | 0.145991689 |
| SNP21 | 43531008 | UMODL1 | rs220126 | 0.14637055 |
| SNP17 | 8243661 | ODF4 | rs12601097 | 0.146861702 |
| SNP10 | 44112245 | ZNF485 | rs12354886 | 0.147071345 |
| SNP22 | 51117580 | SHANK3 | rs9616915 | 0.147179626 |
| SNP19 | 17273893 | MYO9B | rs7256689 | 0.147522014 |
| SNP8 | 141551407 | EIF2C2 | rs2293939 | 0.147772103 |
| SNP19 | 4288332 | SHD | rs888930 | 0.148113911 |
| SNP19 | 4288332 | SHD | rs61739549 | 0.148113911 |
| SNP3 | 52430526 | DNAH1 | rs12163565 | 0.148826492 |
| SNP22 | 19969106 | ARVCF | rs2240717 | 0.149298531 |
| SNP19 | 48519205 | ELSPBP1 | rs34647554 | 0.149973746 |
| SNP20 | 61981104 | CHRNA4 | rs1044397 | 0.150205484 |
| SNP16 | 23672560 | DCTN5 | rs11545871 | 0.150215817 |
| SNP17 | 43192821 | PLCD3 | rs713101 | 0.150466037 |
| SNP3 | 89521693 | EPHA3 | rs35124509 | 0.150473507 |
| SNP3 | 89521693 | EPHA3 | rs150555764 | 0.150473507 |
| SNP9 | 139656670 | LCN15 | rs2297722 | 0.150649832 |
| SNP5 | 96130836 | ERAP1 | rs26618 | 0.150949095 |
| SNP4 | 39921942 | PDS5A | rs28449663 | 0.150963415 |
| SNP10 | 72307101 | KIAA1274 | rs3740447 | 0.151234233 |
| SNP14 | 24033027 | AP1G2 | rs12897422 | 0.151608089 |
| SNP19 | 55086775 | LILRA2 | rs1052120 | 0.151671981 |
| SNP10 | 459940 | DIP2C | rs4881274 | 0.151775535 |
| SNP10 | 459940 | DIP2C | rs140052373 | 0.151775535 |
| SNP8 | 144654249 | C8orf73 | rs10866911 | 0.152165879 |
| SNP19 | 6702157 | C3 | rs428453 | 0.152222096 |
| SNP8 | 143746050 | AC145123.1 | rs2976399 | 0.152264467 |
| SNP14 | 105414280 | AHNAK2 | rs2819429 | 0.152666453 |
| SNP3 | 54919351 | CACNA2D3 | rs10510774 | 0.152856992 |
| SNP3 | 39307162 | CX3CR1 | rs3732378 | 0.152950455 |
| SNP11 | 5153261 | OR52A5 | rs2472530 | 0.153904251 |
| SNP4 | 75675841 | BTC | rs11938093 | 0.153949112 |
| SNP14 | 21623648 | OR5AU1 | rs7161544 | 0.154622794 |
| SNP15 | 99023968 | FAM169B | rs4528551 | 0.154641096 |
| SNP10 | 122649482 | WDR11 | rs2289337 | 0.156106892 |
| SNP10 | 122649482 | WDR11 | rs61761620 | 0.156106892 |
| SNP20 | 37601243 | DHX35 | rs16987712 | 0.157208423 |
| SNP1 | 19595137 | AKR7L | rs2235795 | 0.158211031 |
| SNP12 | 51236802 | TMPRSS12 | rs10876100 | 0.158251293 |
| SNP19 | 55614923 | PPP1R12C | rs2532060 | 0.158480184 |
| SNP3 | 183558402 | PARL | rs3732581 | 0.159005785 |
| SNP9 | 129102840 | FAM125B | rs1888156 | 0.159402013 |
| SNP19 | 829568 | AZU1 | rs12460890 | 0.159678649 |
| SNP11 | 32460624 | | rs3087923 | 0.159728342 |
| SNP22 | 19048413 | DGCR2 | rs2238739 | 0.159752077 |
| SNP12 | 9833628 | CLEC2D | rs3764021 | 0.160379024 |
| SNP11 | 62458275 | BSCL2 | rs6856 | 0.160833293 |
| SNP7 | 55238874 | EGFR | rs17337023 | 0.161461289 |
| SNP9 | 20764870 | KIAA1797 | rs10511687 | 0.163452891 |
| SNP16 | 21222672 | ZP2 | rs2075520 | 0.163954902 |
| SNP19 | 629846 | POLRMT | rs11550305 | 0.16418521 |
| SNP19 | 629846 | POLRMT | rs55649970 | 0.16418521 |
| SNP20 | 18794754 | C20orf79 | rs1053839 | 0.16418521 |
| SNP6 | 28543264 | SCAND3 | rs17336532 | 0.164454215 |
| SNP6 | 28543264 | SCAND3 | rs114476214 | 0.164454215 |
| SNP19 | 37441462 | ZNF568 | rs546589 | 0.164506899 |
| SNP3 | 1424718 | CNTN6 | rs2291101 | 0.164539431 |
| SNP7 | 128140982 | METTL2B | rs10257897 | 0.164779202 |
| SNP7 | 128140982 | METTL2B | rs80061969 | 0.164779202 |
| SNP7 | 128140982 | METTL2B | rs139332186 | 0.164779202 |
| SNP14 | 94417421 | ASB2 | rs7147919 | 0.165339099 |
| SNP6 | 116720487 | DSE | rs10485183 | 0.165360596 |
| SNP6 | 116720487 | DSE | rs41352449 | 0.165360596 |
| SNP5 | 22078584 | CDH12 | rs4371716 | 0.165656272 |
| SNP11 | 1252708 | MUC5B | rs2735709 | 0.165672953 |
| SNP17 | 18775900 | PRPSAP2 | rs4393623 | 0.166113959 |
| SNP3 | 186443018 | KNG1 | rs1656922 | 0.166306161 |
| SNP9 | 134350323 | PRRC2B | rs10736851 | 0.166601566 |
| SNP8 | 67369355 | ADHFE1 | rs2555588 | 0.167344355 |
| SNP20 | 54961463 | AURKA | rs1047972 | 0.168383413 |
| SNP19 | 3731985 | TJP3 | rs1879040 | 0.168580615 |
| SNP1 | 245245402 | EFCAB2 | rs10927387 | 0.169484604 |
| SNP16 | 81314496 | BCMO1 | rs7501331 | 0.169735921 |
| SNP15 | 53081800 | ONECUT1 | rs61735385 | 0.170081722 |
| SNP2 | 237103623 | ASB18 | rs10166966 | 0.17016277 |
| SNP6 | 38951998 | DNAH8 | rs1537232 | 0.170238426 |
| SNP6 | 38951998 | DNAH8 | rs114095356 | 0.170238426 |
| SNP4 | 189012728 | TRIML2 | rs2279550 | 0.170240434 |
| SNP3 | 195453257 | MUC20 | rs3828406 | 0.170431384 |
| SNP22 | 24179922 | DERL3 | rs3177243 | 0.17063405 |
| SNP2 | 113832333 | IL1F10 | rs6743376 | 0.171038231 |
| SNP14 | 23000062 | | rs227003 | 0.171255002 |
| SNP20 | 10019093 | ANKRD5 | rs575534 | 0.171390162 |
| SNP2 | 97559759 | FAM178B | rs11677797 | 0.171530729 |
| SNP19 | 55143452 | LILRB1 | rs1061680 | 0.17216509 |
| SNP20 | 44005936 | TP53TG5 | rs2231616 | 0.172427998 |
| SNP17 | 78820374 | RPTOR | rs2589156 | 0.172509846 |
| SNP17 | 73016621 | ICT1 | rs1044228 | 0.17264497 |
| SNP22 | 38046695 | SH3BP1 | rs12170939 | 0.172815422 |
| SNP10 | 70405855 | TET1 | rs3998860 | 0.172824241 |
| SNP19 | 56156447 | ZNF581 | rs11084406 | 0.172824746 |
| SNP2 | 105859249 | GPR45 | rs35946826 | 0.172824746 |
| SNP1 | 85009894 | SPATA1 | rs12143652 | 0.173221364 |
| SNP12 | 88440676 | C12orf29 | rs9262 | 0.173764579 |
| SNP22 | 31302233 | OSBP2 | rs3804085 | 0.173783072 |
| SNP3 | 58092528 | FLNB | rs2140104 | 0.174082943 |
| SNP18 | 57022754 | LMAN1 | rs1127220 | 0.174251585 |
| SNP11 | 20112417 | NAV2 | rs2243624 | 0.175086312 |
| SNP9 | 133946909 | LAMC3 | rs10901344 | 0.176007644 |
| SNP9 | 133946915 | LAMC3 | rs10901345 | 0.176007644 |
| SNP6 | 83949261 | ME1 | rs2230902 | 0.176174183 |
| SNP6 | 83949261 | ME1 | rs1180230 | 0.176174183 |
| SNP22 | 19959473 | ARVCF | rs165815 | 0.176590418 |
| SNP1 | 230914729 | CAPN9 | rs1933631 | 0.176835505 |
| SNP20 | 56139403 | PCK1 | rs2070756 | 0.176839558 |
| SNP6 | 161137779 | PLG | rs14224 | 0.177168213 |
| SNP6 | 161137779 | PLG | rs138242513 | 0.177168213 |
| SNP3 | 137843106 | A4GNT | rs2170309 | 0.177695411 |
| SNP3 | 137843476 | A4GNT | rs2246945 | 0.177695411 |
| SNP4 | 95170839 | SMARCAD1 | rs11722476 | 0.17778793 |
| SNP11 | 124947149 | SLC37A2 | rs12276567 | 0.177867917 |
| SNP2 | 234967539 | SPP2 | rs593668 | 0.178090506 |
| SNP14 | 105478102 | CDCA4 | rs3803294 | 0.178348174 |
| SNP1 | 247006051 | AHCTF1 | rs12410563 | 0.178756081 |
| SNP7 | 131864591 | PLXNA4 | rs3734989 | 0.179021559 |
| SNP10 | 72813306 | | rs10762420 | 0.180676154 |
| SNP2 | 20205680 | MATN3 | rs28401180 | 0.180676154 |
| SNP16 | 69970329 | WWP2 | rs1983016 | 0.180771454 |
| SNP16 | 81193321 | PKD1L2 | rs72628271 | 0.180927167 |
| SNP16 | 81193321 | PKD1L2 | rs75485939 | 0.180927167 |
| SNP3 | 186358366 | FETUB | rs1047115 | 0.182069139 |
| SNP1 | 157514097 | FCRL5 | rs6679793 | 0.182083622 |
| SNP9 | 75303653 | TMC1 | rs2589615 | 0.182640333 |
| SNP9 | 75303653 | TMC1 | rs140437301 | 0.182640333 |
| SNP5 | 127614472 | FBN2 | rs190450 | 0.18266526 |
| SNP5 | 127614472 | FBN2 | rs150572769 | 0.18266526 |
| SNP4 | 87684031 | PTPN13 | rs710832 | 0.182943173 |
| SNP4 | 47682174 | CORIN | rs10517195 | 0.183851644 |
| SNP21 | 45107562 | RRP1B | rs9306160 | 0.184104527 |
| SNP21 | 45107562 | RRP1B | rs149494438 | 0.184104527 |
| SNP10 | 90771829 | FAS | rs2234978 | 0.184226152 |
| SNP1 | 52266242 | NRD1 | rs8375 | 0.185614384 |
| SNP6 | 133052616 | VNN3 | rs4895943 | 0.185826593 |
| SNP10 | 101645498 | DNMBP | rs2490763 | 0.185948506 |
| SNP21 | 34924243 | SON | rs16990760 | 0.186282305 |
| SNP4 | 83838262 | THAP9 | rs897945 | 0.186745936 |
| SNP5 | 52193287 | ITGA1 | rs1531545 | 0.186843273 |
| SNP20 | 19565671 | SLC24A3 | rs17293432 | 0.186843324 |
| SNP2 | 43931176 | PLEKHH2 | rs4953002 | 0.187024063 |
| SNP19 | 49376582 | PPP1R15A | rs564196 | 0.187233611 |
| SNP9 | 125391677 | OR1B1 | rs12347681 | 0.187400632 |
| SNP16 | 70726795 | VAC14 | rs2278983 | 0.187548336 |
| SNP17 | 79511135 | C17orf70 | rs8077430 | 0.188374829 |
| SNP8 | 13072177 | DLC1 | rs3739300 | 0.188864186 |
| SNP22 | 22550510 | | rs2073448 | 0.189160772 |
| SNP10 | 124096035 | BTBD16 | rs3817281 | 0.189738883 |
| SNP11 | 17418477 | ABCC8 | rs757110 | 0.189972833 |
| SNP10 | 5203864 | AKR1CL1 | rs11253021 | 0.190731336 |
| SNP10 | 5203864 | AKR1CL1 | rs61729616 | 0.190731336 |
| SNP10 | 5204928 | AKR1CL1 | rs7097295 | 0.190731336 |
| SNP10 | 5204928 | AKR1CL1 | rs150568767 | 0.190731336 |
| SNP7 | 24703298 | MPP6 | rs1053962 | 0.19121587 |
| SNP17 | 7592168 | WRAP53 | rs2287499 | 0.191376446 |
| SNP17 | 7592168 | WRAP53 | rs111433356 | 0.191376446 |
| SNP13 | 52603896 | UTP14C | rs17402034 | 0.191435392 |
| SNP7 | 139026152 | LUC7L2 | rs10265 | 0.191954741 |
| SNP11 | 17981047 | SERGEF | rs211146 | 0.192088531 |
| SNP4 | 437831 | ABCA11P | rs72501950 | 0.192572273 |
| SNP4 | 437831 | ABCA11P | rs145798212 | 0.192572273 |
| SNP15 | 40705225 | IVD | rs2229312 | 0.193640505 |
| SNP9 | 27561628 | C9orf72 | rs113299382 | 0.19463599 |
| SNP9 | 27561628 | C9orf72 | rs17769294 | 0.19463599 |
| SNP6 | 33768897 | MLN | rs2281820 | 0.194645852 |
| SNP6 | 33768897 | MLN | rs149413149 | 0.194645852 |
| SNP6 | 36270130 | PNPLA1 | rs12199580 | 0.194645852 |
| SNP6 | 36270130 | PNPLA1 | rs149324598 | 0.194645852 |
| SNP10 | 124214251 | ARMS2 | rs10490923 | 0.195047763 |
| SNP2 | 100916315 | LONRF2 | rs11123823 | 0.195892788 |
| SNP19 | 56413532 | NLRP13 | rs7258847 | 0.19651183 |
| SNP6 | 155451352 | TIAM2 | rs931312 | 0.196659359 |
| SNP17 | 4712617 | PLD2 | rs2286672 | 0.196970853 |
| SNP4 | 126241253 | FAT4 | rs7657251 | 0.197857668 |
| SNP2 | 231738168 | ITM2C | rs2289235 | 0.198049474 |
| SNP17 | 66364804 | ARSG | rs1558878 | 0.198234633 |
| SNP6 | 32713030 | HLA-DQA2 | rs3208181 | 0.199286765 |
| SNP6 | 32713030 | HLA-DQA2 | rs115339028 | 0.199286765 |
| SNP9 | 140638461 | EHMT1 | rs1129768 | 0.199649183 |
| SNP9 | 14722477 | CER1 | rs3747532 | 0.199755518 |
| SNP17 | 76817090 | USP36 | rs3744793 | 0.199935865 |
| SNP17 | 76817090 | USP36 | rs8065170 | 0.199935865 |
| SNP22 | 50694297 | MAPK12 | rs1129880 | 0.20001554 |
| SNP5 | 178416288 | GRM6 | rs2071246 | 0.200047707 |
| SNP10 | 71164655 | TACR2 | rs2229170 | 0.200287103 |
| SNP10 | 71164655 | TACR2 | rs191515098 | 0.200287103 |
| SNP1 | 1887245 | C1orf222 | rs28575980 | 0.201143589 |
| SNP15 | 35045057 | GJD2 | rs3743123 | 0.201186185 |
| SNP4 | 100510859 | MTTP | rs991811 | 0.201747428 |
| SNP10 | 88722398 | SNCG | rs9864 | 0.201874474 |
| SNP5 | 122359640 | PPIC | rs451195 | 0.202189541 |
| SNP5 | 122359640 | PPIC | rs137956160 | 0.202189541 |
| SNP4 | 90169925 | GPRIN3 | rs7653897 | 0.202699772 |
| SNP7 | 6370305 | C7orf70 | rs3750039 | 0.202699772 |
| SNP16 | 20328685 | GP2 | rs1129818 | 0.203306909 |
| SNP1 | 38289383 | MTF1 | rs2228272 | 0.203362529 |
| SNP1 | 38289383 | MTF1 | rs12751325 | 0.203362529 |
| SNP18 | 72343181 | ZNF407 | rs3794942 | 0.204040144 |
| SNP1 | 55119289 | HEATR8 | rs9332417 | 0.204694018 |
| SNP14 | 103440473 | CDC42BPB | rs8009219 | 0.205874154 |
| SNP8 | 22974450 | TNFRSF10C | rs9644063 | 0.207246575 |
| SNP20 | 42965863 | R3HDML | rs3746570 | 0.207790528 |
| SNP2 | 27550967 | GTF3C2 | rs1049817 | 0.208613303 |
| SNP19 | 39433299 | FBXO17 | rs8113389 | 0.208716718 |
| SNP3 | 145938619 | PLSCR4 | rs3762685 | 0.20997704 |
| SNP20 | 23546639 | CST9L | rs2295564 | 0.211096106 |
| SNP12 | 121017171 | POP5 | rs7174 | 0.211500521 |
| SNP1 | 159923111 | SLAMF9 | rs2789417 | 0.212068582 |
| SNP11 | 102495998 | MMP20 | rs2245803 | 0.212448837 |
| SNP19 | 1388538 | NDUFS7 | rs1142530 | 0.212763644 |
| SNP13 | 77738664 | MYCBP2 | rs2274547 | 0.212986553 |
| SNP11 | 117096652 | PCSK7 | rs2277287 | 0.214105126 |
| SNP8 | 6479178 | MCPH1 | rs2912016 | 0.214291468 |
| SNP15 | 68605169 | ITGA11 | rs4777035 | 0.216223037 |
| SNP5 | 162902516 | HMMR | rs299290 | 0.216232925 |
| SNP15 | 89402239 | ACAN | rs1042631 | 0.216420148 |
| SNP17 | 2595964 | KIAA0664 | rs2302199 | 0.217286867 |
| SNP5 | 146780273 | DPYSL3 | rs10515587 | 0.217671357 |
| SNP18 | 56202768 | ALPK2 | rs3809983 | 0.217682263 |
| SNP1 | 229623338 | NUP133 | rs1065674 | 0.218064706 |
| SNP1 | 229623338 | NUP133 | rs142513540 | 0.218064706 |
| SNP2 | 218940418 | RUFY4 | rs4674246 | 0.218581126 |
| SNP15 | 85333953 | ZNF592 | rs2241645 | 0.218686118 |
| SNP8 | 73982161 | C8orf84 | rs2291219 | 0.219130507 |
| SNP20 | 8737734 | PLCB1 | rs2076413 | 0.219367325 |
| SNP9 | 117853022 | TNC | rs944510 | 0.220032132 |
| SNP17 | 29553485 | NF1 | rs2285892 | 0.220616952 |
| SNP14 | 88477413 | GPR65 | rs6574978 | 0.221578336 |
| SNP17 | 39767744 | KRT16 | rs4796681 | 0.221634141 |
| SNP19 | 6677989 | C3 | rs17030 | 0.221730412 |
| SNP1 | 151867560 | THEM4 | rs3762427 | 0.22247822 |
| SNP1 | 151867560 | THEM4 | rs77308366 | 0.22247822 |
| SNP17 | 79166384 | AZI1 | rs3744150 | 0.223601753 |
| SNP9 | 93641199 | SYK | rs2306040 | 0.223601753 |
| SNP14 | 96871104 | AK7 | rs2275554 | 0.224212218 |
| SNP19 | 12541532 | ZNF443 | rs10422063 | 0.224630886 |
| SNP19 | 2733250 | SLC39A3 | rs61736898 | 0.22496948 |
| SNP9 | 79318921 | PRUNE2 | rs620985 | 0.225959118 |
| SNP5 | 176520243 | FGFR4 | rs351855 | 0.22706016 |
| SNP7 | 123514896 | HYAL4 | rs6949082 | 0.227344051 |
| SNP1 | 16271260 | ZBTB17 | rs9661939 | 0.227600609 |
| SNP2 | 130897218 | CCDC74B | rs13006246 | 0.227784671 |
| SNP21 | 35237608 | ITSN1 | rs9976801 | 0.227844825 |
| SNP20 | 10633237 | JAG1 | rs1131695 | 0.228027711 |
| SNP12 | 122186268 | TMEM120B | rs28651018 | 0.228636765 |
| SNP5 | 96076487 | CAST | rs7724759 | 0.229475768 |
| SNP5 | 112349070 | DCP2 | rs9326869 | 0.229871549 |
| SNP14 | 64908845 | MTHFD1 | rs2236225 | 0.230547805 |
| SNP14 | 102808330 | ZNF839 | rs12590618 | 0.231176226 |
| SNP6 | 62887099 | KHDRBS2 | rs6921170 | 0.231790207 |
| SNP9 | 130984755 | DNM1 | rs3003609 | 0.23270961 |
| SNP1 | 27679797 | SYTL1 | rs3813795 | 0.232867261 |
| SNP8 | 23060256 | TNFRSF10A | rs6557634 | 0.233076451 |
| SNP11 | 65790527 | CATSPER1 | rs2845570 | 0.233988808 |
| SNP10 | 124009124 | TACC2 | rs56075957 | 0.234718779 |
| SNP16 | 20552075 | ACSM2B | rs16970280 | 0.235057098 |
| SNP1 | 167096931 | DUSP27 | rs267746 | 0.235466643 |
| SNP7 | 30915263 | AQP1 | rs10216063 | 0.236469247 |
| SNP16 | 961051 | LMF1 | rs2277893 | 0.236744274 |
| SNP17 | 40322252 | KCNH4 | rs939881 | 0.237708877 |
| SNP2 | 54482964 | TSPYL6 | rs843704 | 0.238222498 |
| SNP10 | 123903133 | TACC2 | rs12765679 | 0.238632838 |
| SNP14 | 90730071 | PSMC1 | rs4811 | 0.238676473 |
| SNP14 | 90730071 | PSMC1 | rs11554757 | 0.238676473 |
| SNP6 | 31324562 | HLA-B | rs1050564 | 0.239458302 |
| SNP6 | 31324562 | HLA-B | rs144389754 | 0.239458302 |
| SNP6 | 31324562 | HLA-B | rs41551516 | 0.239458302 |
| SNP11 | 40137543 | LRRC4C | rs2953310 | 0.239471044 |
| SNP3 | 97806616 | OR5AC2 | rs4518168 | 0.239503428 |
| SNP5 | 96124330 | ERAP1 | rs30187 | 0.239597257 |
| SNP3 | 126268918 | C3orf22 | rs869463 | 0.239864772 |
| SNP6 | 54219326 | TINAG | rs3195579 | 0.240124082 |
| SNP20 | 36946848 | BPI | rs4358188 | 0.240255409 |
| SNP21 | 45708277 | AIRE | rs878081 | 0.242760672 |
| SNP21 | 45708277 | AIRE | rs148012328 | 0.242760672 |
| SNP3 | 32995928 | CCR4 | rs2228428 | 0.243499744 |
| SNP2 | 217012901 | XRCC5 | rs207906 | 0.244582002 |
| SNP4 | 88534235 | DSPP | rs2736982 | 0.244841627 |
| SNP2 | 36704144 | CRIM1 | rs848547 | 0.245846635 |
| SNP3 | 184046470 | EIF4G1 | rs2230571 | 0.246614774 |
| SNP19 | 41381683 | CYP2A7 | rs58682606 | 0.246986987 |
| SNP19 | 41381683 | CYP2A7 | rs71358943 | 0.246986987 |
| SNP18 | 60237388 | ZCCHC2 | rs8096750 | 0.247859393 |
| SNP10 | 124402677 | DMBT1 | rs1052715 | 0.248013755 |
| SNP5 | 176867943 | GRK6 | rs335435 | 0.249965856 |
| SNP20 | 1293247 | SDCBP2 | rs2273959 | 0.251186718 |
| SNP20 | 1293247 | SDCBP2 | rs151023134 | 0.251186718 |
| SNP21 | 46011468 | KRTAP10-6 | rs465279 | 0.251192583 |
| SNP8 | 38848850 | | rs4733895 | 0.251241944 |
| SNP14 | 24653954 | IPO4 | rs7146310 | 0.252340036 |
| SNP4 | 3234980 | HTT | rs362272 | 0.252340036 |
| SNP19 | 8154990 | FBN3 | rs35002391 | 0.252490377 |
| SNP1 | 171511039 | PRRC2C | rs1687064 | 0.252768886 |
| SNP19 | 3767265 | MRPL54 | rs7239 | 0.254991374 |
| SNP4 | 170663235 | C4orf27 | rs1047642 | 0.255263257 |
| SNP20 | 31596472 | BPIL1 | rs6088066 | 0.256195241 |
| SNP16 | 84070500 | SLC38A8 | rs1317524 | 0.256267686 |
| SNP17 | 7722365 | DNAH2 | rs7213894 | 0.25666015 |
| SNP17 | 36894839 | PCGF2 | rs1138349 | 0.257213536 |
| SNP6 | 118886961 | C6orf204 | rs3734382 | 0.25732181 |
| SNP17 | 16342833 | | rs11540320 | 0.258066026 |
| SNP17 | 16342833 | | rs11871958 | 0.258066026 |
| SNP20 | 31812923 | C20orf71 | rs3818222 | 0.258825405 |
| SNP2 | 214012508 | IKZF2 | rs6709554 | 0.259050619 |
| SNP3 | 170078232 | SKIL | rs3772173 | 0.25946066 |
| SNP19 | 53911510 | ZNF765 | rs8182488 | 0.259544786 |
| SNP20 | 9288522 | PLCB4 | rs6077510 | 0.259545483 |
| SNP20 | 9288522 | PLCB4 | rs145938987 | 0.259545483 |
| SNP7 | 5347914 | TNRC18 | rs11554710 | 0.259970034 |
| SNP17 | 2091765 | SMG6 | rs903160 | 0.259971474 |
| SNP6 | 167549775 | CCR6 | rs3093007 | 0.261970298 |
| SNP16 | 3254972 | OR1F1 | rs2075851 | 0.26210288 |
| SNP17 | 48433958 | XYLT2 | rs4794136 | 0.262158929 |
| SNP19 | 49878115 | DKKL1 | rs1054770 | 0.262703722 |
| SNP11 | 58190136 | OR5B2 | rs4298923 | 0.263212945 |
| SNP11 | 10786175 | CTR9 | rs7118399 | 0.264224345 |
| SNP6 | 160679400 | SLC22A2 | rs624249 | 0.265228253 |
| SNP3 | 97660106 | CRYBG3 | rs4857302 | 0.265357636 |
| SNP21 | 47974582 | DIP2A | rs1107065 | 0.265726052 |
| SNP8 | 145059425 | PARP10 | rs11136344 | 0.265726052 |
| SNP6 | 24653376 | TDP2 | rs1129644 | 0.267000573 |
| SNP19 | 16872940 | NWD1 | rs773852 | 0.267613884 |
| SNP7 | 89982132 | GTPBP10 | rs6972561 | 0.269831327 |
| SNP20 | 55012318 | CASS4 | rs911159 | 0.270099187 |
| SNP19 | 49438363 | DHDH | rs2270941 | 0.271273953 |
| SNP4 | 6619204 | MAN2B2 | rs4689490 | 0.271623808 |
| SNP1 | 247875608 | OR6F1 | rs6587382 | 0.272969206 |
| SNP13 | 32911888 | BRCA2 | rs1801406 | 0.272971319 |
| SNP19 | 53117531 | ZNF83 | rs1056185 | 0.273151587 |
| SNP19 | 53117809 | ZNF83 | rs10406458 | 0.273151587 |
| SNP12 | 4459032 | C12orf5 | rs7309402 | 0.27342094 |
| SNP1 | 53792651 | LRP8 | rs3820198 | 0.274152958 |
| SNP18 | 43316538 | SLC14A1 | rs2298718 | 0.274782766 |
| SNP13 | 23929095 | SACS | rs1536365 | 0.275829416 |
| SNP13 | 23930055 | SACS | rs2031640 | 0.275829416 |
| SNP15 | 100649248 | ADAMTS17 | rs61752832 | 0.27603128 |
| SNP12 | 94645255 | PLXNC1 | rs2230757 | 0.276487175 |
| SNP16 | 3170188 | ZNF205 | rs12032 | 0.276493732 |
| SNP11 | 107299631 | CWF19L2 | rs659040 | 0.276932369 |
| SNP15 | 86940622 | AGBL1 | rs4362360 | 0.277937278 |
| SNP4 | 164272703 | NPY5R | rs11946004 | 0.277988248 |
| SNP1 | 171751236 | METTL13 | rs2294720 | 0.278515721 |
| SNP11 | 2167543 | IGF2AS | rs1003483 | 0.278572462 |
| SNP8 | 28384712 | FZD3 | rs2241802 | 0.280800745 |
| SNP13 | 40229957 | COG6 | rs3812883 | 0.280972768 |
| SNP3 | 15737689 | ANKRD28 | rs2470548 | 0.281612911 |
| SNP8 | 145756170 | ARHGAP39 | rs873884 | 0.282481587 |
| SNP8 | 145756170 | ARHGAP39 | rs143245385 | 0.282481587 |
| SNP20 | 31386347 | DNMT3B | rs6058891 | 0.282851111 |
| SNP11 | 5877979 | OR52E8 | rs12419602 | 0.2831574 |
| SNP19 | 57037103 | ZNF471 | rs16987303 | 0.283487851 |
| SNP3 | 113286405 | SIDT1 | rs2292511 | 0.284689905 |
| SNP2 | 205829991 | PARD3B | rs236843 | 0.285605204 |
| SNP16 | 84691044 | KLHL36 | rs3751762 | 0.286141822 |
| SNP21 | 26965148 | MRPL39 | rs1135638 | 0.286584735 |
| SNP21 | 26965205 | MRPL39 | rs1057885 | 0.286584735 |
| SNP10 | 115405615 | NRAP | rs3127106 | 0.2866084 |
| SNP10 | 115405664 | NRAP | rs3121478 | 0.2866084 |
| SNP18 | 29432624 | TRAPPC8 | rs3737374 | 0.288959185 |
| SNP11 | 63713317 | NAA40 | rs3740637 | 0.290666807 |
| SNP14 | 71199452 | MAP3K9 | rs3829955 | 0.290821209 |
| SNP5 | 73163831 | RP11-428C6.1 | rs2931423 | 0.291119907 |
| SNP9 | 111641825 | IKBKAP | rs1538660 | 0.291299859 |
| SNP9 | 111651620 | IKBKAP | rs3204145 | 0.291299859 |
| SNP9 | 111651620 | IKBKAP | rs140024352 | 0.291299859 |
| SNP3 | 178546026 | KCNMB2 | rs9831934 | 0.292725544 |
| SNP19 | 8191184 | FBN3 | rs35025963 | 0.295528377 |
| SNP19 | 8191184 | FBN3 | rs147627738 | 0.295528377 |
| SNP14 | 106174261 | | rs1407 | 0.296113011 |
| SNP9 | 126132919 | CRB2 | rs33984675 | 0.297408307 |
| SNP2 | 48602252 | FOXN2 | rs17855177 | 0.297478637 |
| SNP5 | 64881936 | PPWD1 | rs27141 | 0.297747609 |
| SNP21 | 37833751 | CLDN14 | rs219779 | 0.298192996 |
| SNP16 | 4033436 | ADCY9 | rs2230739 | 0.298470509 |
| SNP6 | 30075843 | TRIM31 | rs3132679 | 0.300039561 |
| SNP6 | 30075843 | TRIM31 | rs116373424 | 0.300039561 |
| SNP6 | 30075843 | TRIM31 | rs113793434 | 0.300039561 |
| SNP1 | 20490518 | PLA2G2C | rs6426616 | 0.300725313 |
| SNP4 | 138449683 | PCDH18 | rs10018837 | 0.301094834 |
| SNP9 | 131565554 | TBC1D13 | rs1572912 | 0.301967375 |
| SNP17 | 18630995 | TRIM16L | rs8075739 | 0.302651332 |
| SNP17 | 37243927 | PLXDC1 | rs12602945 | 0.304050954 |
| SNP13 | 113527967 | ATP11A | rs1320525 | 0.304397096 |
| SNP14 | 24505722 | DHRS4L1 | rs8005834 | 0.304918823 |
| SNP1 | 226924642 | ITPKB | rs3754415 | 0.306060006 |
| SNP5 | 13829799 | DNAH5 | rs1348689 | 0.306078778 |
| SNP3 | 66444615 | LRIG1 | rs6793110 | 0.306583079 |
| SNP10 | 100190920 | HPS1 | rs1801287 | 0.308353801 |
| SNP2 | 48808152 | STON1 | rs940389 | 0.308881927 |
| SNP20 | 50405502 | SALL4 | rs17802735 | 0.309704832 |
| SNP1 | 26694245 | ZNF683 | rs10794531 | 0.310405292 |
| SNP17 | 72349040 | KIF19 | rs9897137 | 0.310518647 |
| SNP17 | 65104666 | HELZ | rs11657929 | 0.310690907 |
| SNP18 | 67345034 | DOK6 | rs8099030 | 0.312722752 |
| SNP19 | 804327 | PTBP1 | rs10420953 | 0.313005698 |
| SNP12 | 12247616 | BCL2L14 | rs11054683 | 0.313228847 |
| SNP11 | 57137371 | P2RX3 | rs2276039 | 0.313730978 |
| SNP1 | 77759578 | AK5 | rs2815311 | 0.314089482 |
| SNP1 | 201358304 | LAD1 | rs3738281 | 0.315452741 |
| SNP2 | 32983526 | TTC27 | rs2273665 | 0.316331311 |
| SNP1 | 3807388 | C1orf174 | rs4131373 | 0.316354041 |
| SNP19 | 37854580 | HKR1 | rs3745764 | 0.316571578 |
| SNP5 | 146619206 | STK32A | rs4705132 | 0.317431762 |
| SNP3 | 49836707 | CDHR4 | rs7629936 | 0.319458236 |
| SNP2 | 131520178 | FAM123C | rs77687733 | 0.319926251 |
| SNP19 | 56392920 | NLRP4 | rs12975929 | 0.322029315 |
| SNP19 | 48621036 | LIG1 | rs13436 | 0.322746481 |
| SNP11 | 22242729 | ANO5 | rs4312063 | 0.323446796 |
| SNP1 | 37325477 | GRIK3 | rs6691840 | 0.323557394 |
| SNP21 | 30341891 | LTN1 | rs2254796 | 0.323681197 |
| SNP20 | 44238741 | WFDC9 | rs2245898 | 0.324195856 |
| SNP4 | 7802292 | AFAP1 | rs11728778 | 0.324370868 |
| SNP17 | 27889578 | | rs539307 | 0.324993085 |
| SNP7 | 73969541 | GTF2IRD1 | rs2301895 | 0.325868556 |
| SNP4 | 187179210 | KLKB1 | rs925453 | 0.327663935 |
| SNP8 | 23423758 | SLC25A37 | rs10992 | 0.328417879 |
| SNP13 | 109661359 | MYO16 | rs3825491 | 0.32890495 |
| SNP10 | 123846485 | TACC2 | rs4751871 | 0.329516948 |
| SNP10 | 123846860 | TACC2 | rs11599291 | 0.329516948 |
| SNP1 | 18023509 | ARHGEF10L | rs2270977 | 0.32954105 |
| SNP22 | 21348914 | LZTR1 | rs4822790 | 0.329730286 |
| SNP19 | 48715153 | CARD8 | rs3745718 | 0.329951711 |
| SNP19 | 36048741 | ATP4A | rs2230181 | 0.330088015 |
| SNP4 | 141600320 | TBC1D9 | rs4956329 | 0.331266677 |
| SNP17 | 72368550 | GPR142 | rs11658891 | 0.332078561 |
| SNP16 | 11852354 | ZC3H7A | rs8743 | 0.334821608 |
| SNP10 | 13230950 | MCM10 | rs2296222 | 0.336576799 |
| SNP12 | 60173356 | SLC16A7 | rs3763980 | 0.336584411 |
| SNP2 | 171910313 | TLK1 | rs11553951 | 0.337146823 |
| SNP10 | 102746503 | MRPL43 | rs2863095 | 0.33812384 |
| SNP16 | 56396486 | AMFR | rs4924 | 0.33885706 |
| SNP1 | 227935444 | SNAP47 | rs2236359 | 0.340975817 |
| SNP19 | 18285944 | IFI30 | rs11554159 | 0.341628477 |
| SNP3 | 46501213 | LTF | rs1126478 | 0.34179197 |
| SNP2 | 228883721 | SPHKAP | rs3811514 | 0.341908211 |
| SNP20 | 45853037 | ZMYND8 | rs2664544 | 0.342438963 |
| SNP21 | 46705621 | POFUT2 | rs2297285 | 0.342685384 |
| SNP14 | 35592538 | KIAA0391 | rs941653 | 0.344088212 |
| SNP12 | 75884254 | GLIPR1 | rs28932170 | 0.345343521 |
| SNP7 | 128038555 | IMPDH1 | rs2288550 | 0.345780711 |
| SNP19 | 57325083 | ZIM2 | rs34051133 | 0.345978987 |
| SNP1 | 9324213 | H6PD | rs17368528 | 0.34714901 |
| SNP4 | 70807771 | CSN1S1 | rs10030475 | 0.347328861 |
| SNP1 | 75172001 | CRYZ | rs7527057 | 0.348326886 |
| SNP7 | 129330353 | NRF1 | rs3735006 | 0.348469257 |
| SNP15 | 89401109 | ACAN | rs4932439 | 0.348621316 |
| SNP10 | 73574843 | CDH23 | rs2290021 | 0.34929567 |
| SNP1 | 1849529 | TMEM52 | rs28640257 | 0.350957986 |
| SNP1 | 1849529 | TMEM52 | rs4459050 | 0.350957986 |
| SNP1 | 236413230 | ERO1LB | rs2477599 | 0.351177936 |
| SNP8 | 135649848 | ZFAT | rs12541381 | 0.351188893 |
| SNP19 | 46812451 | HIF3A | rs61750957 | 0.352636827 |
| SNP19 | 46812451 | HIF3A | rs145463725 | 0.352636827 |
| SNP7 | 154862652 | HTR5A | rs11575874 | 0.353644063 |
| SNP14 | 33046388 | AKAP6 | rs1950703 | 0.354162749 |
| SNP20 | 21142813 | | rs2236178 | 0.354471056 |
| SNP7 | 154429560 | DPP6 | rs11243339 | 0.356308887 |
| SNP19 | 36224705 | AD000671.1 | rs231591 | 0.357732846 |
| SNP10 | 50532588 | C10orf71 | rs10857469 | 0.358053279 |
| SNP14 | 68045935 | PLEKHH1 | rs6573781 | 0.360033713 |
| SNP5 | 150647012 | GM2A | rs1048723 | 0.360610415 |
| SNP17 | 74056413 | SRP68 | rs2665998 | 0.360703876 |
| SNP16 | 55536727 | MMP2 | rs14070 | 0.360710435 |
| SNP19 | 18047283 | CCDC124 | rs4808722 | 0.362878798 |
| SNP3 | 191179193 | PYDC2 | rs293833 | 0.363134654 |
| SNP1 | 22835677 | ZBTB40 | rs209729 | 0.363497739 |
| SNP4 | 11401012 | HS3ST1 | rs1047385 | 0.363753845 |
| SNP4 | 11401087 | HS3ST1 | rs1047389 | 0.363753845 |
| SNP2 | 220371035 | GMPPA | rs1046474 | 0.365515404 |
| SNP16 | 25268368 | ZKSCAN2 | rs4787310 | 0.366305996 |
| SNP12 | 104645405 | TXNRD1 | rs7975161 | 0.367096314 |
| SNP22 | 45821887 | RIBC2 | rs1022477 | 0.368922312 |
| SNP1 | 173878832 | SERPINC1 | rs5878 | 0.369285468 |
| SNP17 | 21826256 | FAM27L | rs9989450 | 0.371409206 |
| SNP20 | 4880308 | SLC23A2 | rs1776964 | 0.372061915 |
| SNP2 | 30976058 | CAPN13 | rs2926304 | 0.372373423 |
| SNP11 | 63487386 | RTN3 | rs542998 | 0.372453014 |
| SNP3 | 33138549 | GLB1 | rs7637099 | 0.37315226 |
| SNP9 | 88959938 | ZCCHC6 | rs791323 | 0.373251741 |
| SNP2 | 127453664 | GYPC | rs1050967 | 0.373365719 |
| SNP10 | 53458047 | CSTF2T | rs3740228 | 0.374498538 |
| SNP19 | 11304498 | KANK2 | rs755238 | 0.375707396 |
| SNP14 | 96157187 | TCL1B | rs1064017 | 0.376770732 |
| SNP11 | 12225946 | MICAL2 | rs3763823 | 0.377257094 |
| SNP6 | 108068003 | SCML4 | rs6934505 | 0.377622123 |
| SNP19 | 39322087 | ECH1 | rs9419 | 0.378992477 |
| SNP19 | 11890921 | ZNF441 | rs33949590 | 0.379862215 |
| SNP7 | 123152035 | IQUB | rs10270705 | 0.379903798 |
| SNP21 | 26969703 | MRPL39 | rs1135618 | 0.380542689 |
| SNP17 | 11651057 | DNAH9 | rs3744581 | 0.380923177 |
| SNP3 | 4856234 | ITPR1 | rs901854 | 0.382441986 |
| SNP3 | 148872987 | HPS3 | rs6440589 | 0.382849756 |
| SNP2 | 68753240 | APLF | rs35002937 | 0.384135167 |
| SNP19 | 35506729 | GRAMD1A | rs2290647 | 0.385382361 |
| SNP2 | 26798893 | C2orf70 | rs13002673 | 0.38614609 |
| SNP17 | 39883595 | HAP1 | rs7216154 | 0.386285413 |
| SNP17 | 39883595 | HAP1 | rs143134174 | 0.386285413 |
| SNP5 | 115341638 | AC010282.1 | rs10078759 | 0.388302067 |
| SNP11 | 6942476 | OR2D3 | rs10839658 | 0.38955615 |
| SNP2 | 233757679 | NGEF | rs748002 | 0.389858188 |
| SNP4 | 140811135 | MAML3 | rs3733382 | 0.390714495 |
| SNP6 | 116574455 | TSPYL4 | rs2232472 | 0.391210017 |
| SNP14 | 101429459 | | rs72700530 | 0.393335487 |
| SNP7 | 99971313 | PILRA | rs2405442 | 0.393447331 |
| SNP20 | 56099114 | CTCFL | rs6070128 | 0.39380423 |
| SNP1 | 17402255 | PADI2 | rs3818032 | 0.394806 |
| SNP14 | 96707457 | BDKRB2 | rs5224 | 0.395044313 |
| SNP6 | 166889281 | RPS6KA2 | rs4373344 | 0.395766043 |
| SNP2 | 162929979 | DPP4 | rs17574 | 0.39693818 |
| SNP5 | 141391532 | GNPDA1 | rs164080 | 0.398813917 |
| SNP1 | 75681511 | SLC44A5 | rs595009 | 0.398966855 |
| SNP1 | 234041422 | SLC35F3 | rs57010808 | 0.399959454 |
| SNP5 | 1880891 | IRX4 | rs2232376 | 0.400557913 |
| SNP22 | 36922052 | EIF3D | rs9622410 | 0.403338294 |
| SNP6 | 166571935 | T | rs35819705 | 0.404732076 |
| SNP6 | 42932202 | PEX6 | | 0.407443812 |
| SNP18 | 21441717 | LAMA3 | rs12965685 | 0.407462381 |
| SNP17 | 6733672 | TEKT1 | rs8078571 | 0.407818052 |
| SNP17 | 38938591 | KRT27 | rs2469826 | 0.409901285 |
| SNP8 | 124121798 | WDR67 | rs16897969 | 0.410593697 |
| SNP17 | 65362552 | PSMD12 | rs11079691 | 0.4124315 |
| SNP14 | 22133416 | OR4E2 | rs12717305 | 0.413309186 |
| SNP16 | 3811556 | | rs886528 | 0.414817672 |
| SNP8 | 26441477 | DPYSL2 | rs11786691 | 0.417890097 |
| SNP22 | 23503170 | RAB36 | rs5759612 | 0.419410384 |
| SNP12 | 109994870 | MMAB | rs9593 | 0.419591792 |
| SNP13 | 110818598 | COL4A1 | rs3742207 | 0.419756349 |
| SNP2 | 31589847 | XDH | rs2295475 | 0.420988482 |
| SNP2 | 31589847 | XDH | rs72549366 | 0.420988482 |
| SNP17 | 42449789 | ITGA2B | rs5910 | 0.424931548 |
| SNP8 | 16907125 | | rs2014286 | 0.425964062 |
| SNP11 | 607175 | PHRF1 | rs936468 | 0.426348096 |
| SNP12 | 12871099 | CDKN1B | rs2066827 | 0.426348096 |
| SNP1 | 151733335 | MRPL9 | rs8480 | 0.427779935 |
| SNP11 | 117163824 | BACE1 | rs638405 | 0.428744088 |
| SNP11 | 110035301 | ZC3H12C | rs1026607 | 0.429164245 |
| SNP5 | 179740827 | GFPT2 | rs2303007 | 0.429220075 |
| SNP19 | 35649281 | FXYD5 | rs1128882 | 0.429793838 |
| SNP8 | 39872935 | IDO2 | rs4503083 | 0.430091939 |
| SNP22 | 17446991 | GAB4 | rs4819925 | 0.431328123 |
| SNP7 | 18767343 | HDAC9 | rs1178127 | 0.431328123 |
| SNP1 | 220154768 | EPRS | rs1061160 | 0.431680475 |
| SNP20 | 61288038 | SLCO4A1 | rs1047099 | 0.432982335 |
| SNP17 | 4638737 | CXCL16 | rs1050998 | 0.433328591 |
| SNP15 | 57835903 | CGNL1 | rs1620402 | 0.435414729 |
| SNP8 | 37730456 | RAB11FIP1 | rs7341564 | 0.435445449 |
| SNP5 | 176378574 | UIMC1 | rs365132 | 0.435490567 |
| SNP20 | 33565755 | MYH7B | rs11906160 | 0.436543561 |
| SNP2 | 65299300 | CEP68 | rs2723089 | 0.437343347 |
| SNP19 | 49621964 | C19orf73 | rs2232003 | 0.438682088 |
| SNP19 | 49621964 | C19orf73 | rs186208217 | 0.438682088 |
| SNP17 | 42925559 | HIGD1B | rs12164 | 0.439173138 |
| SNP19 | 52620269 | ZNF616 | rs35582075 | 0.441980592 |
| SNP15 | 25925094 | ATP10A | rs3816800 | 0.443511716 |
| SNP15 | 25925094 | ATP10A | rs138394113 | 0.443511716 |
| SNP16 | 16228242 | ABCC1 | rs2230671 | 0.443956434 |
| SNP6 | 158505088 | SYNJ2 | rs1744173 | 0.444587722 |
| SNP1 | 176992553 | ASTN1 | rs172917 | 0.452105963 |
| SNP11 | 11292700 | GALNTL4 | rs10831567 | 0.452180485 |
| SNP6 | 152683413 | SYNE1 | rs4407724 | 0.452326779 |
| SNP2 | 121989489 | TFCP2L1 | rs2304667 | 0.454803162 |
| SNP10 | 70045157 | PBLD | rs35708572 | 0.454929199 |
| SNP5 | 180043388 | FLT4 | rs1130378 | 0.455353431 |
| SNP11 | 61897359 | INCENP | rs1675133 | 0.456091323 |
| SNP3 | 108703587 | MORC1 | rs2290057 | 0.458435248 |
| SNP11 | 5141902 | OR52A4 | rs10837374 | 0.462523048 |
| SNP6 | 169637763 | THBS2 | rs9766671 | 0.464418469 |
| SNP6 | 35477032 | TULP1 | rs2064317 | 0.46565902 |
| SNP6 | 83075914 | TPBG | rs700494 | 0.467459593 |
| SNP17 | 79095144 | AATK | rs8073904 | 0.467865857 |
| SNP10 | 93608142 | TNKS2 | rs3758499 | 0.468564212 |
| SNP11 | 124793682 | HEPACAM | rs10790715 | 0.474871234 |
| SNP14 | 22038210 | OR10G3 | rs17792766 | 0.478074824 |
| SNP12 | 104709559 | TXNRD1 | rs4964287 | 0.479633338 |
| SNP1 | 21795388 | NBPF3 | rs1827293 | 0.479733259 |
| SNP2 | 240961728 | NDUFA10 | rs2083411 | 0.485309229 |
| SNP19 | 51381777 | KLK2 | rs198977 | 0.486683436 |
| SNP1 | 100316589 | | rs2307130 | 0.487065559 |
| SNP8 | 39091526 | ADAM32 | rs4515515 | 0.487838177 |
| SNP11 | 6631016 | ILK | rs2292195 | 0.488233615 |
| SNP18 | 9887394 | TXNDC2 | rs2240910 | 0.488614967 |
| SNP9 | 103054951 | INVS | rs2787374 | 0.488706208 |
| SNP7 | 92985252 | CCDC132 | rs2374639 | 0.490880993 |
| SNP16 | 55880534 | CES5A | rs11860488 | 0.492239522 |
| SNP1 | 43296173 | ERMAP | rs33950227 | 0.495083231 |
| SNP5 | 137892170 | HSPA9 | rs10117 | 0.496441667 |
| SNP10 | 11805339 | ECHDC3 | rs17850531 | 0.49910912 |
| SNP9 | 125562527 | OR1K1 | rs10985782 | 0.501594858 |
| SNP2 | 152476028 | NEB | rs10172023 | 0.50210898 |
| SNP1 | 201334382 | TNNT2 | rs3729547 | 0.5038732 |
| SNP6 | 127476516 | RSPO3 | rs1892172 | 0.505536938 |
| SNP19 | 42312933 | CEACAM3 | rs3752172 | 0.505750662 |
| SNP19 | 52660783 | ZNF836 | rs17696575 | 0.508425215 |
| SNP3 | 194063300 | CPN2 | rs62288099 | 0.510900154 |
| SNP19 | 58003613 | | rs2074072 | 0.511218992 |
| SNP19 | 58004346 | ZNF419 | rs2074076 | 0.511218992 |
| SNP19 | 12863429 | BEST2 | rs79300835 | 0.511350867 |
| SNP14 | 65253232 | SPTB | rs77806 | 0.511452735 |
| SNP17 | 77704912 | ENPP7 | rs8074547 | 0.511544449 |
| SNP1 | 39340282 | GJA9 | rs880303 | 0.514771248 |
| SNP1 | 39352271 | RHBDL2 | rs2147914 | 0.514771248 |
| SNP11 | 4703165 | OR51E2 | rs1123991 | 0.517372275 |
| SNP7 | 64452830 | ERV3 | rs34639489 | 0.521773858 |
| SNP2 | 17998331 | MSGN1 | rs13001625 | 0.521910208 |
| SNP12 | 13764774 | GRIN2B | rs1805482 | 0.523488812 |
| SNP6 | 158927651 | TULP4 | rs3828712 | 0.525883929 |
| SNP22 | 40814500 | MKL1 | rs878756 | 0.526116035 |
| SNP2 | 96795608 | ASTL | rs1657502 | 0.530086368 |
| SNP11 | 124793716 | HEPACAM | rs10790716 | 0.531737423 |
| SNP8 | 97614661 | SDC2 | rs1042381 | 0.532036412 |
| SNP14 | 94187832 | PRIMA1 | rs1887197 | 0.532856627 |
| SNP3 | 42728144 | KBTBD5 | rs6805421 | 0.536573167 |
| SNP1 | 19447843 | UBR4 | rs1044010 | 0.538956906 |
| SNP22 | 17589567 | IL17RA | rs879575 | 0.539703087 |
| SNP2 | 71417134 | PAIP2B | rs357776 | 0.542899417 |
| SNP1 | 151288172 | PI4KB | rs1056847 | 0.543761997 |
| SNP10 | 102824349 | KAZALD1 | rs807037 | 0.544007866 |
| SNP2 | 74300717 | TET3 | rs7560668 | 0.544952758 |
| SNP4 | 77637502 | SHROOM3 | rs344122 | 0.545930456 |
| SNP12 | 122079189 | ORAI1 | rs3741595 | 0.55086055 |
| SNP7 | 94898811 | PPP1R9A | rs854524 | 0.556582188 |
| SNP15 | 100252892 | MEF2A | rs34851361 | 0.556737073 |
| SNP15 | 100252892 | MEF2A | rs111748677 | 0.556737073 |
| SNP1 | 118644524 | SPAG17 | rs17185492 | 0.556771904 |
| SNP14 | 21551058 | ARHGEF40 | rs1958395 | 0.5571679 |
| SNP1 | 109486196 | CLCC1 | rs338466 | 0.557926705 |
| SNP22 | 39496336 | APOBEC3H | rs139293 | 0.55828053 |
| SNP1 | 230898494 | CAPN9 | rs2282319 | 0.562212116 |
| SNP6 | 33139328 | COL11A2 | rs1799910 | 0.565861034 |
| SNP6 | 33139328 | COL11A2 | rs143029688 | 0.565861034 |
| SNP19 | 282753 | PPAP2C | rs17425002 | 0.567494092 |
| SNP19 | 282753 | PPAP2C | rs1138439 | 0.567494092 |
| SNP18 | 55027334 | ST8SIA3 | rs1153626 | 0.567913001 |
| SNP18 | 55027334 | ST8SIA3 | rs112792874 | 0.567913001 |
| SNP19 | 3728609 | TJP3 | rs2067019 | 0.570709417 |
| SNP14 | 69521345 | DCAF5 | rs61741172 | 0.571882801 |
| SNP22 | 16449050 | OR11H1 | rs78350717 | 0.573216582 |
| SNP22 | 16449050 | OR11H1 | rs78804300 | 0.573216582 |
| SNP22 | 16449050 | OR11H1 | rs148001219 | 0.573216582 |
| SNP9 | 390512 | DOCK8 | rs2297075 | 0.574425191 |
| SNP1 | 57340727 | C8A | rs652785 | 0.576160157 |
| SNP7 | 128388648 | CALU | rs2290228 | 0.577957975 |
| SNP8 | 134108546 | TG | rs2069569 | 0.578711719 |
| SNP8 | 134108546 | TG | rs56541861 | 0.578711719 |
| SNP8 | 134108546 | TG | rs116726914 | 0.578711719 |
| SNP13 | 48671081 | | rs1571256 | 0.579779943 |
| SNP2 | 228883029 | SPHKAP | rs3811515 | 0.580076378 |
| SNP19 | 17351535 | NR2F6 | rs2288539 | 0.580816683 |
| SNP2 | 224824617 | MRPL44 | rs11546405 | 0.581803934 |
| SNP9 | 129854199 | ANGPTL2 | rs2297866 | 0.581822721 |
| SNP7 | 75051375 | POM121C | rs398016 | 0.583630237 |
| SNP20 | 48491258 | SLC9A8 | rs542234 | 0.587045982 |
| SNP17 | 15341183 | FAM18B2 | rs2954759 | 0.590674096 |
| SNP2 | 29001691 | PPP1CB | rs1128416 | 0.590821652 |
| SNP15 | 91083353 | CRTC3 | rs8033595 | 0.593215422 |
| SNP12 | 117669914 | NOS1 | rs3741475 | 0.599568669 |
| SNP11 | 117222592 | CEP164 | rs490262 | 0.599630333 |
| SNP19 | 36207510 | ZBTB32 | rs1052491 | 0.601238505 |
| SNP19 | 6755007 | SH2D3A | rs2305806 | 0.605314834 |
| SNP6 | 144869785 | UTRN | rs1534443 | 0.608856161 |
| SNP12 | 5603632 | NTF3 | rs6332 | 0.610157845 |
| SNP22 | 22730634 | | rs2009587 | 0.612482709 |
| SNP22 | 22730634 | | rs61731372 | 0.612482709 |
| SNP2 | 205986321 | PARD3B | rs2289023 | 0.617996972 |
| SNP22 | 31529463 | INPP5J | rs35342535 | 0.620389665 |
| SNP11 | 33680371 | C11orf41 | rs12280103 | 0.620774168 |
| SNP1 | 207238419 | PFKFB2 | rs72741390 | 0.624290875 |
| SNP12 | 48723324 | H1FNT | rs2732441 | 0.625593006 |
| SNP1 | 145562087 | ANKRD35 | rs41315701 | 0.626623492 |
| SNP2 | 217148417 | 4-Mar | rs876771 | 0.629508843 |
| SNP3 | 183976241 | ECE2 | rs6767237 | 0.635699343 |
| SNP7 | 38289121 | | rs1053756 | 0.641015783 |
| SNP10 | 63170292 | TMEM26 | rs7083475 | 0.641249974 |
| SNP14 | 39703324 | MIA2 | rs7141840 | 0.642420016 |
| SNP14 | 72117156 | SIPA1L1 | rs8017465 | 0.643689097 |
| SNP1 | 177001896 | ASTN1 | rs2076069 | 0.646055406 |
| SNP11 | 10650329 | MRVI1 | rs2278125 | 0.647190572 |
| SNP1 | 17668563 | PADI4 | rs35903413 | 0.648548514 |
| SNP9 | 116931666 | COL27A1 | rs2567705 | 0.648807831 |
| SNP1 | 66831370 | PDE4B | rs783036 | 0.64954007 |
| SNP19 | 2078176 | MOBKL2A | rs35452475 | 0.649676101 |
| SNP2 | 8871342 | KIDINS220 | rs1044280 | 0.650160785 |
| SNP11 | 3129027 | OSBPL5 | rs3741350 | 0.651744032 |
| SNP1 | 248525780 | OR2T4 | rs28698997 | 0.652706933 |
| SNP1 | 248525780 | OR2T4 | rs77428015 | 0.652706933 |
| SNP6 | 99893938 | USP45 | rs4504482 | 0.656163736 |
| SNP1 | 68903942 | RPE65 | rs12145904 | 0.658030894 |
| SNP11 | 69934085 | ANO1 | rs10898112 | 0.660280003 |
| SNP22 | 44489896 | PARVB | rs738479 | 0.661955099 |
| SNP22 | 44489896 | PARVB | rs187361838 | 0.661955099 |
| SNP12 | 53294381 | KRT8 | rs8608 | 0.668782502 |
| SNP6 | 101296389 | ASCC3 | rs9390698 | 0.669275896 |
| SNP2 | 240981375 | PRR21 | rs6732185 | 0.669850329 |
| SNP20 | 60768497 | GTPBP5 | rs2236527 | 0.672878237 |
| SNP4 | 13606576 | BOD1L | rs1971278 | 0.675702764 |
| SNP7 | 150491084 | TMEM176B | rs3173833 | 0.676817514 |
| SNP15 | 78585106 | WDR61 | rs2280364 | 0.679411593 |
| SNP11 | 103229027 | DYNC2H1 | rs2566913 | 0.680742694 |
| SNP10 | 134218269 | PWWP2B | rs11146363 | 0.685206165 |
| SNP18 | 67365668 | DOK6 | rs4426448 | 0.685841569 |
| SNP3 | 108081277 | HHLA2 | rs6779094 | 0.688343631 |
| SNP13 | 101287404 | TMTC4 | rs2297943 | 0.690001227 |
| SNP1 | 7723957 | CAMTA1 | rs12128526 | 0.690135057 |
| SNP10 | 86004873 | RGR | rs2279227 | 0.690218396 |
| SNP12 | 9020489 | A2ML1 | rs7308811 | 0.692135807 |
| SNP15 | 45779810 | SLC30A4 | rs2453531 | 0.692666048 |
| SNP7 | 31594508 | CCDC129 | rs2286711 | 0.696686527 |
| SNP13 | 100518580 | CLYBL | rs3783185 | 0.697367673 |
| SNP8 | 63951312 | GGH | rs1800909 | 0.702737614 |
| SNP4 | 156276289 | MAP9 | rs2341894 | 0.704097707 |
| SNP16 | 70602221 | SF3B3 | rs12909 | 0.706713481 |
| SNP19 | 8321946 | LASS4 | rs36247 | 0.708661671 |
| SNP11 | 117869670 | IL10RA | rs2229113 | 0.715577626 |
| SNP17 | 34416537 | CCL3 | rs1130371 | 0.717882901 |
| SNP3 | 14175262 | TMEM43 | rs2340917 | 0.722539465 |
| SNP19 | 11105608 | SMARCA4 | rs7935 | 0.725963887 |
| SNP19 | 1233597 | C19orf26 | rs3746102 | 0.726667345 |
| SNP6 | 167754702 | TTLL2 | rs909546 | 0.731009291 |
| SNP20 | 44052992 | PIGT | rs13217 | 0.731733172 |
| SNP17 | 77040128 | C1QTNF1 | rs4789922 | 0.733955786 |
| SNP22 | 24459438 | CABIN1 | rs762273 | 0.735752143 |
| SNP22 | 24459438 | CABIN1 | rs77495199 | 0.735752143 |
| SNP1 | 246021941 | SMYD3 | rs2362587 | 0.740217611 |
| SNP17 | 80863857 | TBCD | rs2292971 | 0.740644031 |
| SNP7 | 152143744 | | rs6953943 | 0.744430423 |
| SNP22 | 44395451 | PARVB | rs1007863 | 0.74492585 |
| SNP12 | 6933787 | GPR162 | rs1051409 | 0.746496035 |
| SNP19 | 50865535 | NAPSA | rs676314 | 0.753319516 |
| SNP2 | 1520676 | TPO | rs1126799 | 0.754045206 |
| SNP2 | 1520676 | TPO | rs61734476 | 0.754045206 |
| SNP19 | 53612311 | ZNF415 | rs4803051 | 0.75471756 |
| SNP19 | 53612311 | ZNF415 | rs147077991 | 0.75471756 |
| SNP7 | 107849908 | NRCAM | rs1269621 | 0.75471756 |
| SNP9 | 8465598 | PTPRD | rs2281747 | 0.760889235 |
| SNP2 | 159195282 | CCDC148 | rs16842890 | 0.761631581 |
| SNP2 | 159196771 | CCDC148 | rs12620556 | 0.761631581 |
| SNP2 | 159196771 | CCDC148 | rs137873189 | 0.761631581 |
| SNP5 | 131411460 | CSF2 | rs25882 | 0.762589333 |
| SNP8 | 10468172 | RP1L1 | rs4840502 | 0.762826115 |
| SNP6 | 131995313 | ENPP3 | rs9493048 | 0.768884421 |
| SNP8 | 1905132 | ARHGEF10 | rs3735876 | 0.773859731 |
| SNP10 | 97990583 | BLNK | rs727852 | 0.774252571 |
| SNP6 | 6152137 | F13A1 | rs5988 | 0.774496967 |
| SNP11 | 125848261 | CDON | rs3740904 | 0.776252184 |
| SNP22 | 25601196 | CRYBB3 | rs9608378 | 0.776743595 |
| SNP1 | 235940450 | LYST | rs2273584 | 0.784643433 |
| SNP1 | 235972435 | LYST | rs3820553 | 0.784643433 |
| SNP2 | 241979550 | SNED1 | rs7571117 | 0.784888177 |
| SNP8 | 22526559 | BIN3 | rs17088526 | 0.790774534 |
| SNP4 | 57273840 | PPAT | rs11538098 | 0.792937769 |
| SNP5 | 95733112 | PCSK1 | rs6233 | 0.793972096 |
| SNP16 | 20376755 | PDILT | rs8054266 | 0.804053468 |
| SNP19 | 4950770 | UHRF1 | rs2307213 | 0.805634638 |
| SNP11 | 7022038 | ZNF214 | rs12575236 | 0.807510183 |
| SNP15 | 24923390 | C15orf2 | rs60574723 | 0.808325922 |
| SNP1 | 158736445 | OR6N1 | rs1864346 | 0.809556066 |
| SNP7 | 154862621 | HTR5A | rs6320 | 0.811860752 |
| SNP2 | 71190384 | ATP6V1B1 | rs2072462 | 0.812944944 |
| SNP17 | 39081713 | KRT23 | rs8037 | 0.813564177 |
| SNP4 | 122301622 | QRFPR | rs17438900 | 0.815574899 |
| SNP17 | 32483237 | ACCN1 | rs2228990 | 0.815876723 |
| SNP19 | 56704101 | ZSCAN5B | rs11084427 | 0.819586929 |
| SNP19 | 56704101 | ZSCAN5B | rs61747508 | 0.819586929 |
| SNP13 | 76409436 | LMO7 | rs2273997 | 0.821264557 |
| SNP11 | 14264916 | SPON1 | rs2303973 | 0.821773961 |
| SNP12 | 101732655 | UTP20 | rs56265469 | 0.822173208 |
| SNP8 | 133047071 | OC90 | rs11779306 | 0.824543295 |
| SNP11 | 197337 | ODF3 | rs3802984 | 0.824640894 |
| SNP8 | 98943545 | MATN2 | rs2290471 | 0.831207776 |
| SNP1 | 175335234 | TNR | rs1385541 | 0.831254195 |
| SNP19 | 52249680 | FPR1 | rs5030880 | 0.840537841 |
| SNP1 | 42880516 | RIMKLA | rs1055055 | 0.842873543 |
| SNP22 | 45691594 | UPK3A | rs1057356 | 0.845534364 |
| SNP1 | 5926507 | NPHP4 | rs555164 | 0.849051327 |
| SNP5 | 14368975 | TRIO | rs13189406 | 0.855133728 |
| SNP9 | 34310927 | KIF24 | rs10972048 | 0.856339857 |
| SNP11 | 11354346 | GALNTL4 | rs901553 | 0.858701594 |
| SNP20 | 3624830 | ATRN | rs2246808 | 0.859036365 |
| SNP6 | 30680916 | MDC1 | rs9262152 | 0.860417242 |
| SNP6 | 30680916 | MDC1 | rs116665440 | 0.860417242 |
| SNP2 | 54127041 | PSME4 | rs805423 | 0.862215116 |
| SNP22 | 31032920 | SLC35E4 | rs5997714 | 0.863927226 |
| SNP19 | 1357082 | MUM1 | rs713042 | 0.866130382 |
| SNP2 | 99812070 | MRPL30 | rs1044575 | 0.868181897 |
| SNP19 | 17091368 | CPAMD8 | rs8103646 | 0.869732754 |
| SNP2 | 32822957 | BIRC6 | rs2710625 | 0.876582647 |
| SNP15 | 81637284 | TMC3 | rs4278705 | 0.877609369 |
| SNP6 | 4998963 | RPP40 | rs1749144 | 0.877609369 |
| SNP12 | 2973615 | FOXM1 | rs2072360 | 0.884574928 |
| SNP10 | 113917085 | GPAM | rs2254537 | 0.884923407 |
| SNP3 | 42772038 | CCDC13 | rs12495805 | 0.89810585 |
| SNP5 | 156479426 | HAVCR1 | rs12522248 | 0.904701447 |
| SNP8 | 124138855 | WDR67 | rs16898012 | 0.908087734 |
| SNP8 | 124138855 | WDR67 | rs192939621 | 0.908087734 |
| SNP7 | 139797431 | JHDM1D | rs6950119 | 0.909102036 |
| SNP17 | 33269648 | CCT6B | rs2230553 | 0.915156352 |
| SNP19 | 41129842 | LTBP4 | rs7367 | 0.939602869 |
| SNP17 | 64025331 | CCDC46 | rs11652766 | 0.942397865 |
| SNP1 | 111857951 | CHIA | rs61756687 | 0.943233497 |
| SNP14 | 74876181 | TMEM90A | rs12590672 | 0.951416073 |
| SNP15 | 47873549 | | rs4775699 | 0.95407829 |
| SNP2 | 220402680 | ACCN4 | rs11695248 | 0.954443996 |
| SNP11 | 27076977 | | rs2305095 | 0.955326217 |
| SNP5 | 82786194 | VCAN | rs12332199 | 0.96016265 |
| SNP19 | 50140092 | RRAS | rs1865077 | 0.961960365 |
| SNP22 | 29456733 | C22orf31 | rs6005977 | 0.964194175 |
| SNP3 | 38766701 | SCN10A | rs6791171 | 0.966014083 |
| SNP4 | 46995366 | GABRA4 | rs2229940 | 0.966841214 |
| SNP17 | 49713300 | CA10 | rs11870209 | 0.975428792 |
| SNP10 | 64159333 | ZNF365 | rs3758490 | 0.978482697 |
| SNP21 | 31744127 | KRTAP13-2 | rs877346 | 0.980758052 |
| SNP1 | 159175354 | DARC | rs12075 | 0.98397315 |
| SNP19 | 17741047 | UNC13A | rs10413821 | 0.984573573 |